\documentclass[final]{siamltex}
\usepackage{verbatim}
\usepackage{graphicx, epsfig}
\usepackage{subcaption}
\usepackage{float}
\usepackage{amsmath,amssymb}
\usepackage{hyperref}
\usepackage{xcolor}
\usepackage{multirow}
\usepackage{bbm}
\usepackage{float}
\usepackage{soul}
\usepackage[makeroom]{cancel}
\usepackage[normalem]{ulem}
\usepackage{bm}
\usepackage{tikz}
\usepackage{algorithm}
\usepackage{algpseudocode}
\usepackage{comment}
\usepackage{url}
\usepackage{booktabs}
\usepackage{adjustbox}

\usepackage[]{lineno}

\title{Robust Parameter and State Estimation in Multiscale Neuronal Systems Using Physics-Informed Neural Networks}
\author{Changliang, Wei\thanks{Department of Mathematics, University of Iowa, Iowa City, IA 52246, USA. Email: 
changliang-wei@uiowa.edu.}
\and Yangyang, Wang\thanks{Department of Mathematics, Volen National Center for Complex Systems, Brandeis University, Waltham, MA 02453, USA. Email: 
yangyangwang@brandeis.edu.}
\and Xueyu Zhu \thanks{Department of Mathematics, University of Iowa, Iowa City, IA 52246. USA. Email: xueyu-zhu@uiowa.edu.}}

\begin{document}

\maketitle

\begin{abstract}
Inferring biophysical parameters and hidden state variables from partial and noisy observations is a fundamental challenge in computational neuroscience. This problem is particularly difficult for fast–slow spiking and bursting models, where strong nonlinearities, multiscale dynamics, and limited observational data often lead to severe sensitivity to initial parameter guesses and convergence failure in the methods replying on the traditional numerical forward solvers. In this work, we developed a physics-informed neural network (PINN) framework for the joint reconstruction of unobserved state variables and the estimation of unknown biophysical parameters in neuronal models. We demonstrate the effectiveness of the method on biophysical neuron models, including the Morris--Lecar model across multiple spiking and bursting regimes and a respiratory model neuron. The method requires only partial voltage observations over short observation windows and remains robust even when initialized with non-informative parameter guesses. These results suggest that PINN can deliver robust and accurate parameter inference and state reconstruction, providing a promising alternative for inverse problems in multiscale neuronal dynamics, where traditional techniques often struggle.

\end{abstract}


\pagestyle{myheadings}
\thispagestyle{plain}

\section{Introduction}
\label{sec:Intro}

 Multiple-timescale dynamics are ubiquitous in biological systems, arising from the interaction of fast and slow physiological processes \cite{keener2009mathematicalI,keener2009mathematicalII}. Neural systems provide a canonical example of such multiscale behavior \cite{ermentrout2010mathematical}, in which rapid membrane potential dynamics are tightly coupled to slower ion-channel gating variables, intracellular signaling pathways, and neuromodulatory processes. These interactions generate a rich repertoire of oscillatory phenomena, including action potentials, bursting, and mixed-mode oscillations characterized by alternating large- and small-amplitude dynamics \cite{rinzel1989analysis,izhikevich2007dynamical,moehlis2006canards,iglesias2011mixed,desroches2012mixed,wang2016multiple,wang2017timescales,phan2024mixed}. Capturing these behaviors requires mathematical models that can explicitly represent the interaction of fast and slow processes within a unified dynamical framework.

Conductance-based neuronal models provide such a framework by directly coupling membrane voltage to ion-channel kinetics and intracellular processes through systems of nonlinear ordinary differential equations, thereby exhibiting distinct timescales. For example, membrane potential typically oscillates on the order of milliseconds, while ion channel conductances and intracellular calcium-dependent processes may evolve over much slower timescales. Typical simple neuron models of membrane potential oscillations such as the Morris-Lecar model \cite{morris1981voltage} exhibit at least two distinct timescales \cite{ermentrout2010mathematical}, while more detailed formulations can involve three or more timescales \cite{nan2015understanding,wang2020complex,venkatakrishnan2025dual}. At the same time, this model capability comes at a cost: conductance-based models are highly sensitive to their underlying biophysical parameters, with small perturbations capable of inducing qualitative changes in dynamics through bifurcations marking the boundary between distinct regimes \cite{izhikevich2007dynamical}. Moreover, in most electrophysiological settings, such as current-clamp experiments \cite{johnston1994foundations}, membrane voltage is the only directly observed variable, while other state variables remain unmeasured. The combination of strong multiscale nonlinear dynamics, parameter sensitivity, and severe partial observability renders parameter estimation an intrinsically ill-posed inverse problem, particularly for models with many parameters and with pronounced timescale separation  \cite{paninski2009statistical,raue2009structural}.

Over the past few years, a range of methodologies has been developed to address neuronal state and parameter estimation under these constraints. Data assimilation approaches, including Kalman-filter–based methods and variational optimization techniques, leverage the traditional forward solvers based on governing equations to jointly infer hidden states and biophysical parameters from partial observations \cite{moye2018data,vlachas2020ukf,bano2021daily,li2023adaptiveUKF}. Bayesian inference methods tailored to neuronal systems provide principled uncertainty quantification and have been applied to a variety of conductance-based models \cite{golightly2011bayesian,calderhead2011accelerating,transtrum2015sloppiness}. More recently, simulation-based inference approaches have been explored in computational neuroscience to amortize inference across large numbers of neuronal simulations, including applications to both single-neuron and network-level models \cite{gonccalves2023sbi,zhang2022neuralsbi}. 

Despite these advances,  a key ingredient in these approaches, is their reliance on the traditional numerical ODE solvers to integrate forward in time.
Since parameter estimation proceeds sequentially through forward integration, local numerical errors can accumulate and performance becomes sensitive to the initial guesses of biological parameters. In the absence of informative prior knowledge, poor initialization may result in slow convergence, convergence to incorrect solutions, or failure due to numerical instability or nonphysical parameter values \cite{moye2018data}.  Beyond these general challenges, commonly used estimation paradigms exhibit distinct method-specific limitations. Variational approaches such as 4D-Var formulate parameter estimation as a large-scale optimization problem over the full observation window, whose dimensionality of the estimation problem grows with both the state dimension and the window length \cite{asch2016data}. 
This scaling can make 4D-Var computationally impractical for long or high-resolution neuronal recordings. In contrast, sequential filtering methods such as the Unscented Kalman Filter (UKF), while computationally efficient, typically require long observation windows or enough datasets to achieve accurate parameter estimates \cite{ghorbani2025physics}, particularly under partial observability where parameters are only weakly informed by voltage measurements \cite{moye2018data}. 
Taken together, these limitations motivate us to develop an effective alternative framework that does not rely on sequential time marching, is less sensitive to initialization, and can better exploit known model structure with partial observed state variable.

Physics-informed neural networks (PINNs) have recently emerged as a powerful framework for state and parameter estimation in dynamical systems governed by differential equations \cite{raissi2019physics}. By embedding the governing equations directly into the training loss, PINNs approximate system dynamics as global functions over the observation window, constrained simultaneously by data fidelity and physical laws, with large or sparse (noisy) data sets. It has been particularly effective for scenarios where only partial observations are available, and the remaining state variables must be inferred indirectly, as demonstrated in \cite{bertaglia2022asymptotic}. Importantly, this formulation does not rely on time marching. As a result, PINNs avoid the accumulation of local integration errors and reduce sensitivity to informative initial parameter guesses that are often required to maintain convergence and biological plausibility in traditional data assimilation methods.

Within the context of neuronal dynamical systems, several recent studies have established the feasibility of PINN-based formulations for joint state and parameter estimation. For example, standard PINNs have been applied to Hodgkin–Huxley types of models to recover hidden gating variables and biophysical parameters from voltage recordings \cite{alamstotter2022hhpinn}.
Recently, it has also been explored for Hodgkin-Huxley models under full observability, focusing on learning system dynamics without addressing parameter estimation \cite{kainth2025physics}.
While these studies consider different observational and inferential settings, together they highlight the flexibility of physics-informed learning for modeling neuronal dynamics and position PINNs as a promising alternative to existing inverse problem approaches in neuroscience.
Nevertheless, the application of PINNs to complex neuronal dynamics is still at an early stage, with persistent challenges in effectiveness and robustness, particularly for stiff, multiscale, and regime-switching systems. Recent methodological advances, such as Fourier feature embedding, adaptive loss balancing, and residual scaling, have improved the effectiveness of PINN training for high-frequency and multiscale problems \cite{wang2021understanding}, thereby motivating further investigations of PINN-based approaches for biophysical neuronal models.

In this work, we develop a physics-informed neural network framework for joint state reconstruction and parameter estimation in biophysical neuronal models under partial and noisy observability. The main contributions are as follows:

\begin{itemize}
    \item We combine data-driven Fourier embeddings extracted from voltage recordings with trainable Fourier features to represent both dominant oscillatory components and remaining fast--slow temporal dynamics.
    
    \item We introduce a two-stage training strategy: the network is first pretrained using voltage observations to capture the representation of observed state variable, followed by physics-informed training that incorporates the full system of state equations and biophysical parameters.
    
    \item We incorporate recent methodological advances in PINNs, including random weight factorization, adaptive loss balancing, and residual scaling, to improve training robustness for nonlinear and multiscale problems. In addition, we employ separate learning rate schedules for the network parameters and the biophysical parameters to  improve convergence.
    
    \item We systematically evaluate the proposed framework across multiple conductance-based neuronal models with varying dynamical complexity, including spiking and bursting regimes of the Morris--Lecar model \cite{ermentrout2010mathematical} and a biophysical model of pre-B\"otzinger complex respiratory neurons \cite{wang2016multiple}. In particular, we focus on short observation windows containing only one or two oscillatory cycles, a regime in which limited data substantially restricts parameter identifiability and renders the inverse problem more challenging. Nonetheless, our results demonstrate that the proposed PINN framework remains robust even under non-informative and regime-mismatched initial parameter guesses.
\end{itemize}

The rest of the paper is organized as follows. Section~\ref{sec:bg} introduces the problem setup, methodological details, and necessary background. In particular, Section~\ref{subsec:PINNs} reviews the formulation of physics-informed neural networks (PINNs), and Section~\ref{subsec:network_architecture} describes the network architecture and training strategy adopted in this work. Section~\ref{sec:NumExample} presents a series of numerical examples, including both spiking and bursting dynamics that demonstrate the performance of the proposed approach across different neuronal models and noise settings.   All codes
and data accompanying this manuscript will be made publicly available. 

\section{Problem Setup and Background}
\label{sec:bg}
\label{subsec:problem}
We focus on multiscale systems, with conductance-based biophysical neuronal models serving as a representative class of biologically motivated fast-slow systems. These models were firstly introduced by Alan Hodgkin and Andrew Huxley in 1952 \cite{hodgkin1952quantitative} to describe the action potential generation in the squid giant axon. Using an equivalent circuit framework, the dynamics of membrane voltage $V$ are described by
\begin{equation}
    C_m \frac{dV}{dt} = I_{\textrm{app}} - \sum_{\textrm{ion}} I_{\textrm{ion}},
\end{equation}
where $I_{\textrm{app}}$ is the externally injected current and $I_{ion}$ denotes ionic currents associated with different channel species, such as sodium ($\textrm{Na}^+$), potassium ($\textrm{K}^+$), or calcium ($\textrm{Ca}^{2+}$) channels. For example, the potassium $\textrm{K}^+$ current takes the standard Hodgkin-Huxley form
\begin{equation}
    I_{\textrm{K}} = g_{\textrm K} n^4 (V - E_{\textrm{K}}),
    \label{eq:classical_current}
\end{equation}
where $g_{\textrm K}$ is the maximal conductance, $E_{\textrm{K}}$ is the potassium reversal potential, and $n$ is a gating variable with $n^4$ representing the probability that a $\rm K^+$ channel is open. The dynamics of gating variables are typically governed by first-order kinetics
\begin{equation}
    \frac{dx}{dt} = \alpha_x(V) (1-x) - \beta_x(V)x = \frac{x_{\infty}(V) - x}{\tau_x(V)},
\end{equation}
where $x_{\infty}(V) = \frac{\alpha_x(V)}{\alpha_x(V) + \beta_x(V)}$ denotes the steady-state activation and $\tau_x(V) =\frac{1}{\alpha_x(V) + \beta_x(V)}$ is the voltage-dependent time constant.

To  unify notation independent of the specific neuronal model, we denote the state vector as 
\begin{equation}
    \mathbf{X}(t) = \begin{bmatrix}
    V(t) \\
    x_1(t) \\
    x_2(t) \\
    \vdots \\
    x_m(t)
    \end{bmatrix},
\end{equation}
where $x_1, \cdots, x_m$ represent all gating variables and slow variables. Then, the general conductance-based model can be written as
\begin{equation}
    \frac{d\mathbf{X}}{dt} = \mathbf{F}(\mathbf{X}(t); \boldsymbol{\lambda}),
    \label{eqn:general_model}
\end{equation}
where $\boldsymbol{\lambda}$ represents all biophysical parameters, such as $g_{\textrm{K}}$ and $E_{\textrm{K}}$ in \eqref{eq:classical_current}. 

In this work, we assume that only the membrane voltage $V$ is observed, as in a current-clamp protocol where one records voltage traces in response to injected currents $I_{\rm app}$ \cite{johnston1994foundations}. In addition, measurements are assumed to be contaminated by additive noise. Specifically,
\begin{equation}
    V^{\mathrm{obs}}(t_i) = V^{\mathrm{true}}(t_i) + \varepsilon_i, \qquad i = 1,\dots,N_u,
\end{equation}
where $\varepsilon_i$ denotes measurement noise and $N_u$ denotes the total number of observed voltage data points. Given the partially (noisy) observed system, the goal is to estimate important biophysical parameters $\boldsymbol{\lambda}$ and reconstruct the full system state trajectory.  

\subsection{Physics Informed Neural Networks PINNs}
\label{subsec:PINNs} 

Over the past several years, scientific machine learning (SciML) methods for solving inverse problems have attracted increasing attention \cite{baker_2019,karniadakis2021physics}, and one of the most popular approaches is the Physics Informed Neural Networks (PINNs). PINNs enable the simultaneous approximation of the solution of differential equations (DEs) and the estimation of unknown physical parameters.

Given a dynamical system \eqref{eqn:general_model}, the state variable is approximated by a fully connected neural network surrogate $\widehat{\mathbf{X}}(t;\boldsymbol{\theta})$, where $\boldsymbol{\theta}\in\mathbb{R}^{N_\theta}$ denotes the collection of neural network parameters. In the inverse problem setting, both the neural network parameters and the unknown biophysical parameters are treated as trainable variables. We therefore define the full set of learnable parameters as
\[
\boldsymbol{\xi} = [\boldsymbol{\theta},\, \boldsymbol{\lambda}],
\]
where $\boldsymbol{\lambda}$ represents the unknown physiological parameters to be estimated.

Let $\mathcal{D}_u$ denote the observational data set, and $\mathcal{D}_f$ the residual collocation set used to enforce the governing equations. These sets are defined as
\begin{align}
\mathcal{D}_u &= \{(t_u^i, \mathbf{X}_u^i)\}_{i=1}^{N_u}, \\
\mathcal{D}_f &= \{t_f^i\}_{i=1}^{N_f},
\end{align}
where $\mathbf{X}_u^i$ denotes the observed neuronal state variable at time $t_u^i$.

At each residual point $t_f^i$, the governing equation residual  of \eqref{eqn:general_model} is defined by
\begin{equation}
\mathcal{N}(t_f^i; \boldsymbol{\xi}) 
= \frac{d\widehat{\mathbf{X}}}{dt}(t_f^i;\boldsymbol{\xi})
- \mathbf{F}\big(\widehat{\mathbf{X}}(t_f^i;\boldsymbol{\xi});\, \boldsymbol{\lambda}\big),
\end{equation}
where the temporal derivative is evaluated via automatic differentiation.

Following the standard PINNs formulation \cite{RAISSI2019686,wang2023}, the total loss function is constructed as a weighted combination of the data misfit and the residual loss:
\begin{equation}
\mathcal{L}(\boldsymbol{\xi}) 
= \omega_u \mathcal{L}_u(\boldsymbol{\xi})
+ \omega_f \mathcal{L}_f(\boldsymbol{\xi}),
\end{equation}
with
\begin{align}
\mathcal{L}_{u}(\boldsymbol{\xi}) &= \frac{1}{N_u} \sum_{i=1}^{N_u}
\Big\|\mathbf{X}_u^i - \widehat{\mathbf{X}}(t_u^i;\boldsymbol{\xi})\Big\|^2, \\ 
\mathcal{L}_{f}(\boldsymbol{\xi}) &= \frac{1}{N_f} \sum_{i=1}^{N_f}
\Big\|\mathcal{N}(t_f^i;\boldsymbol{\xi})\Big\|^.
\end{align}
Here, $\omega_u$, and $\omega_f$ are positive weighting coefficients balancing each loss.

In practice, the parameter vector $\boldsymbol{\xi}$ is optimized using gradient-based methods such as Adam or L-BFGS \cite{kingma2017adam, liu1989lbfgs}. This formulation allows the neuronal state and the unknown biophysical parameters to be inferred simultaneously within a unified physics-informed learning framework.

\subsection{Network Architecture and Training Strategy}
\label{subsec:network_architecture} 

In this section, we present the design of the network architecture and training strategy underlying the proposed PINN framework, with the central goal of improving representation capacity, optimization stability, and robustness for inverse problems in nonlinear neuronal dynamics.

\subsubsection{Fourier Feature Embedding}
\label{subsubsec:ffe} 

Neuronal signals typically exhibit intrinsically oscillatory and multi-scale temporal structures, characterized by the coexistence of low-frequency bursting patterns and high-frequency spiking behaviors. Standard fully connected neural networks driven solely by raw time inputs often suffer from spectral bias and therefore struggle to accurately resolve such multi-scale oscillatory behaviors. To enhance the expressive capacity of the network in the frequency domain, we introduce a Fourier feature embedding strategy that integrates data-driven dominant frequencies extracted from observations with additional trainable frequencies.

Following the general random Fourier feature (RFF) formulation \cite{tancik2020,WANG2021113938}, a Fourier feature mapping is defined by
\begin{equation}
\boldsymbol{\gamma}(t) = 
\begin{bmatrix}
\cos(\mathbf{B}t) \\
\sin(\mathbf{B}t)
\end{bmatrix},
\end{equation}
where $\mathbf{B}\in\mathbb{R}^{m}$ denotes a frequency vector. In classical RFF-based embeddings, the entries of $\mathbf{B}$ are sampled from a Gaussian distribution $\mathcal{N}(0,\sigma^2)$ and remain fixed throughout training. 

Such random embeddings are particularly effective when the observational data are sparse, making the underlying frequency content impossible to estimate directly from data. In that setting, the randomly sampled bands provide a broad frequency coverage that compensates for the lack of information. While such random embeddings can partially alleviate spectral bias, they do not explicitly reflect the dominant oscillatory structures present in neuronal data. In contrast, our problem lies in a fundamentally different regime. The observed neuronal signals are densely sampled, and their oscillatory structure which included dominant frequencies, can be precisely extracted using Fourier transform analysis. Therefore, rather than relying on randomly drawn frequencies, we construct a deterministic, data-driven Fourier embedding using Fast Fourier Transform (FFT). \\

\noindent\textbf{FFT-based Fourier Embedding.}
To construct a data-informed frequency representation, we perform a Fast Fourier Transform (FFT) on the observed voltage $\{V^{\mathrm{obs}}(t_i)\}_{i=1}^{N}$. The discrete Fourier transform is given by
\begin{equation}
\hat{V}(b_k) = \sum_{j=1}^{N} V_j^{\mathrm{obs}} e^{-2\pi i b_k j},
\qquad k=1,\ldots,K,
\end{equation}
and the corresponding power spectral density (PSD) is defined as
\begin{equation}
\mathrm{PSD}(b_k) = \left|\hat{V}(b_k)\right|^2.
\end{equation}
The frequencies are then sorted in descending order according to their PSD magnitudes,
\begin{equation}
\mathrm{PSD}(b_1) \ge \mathrm{PSD}(b_2) \ge \cdots \ge \mathrm{PSD}(b_K).
\end{equation}
We introduce the cumulative normalized spectral energy
\begin{equation}
E_m = \frac{\sum_{k=1}^{m} \mathrm{PSD}(b_k)}{\sum_{k=1}^{K} \mathrm{PSD}(b_k)},
\end{equation}
and select the minimal integer $m^*$ such that $E_{m^*} \ge p/100$. The resulting dominant frequency set is denoted by
\[
\mathbf{B}_{\mathrm{FFT}} = \{b_k\}_{k=1}^{m^*}.
\]
The FFT-based Fourier feature embedding of time is then constructed as
\begin{equation}
\boldsymbol{\gamma}(t) =
\big[
\sin(b_1 t),\, \cos(b_1 t),\,
\sin(b_2 t),\, \cos(b_2 t),\,
\ldots,\,
\sin(b_{m^*} t),\, \cos(b_{m^*} t)
\big].
\label{eq:B-FFT}
\end{equation}

\begin{figure}[htbp]
    \centering
    \begin{subfigure}[b]{0.48\textwidth}
        \centering
        \includegraphics[width=\textwidth]{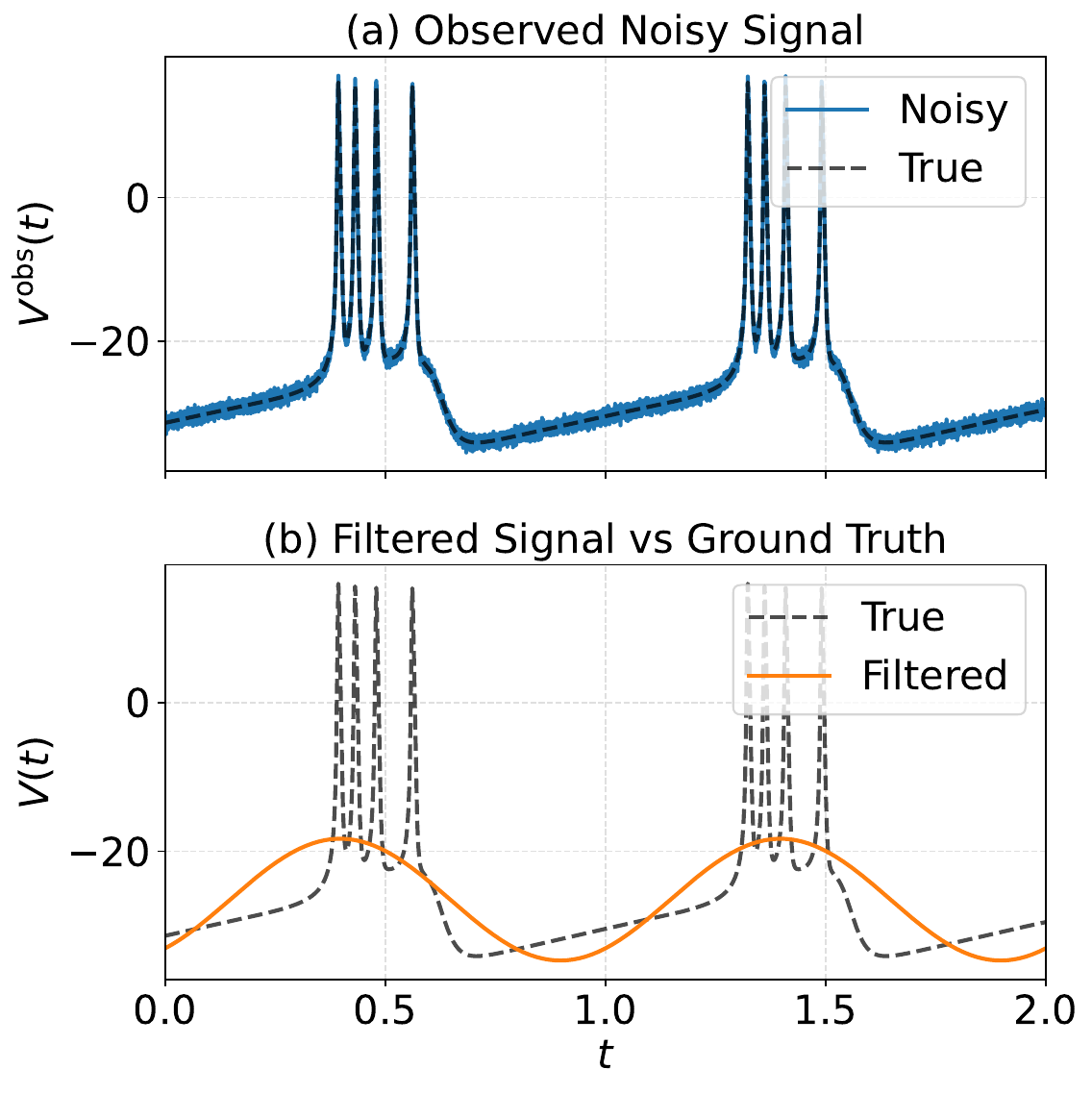}
        \caption{$p=95$}
        \label{fig:signal_filtering_comparison_0.95}
    \end{subfigure}
    \begin{subfigure}[b]{0.48\textwidth}
        \centering
        \includegraphics[width=\textwidth]{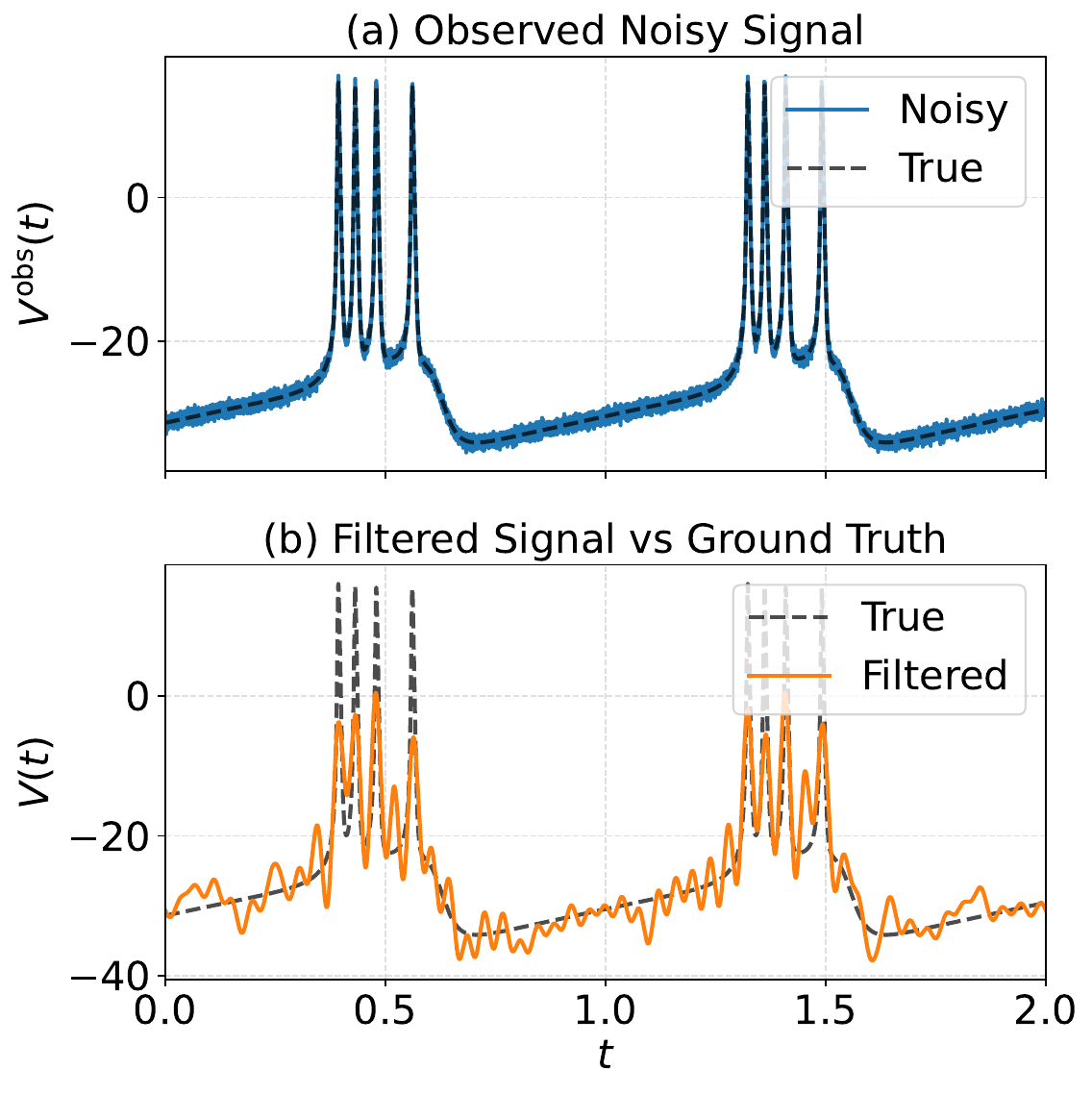}
        \caption{$p=99$}
        \label{fig:signal_filtering_comparison_0.99}
    \end{subfigure}  
    \caption{Comparison of the signal filtering results under different parameter choices. The left panel corresponds to $p=95$, while the right panel corresponds to $p=99$. In each subplot, the observed noisy signal is filtered and compared against the ground-truth signal.}
    \label{fig:signal_filtering_comparison}
\end{figure}

We remark that the hyperparameter $p$ controls the proportion of retained spectral energy. In our numerical experiments, $p$ is typically chosen as $95$ or $99$. As an illustrative example, we consider the bursting Morris--Lecar model \eqref{BR_ML_Eq_V}-\eqref{BR_ML_Eq_Ca} which is introduced in Section \ref{sec:NumExample}. When $p=95$, only two dominant frequencies are retained, and the resulting filtered signal fails to capture the principal bursting oscillations (Figure~\ref{fig:signal_filtering_comparison_0.95}). In contrast, when $p=99$, a total of $19$ dominant frequencies are selected, and the reconstructed signal closely matches the ground-truth voltage trace (Figure~\ref{fig:signal_filtering_comparison_0.99}). For noisy observations, $p$ should not be chosen excessively large, since high-frequency components may be dominated by noise rather than the intrinsic neuronal dynamics. \\

\noindent\textbf{Hybrid Frequency Embedding.}
For unobserved state variables, we adopt a hybrid frequency selection strategy. First, the dominant frequency set $\mathbf{B}_{\mathrm{FFT}}$ extracted from the observed voltage $V(t)$ is retained, under the assumption that all neuronal state variables share common fundamental oscillatory timescales. To further enhance representation expressivity, we introduce an additional set of trainable frequencies
\[
\mathbf{B}_{\mathrm{train}} = \{b_j^{\mathrm{train}}\}_{j=1}^{m^*},
\]
which are initialized from a uniform distribution over the FFT-derived frequency range,
\begin{equation}
b_j^{\mathrm{train}} \sim \mathcal{U}(b_{m^*},\, b_1), 
\qquad j=1,\ldots,m^*,
\end{equation}
where $b_1$ and $b_{m^*}$ denote the maximum and minimum FFT-derived frequencies, respectively. The total frequency set for each unobserved variable is therefore given by
\[
\mathbf{B}_{\mathrm{unobs}} = \mathbf{B}_{\mathrm{FFT}} \cup \mathbf{B}_{\mathrm{train}}.
\]
The resulting Fourier feature embedding is
\begin{equation}
\boldsymbol{\gamma}(t) =
\big[
\sin(b_1 t),\, \cos(b_1 t),\,
\ldots,\,
\sin(b_{2m^*} t),\, \cos(b_{2m^*} t)
\big],
\qquad b_j \in \mathbf{B}_{\mathrm{unobs}}.
\end{equation}
This hybrid embedding simultaneously leverages data-driven dominant oscillatory structures and adaptive trainable frequencies, thereby improving the network’s ability to represent multi-scale neuronal dynamics.

\subsubsection{Random Weight Factorization}
\label{subsubsec:rwf}

Despite the use of enhanced Fourier feature embeddings, training PINNs for stiff and strongly nonlinear neuronal dynamical systems remains a highly ill-conditioned optimization problem, due to the coexistence of multiple time scales and sharp transitions in the state trajectories. To further improve the conditioning of the optimization landscape and stabilize gradient propagation across layers, we incorporate the Random Weight Factorization (RWF) strategy proposed by Wang et al.~\cite{wang2022randomweightfactorizationimproves}. RWF improves training stability by explicitly decoupling the scale and direction of the neural network weights.

Specifically, for each hidden layer $l=1,2,\ldots,L$, the weight matrix $\mathbf{W}^{(l)} \in \mathbb{R}^{d_l \times d_{l+1}}$ is reparameterized as
\begin{equation}
\mathbf{W}^{(l)} = \mathrm{diag}\big(\exp(\mathbf{s}^{(l)})\big)\, \mathbf{W}_N^{(l)},
\end{equation}
where $\mathbf{s}^{(l)} \in \mathbb{R}^{d_l}$ is a trainable vector of scalar factors that controls the magnitude of each neuron, and $\mathbf{W}_N^{(l)} \in \mathbb{R}^{d_l \times d_{l+1}}$ represents the corresponding normalized weight directions. The exponential parameterization guarantees strictly positive scaling and prevents degeneracy due to vanishing or negative scales.

Under this formulation, the scale parameters $\mathbf{s}^{(l)}$ and the directional parameters $\mathbf{W}_N^{(l)}$ are optimized independently, allowing the network to adapt both activation amplitudes and feature directions in a more flexible and numerically stable manner. The full set of trainable network parameters is therefore given by
\[
\boldsymbol{\theta} = \{\mathbf{s}^{(l)},\, \mathbf{W}_N^{(l)},\, \mathbf{b}^{(l)}\}_{l=1}^{L}.
\]

Different from standard network initialization, the introduction of RWF requires each weight matrix to be further reparameterized by a trainable scalar factor. Specifically, for each layer, a scalar variable $s^{(l)}$ is introduced and initialized by sampling from a normal distribution $\mathcal{N}(\mu, \sigma^2)$. The base matrices $\mathbf{W}_N^{(l)}$ and biases $\mathbf{b}^{(l)}$ are initialized using the Glorot scheme \cite{glorot2010}.  We remark that the original RWF formulation recommends $\mu=1.0$ and $\sigma=0.1$. However, in the context of neuronal dynamics characterized by sharp voltage transitions and stiffness, we observed that $\mu=1.0$ often leads to excessively large effective weights at early training stages, resulting in unstable gradient updates. To alleviate this issue, we adopt a smaller mean value $\mu=0.5$ with $\sigma=0.1$ in most of our experiments, which consistently provides improved training stability and faster convergence.

\subsubsection{Two-Stage Training Strategy}
\label{subsubsec:pre-trian} 

In contrast to the traditional PINN setup, where the observational data are typically sparse, our dataset provides relatively dense measurements of the voltage $V^{\mathrm{obs}}(t)$ even over a short observation window. This allows  a data-driven pre-training stage for the observed state variable with prior to incorporating physics-informed constraints. This pre-training stage is easier to optimize than training with multiple loss terms simultaneously. Following this data-driven pre-training, we proceed to a second training stage focused on recovering the unobserved state variables and the associated physical parameters by incorporating physics-informed loss terms.

\vspace{0.4em}
\noindent\textbf{Data-driven pre-training stage.}
We first train a fully connected neural network $\widehat{V}(t)$ for the voltage variable $V$ using only the data misfit term. The loss function is defined as
\begin{equation}
\mathcal{L}_{u}(\boldsymbol{\theta}_V)
= \frac{1}{N_u}\sum_{i=1}^{N_u}
\left|\widehat{V}(t_i;\boldsymbol{\theta}_V) - V^{\mathrm{obs}}(t_i)\right|^2,
\end{equation}
where $\boldsymbol{\theta}_V$ denotes the weights of the neural network surrogate $\widehat{V}(t;\boldsymbol{\theta}_V)$.

\vspace{0.4em}
\noindent\textbf{Physics-informed training stage.}
After the data pre-training stage described above, during which the network output $\widehat{V}(t)$ is fitted to the observed voltage data, we proceed to the second stage of training, which focuses on recovering the unobserved variables and physical parameters. In this stage, the loss function is defined as
\begin{equation}
\mathcal{L}_{f}(\boldsymbol{\xi}) 
= \frac{1}{N_f} \sum_{i=1}^{N_f}
\|\mathcal{N}(t_f^i; \boldsymbol{\xi})\|_{\boldsymbol{\omega}}^2,
\end{equation}
where $\|\cdot\|_{\boldsymbol{\omega}}$ denotes the weighted norm.

\begin{figure}
    \centering
    \includegraphics[width=\linewidth]{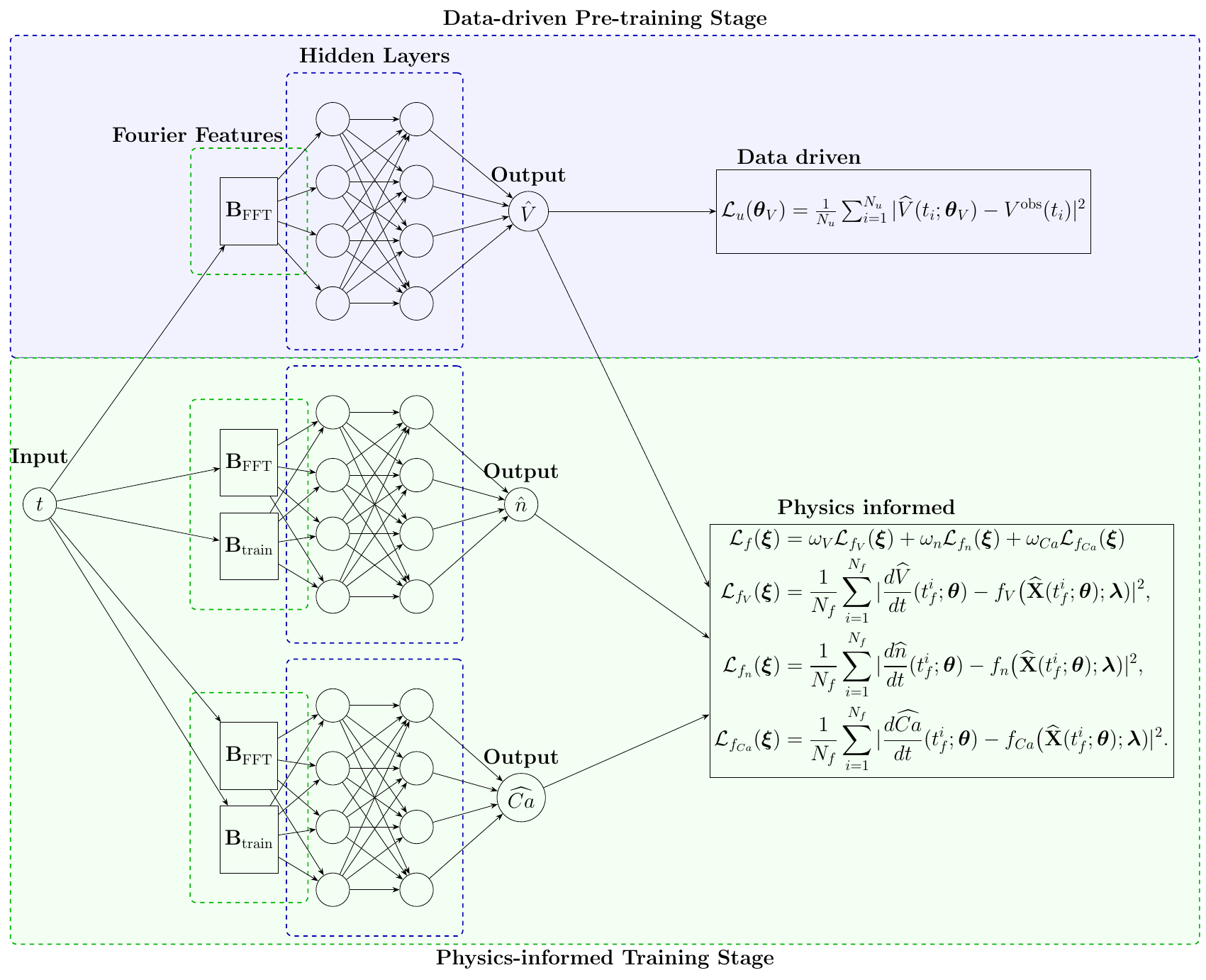}
    \caption{Illustration of the architecture of the proposed PINN, using model \eqref{eq:ex_model_de1}--\eqref{eq:ex_model_de3} with state variables $(V, n, \mathrm{Ca})$ as an illustrative example.}
    \label{fig:PINN_Structure}
\end{figure}

Using the bursting Morris--Lecar model \eqref{BR_ML_Eq_V}-\eqref{BR_ML_Eq_Ca} as an example, the unobserved variables are two-dimensional, and the state vector can be written as
\begin{equation}
\mathbf{X}(t) =
\begin{bmatrix}
V(t) \\
n(t) \\
Ca(t)
\end{bmatrix},
\end{equation}
with the detailed model formulation is introduced in Section \ref{sec:NumExample}.
The conductance-based model can then be written as
\begin{align}
\frac{dV}{dt} &= f_V(\mathbf{X}(t); \boldsymbol{\lambda}), \label{eq:ex_model_de1} \\
\frac{dn}{dt} &= f_n(\mathbf{X}(t); \boldsymbol{\lambda}), \label{eq:ex_model_de2} \\
\frac{dCa}{dt} &= f_{Ca}(\mathbf{X}(t); \boldsymbol{\lambda}). \label{eq:ex_model_de3}
\end{align}
Accordingly, the loss function can be rewritten as
\begin{equation}
\mathcal{L}_{f}(\boldsymbol{\xi})
= \omega_V \mathcal{L}_{f_V}(\boldsymbol{\xi})
+ \omega_n \mathcal{L}_{f_n}(\boldsymbol{\xi})
+ \omega_{Ca} \mathcal{L}_{f_{Ca}}(\boldsymbol{\xi}),
\label{eq:ex_model_loss_function}
\end{equation}
where $\mathcal{L}_{f_V}(\boldsymbol{\xi})$, $\mathcal{L}_{f_n}(\boldsymbol{\xi})$, and $\mathcal{L}_{f_{Ca}}(\boldsymbol{\xi})$ denote the residual losses of the differential equations associated with the voltage and the two additional state variables, respectively. They are defined as
\begin{align}
\mathcal{L}_{f_V}(\boldsymbol{\xi})
&= \frac{1}{N_f}\sum_{i=1}^{N_f}
\left|
\frac{d\widehat{V}}{dt}(t_f^i; \boldsymbol{\theta})
- f_V\big(\widehat{\mathbf{X}}(t_f^i; \boldsymbol{\theta}); \boldsymbol{\lambda}\big)
\right|^2, \\
\mathcal{L}_{f_n}(\boldsymbol{\xi})
&= \frac{1}{N_f}\sum_{i=1}^{N_f}
\left|
\frac{d\widehat{n}}{dt}(t_f^i; \boldsymbol{\theta})
- f_n\big(\widehat{\mathbf{X}}(t_f^i; \boldsymbol{\theta}); \boldsymbol{\lambda}\big)
\right|^2, \\
\mathcal{L}_{f_{Ca}}(\boldsymbol{\xi})
&= \frac{1}{N_f}\sum_{i=1}^{N_f}
\left|
\frac{d\widehat{Ca}}{dt}(t_f^i; \boldsymbol{\theta})
- f_{Ca}\big(\widehat{\mathbf{X}}(t_f^i; \boldsymbol{\theta}); \boldsymbol{\lambda}\big)
\right|^2.
\end{align}
The coefficients $\omega_V$, $\omega_n$, and $\omega_{Ca}$ are weights that balance the contribution of each term.  The corresponding architecture of the network is illustrated in Fig.\ref{fig:PINN_Structure}.

\subsubsection{Residual Loss Balancing Strategy}
\label{subsubsec:loss_balancing} 

For our partially observed neuron model, we face the challenge of balancing multiple residual losses arising from distinct components of the underlying dynamical system. Since these residuals correspond to different equations or state variables with inherently different scales and sensitivities, naive aggregation without proper weighting may lead to biased training dynamics, in which the optimization procedure overly focuses on only a subset of the residuals. To mitigate this issue, we adopt a gradient-based loss balancing strategy \cite{wang2023}, which dynamically adjusts the weights associated with each loss term to equalize their contributions during training. Specifically, we assign weights $\omega_j$ to each residual loss $\mathcal{L}_{f_j}(\boldsymbol{\xi})$ such that the gradient norms of their weighted losses are approximately equal: 
\begin{equation}
\|\omega_{j} \nabla_{\boldsymbol{\xi}} \mathcal{L}_{f_j}(\boldsymbol{\xi})\|
\approx \sum_{j} \|\nabla_{\boldsymbol{\xi}} \mathcal{L}_{f_j}(\boldsymbol{\xi})\|.
\end{equation}

For model \eqref{eq:ex_model_de1}--\eqref{eq:ex_model_de3}, the loss function in the physics-informed training stage is given by \eqref{eq:ex_model_loss_function}. We compute the intermediate weights
$\hat{\boldsymbol{\omega}} = (\hat{\omega}_{f_V}, \hat{\omega}_{f_n}, \hat{\omega}_{f_{Ca}})$ as
\begin{align}  
\hat{\omega}_{f_V}
&= \frac{
\|\nabla_{\boldsymbol{\xi}} \mathcal{L}_{f_V}(\boldsymbol{\xi})\|
+ \|\nabla_{\boldsymbol{\xi}} \mathcal{L}_{f_n}(\boldsymbol{\xi})\|
+ \|\nabla_{\boldsymbol{\xi}} \mathcal{L}_{f_{Ca}}(\boldsymbol{\xi})\|
}{
\|\nabla_{\boldsymbol{\xi}} \mathcal{L}_{f_V}(\boldsymbol{\xi})\|
}, \\ 
\hat{\omega}_{f_n}
&= \frac{
\|\nabla_{\boldsymbol{\xi}} \mathcal{L}_{f_V}(\boldsymbol{\xi})\|
+ \|\nabla_{\boldsymbol{\xi}} \mathcal{L}_{f_n}(\boldsymbol{\xi})\|
+ \|\nabla_{\boldsymbol{\xi}} \mathcal{L}_{f_{Ca}}(\boldsymbol{\xi})\|
}{
\|\nabla_{\boldsymbol{\xi}} \mathcal{L}_{f_n}(\boldsymbol{\xi})\|
}, \\
\hat{\omega}_{f_{Ca}}
&= \frac{
\|\nabla_{\boldsymbol{\xi}} \mathcal{L}_{f_V}(\boldsymbol{\xi})\|
+ \|\nabla_{\boldsymbol{\xi}} \mathcal{L}_{f_n}(\boldsymbol{\xi})\|
+ \|\nabla_{\boldsymbol{\xi}} \mathcal{L}_{f_{Ca}}(\boldsymbol{\xi})\|
}{
\|\nabla_{\boldsymbol{\xi}} \mathcal{L}_{f_{Ca}}(\boldsymbol{\xi})\|
}.
\end{align}
Here, $\|\cdot\|$ denotes the $L^2$ norm. The weights
$\boldsymbol{\omega} = (\omega_{f_V}, \omega_{f_n}, \omega_{f_{Ca}})$
are then updated using a moving average scheme,
\begin{equation}
\boldsymbol{\omega}
= \alpha \boldsymbol{\omega}
+ (1-\alpha)\hat{\boldsymbol{\omega}},
\end{equation}
where the parameter $\alpha$ controls the balance between the previous and newly computed weights. The initial values of all weights are set to $1$, and we choose $\alpha = 0.9$ in our experiments. 

To enhance numerical stability and prevent weight explosion when a residual gradient becomes very small, we add a small constant $\varepsilon$ to each gradient norm - a modification that distinguishes our approach from the original approach in \cite{wang2023}. Specifically, the gradient norm is adjusted as follows:
\begin{equation}
\|\nabla_{\boldsymbol{\xi}} \mathcal{L}_{f_j}(\boldsymbol{\xi})\|
\quad\longrightarrow\quad
\|\nabla_{\boldsymbol{\xi}} \mathcal{L}_{f_j}(\boldsymbol{\xi})\| + \varepsilon.
\end{equation}
Accordingly, the estimation of the weights becomes
\begin{equation}
\hat{\omega}_{f_j}
= \frac{
\sum_{k} \big( \|\nabla_{\boldsymbol{\xi}} \mathcal{L}_{f_k}\| + \varepsilon \big)
}{
\|\nabla_{\boldsymbol{\xi}} \mathcal{L}_{f_j}\| + \varepsilon
}.
\end{equation}
This modification differs from the original gradient-based loss balancing strategy \cite{wang2023} by introducing a small constant $\varepsilon$, which is designed for two primary purposes. First, when a particular residual has been well learned, its gradient norm may become extremely small. Using such a value in the denominator would cause the corresponding weight $\omega_j$ to grow excessively large. By adding $\varepsilon$, we impose a lower bound on the denominator and prevent the weights from blowing up. Second, adding $\varepsilon$ reduces the sensitivity of the weights to gradient noise. This is particularly important in the presence of noisy voltage observations, where gradient norms can fluctuate significantly. The term $\varepsilon$ therefore acts as an additive smoothing factor on the gradient norms, leading to more stable and consistent weight updates. Based on our empirical experience, we choose $\varepsilon = 1\times 10^{-6}$ or $1\times 10^{-8}$.

\subsubsection{Learning rate schedule}

Since the neural network parameters and the biophysical parameters exhibit fundamentally different optimization dynamics and time scales, we adopt separate learning rate schedules for these two parameter groups. Let $\boldsymbol{\theta}$ denote the trainable network parameters, including the trainable Fourier frequencies and network weights, and let $\boldsymbol{\lambda}$ denote the biophysical parameters. Their updates at iteration $k$ are given by
\begin{align}
\boldsymbol{\theta}_{k+1} &= \boldsymbol{\theta}_{k} - \eta_{\theta}(k)\, \nabla_{\boldsymbol{\theta}} \mathcal{L}(\boldsymbol{\theta}_k,\boldsymbol{\lambda}_k), \\
\boldsymbol{\lambda}_{k+1} &= \boldsymbol{\lambda}_{k} - \eta_{\lambda}(k)\, \nabla_{\boldsymbol{\lambda}} \mathcal{L}(\boldsymbol{\theta}_k,\boldsymbol{\lambda}_k),
\end{align}
where $\nabla_{\boldsymbol{\theta}} \mathcal{L}$ and $\nabla_{\boldsymbol{\lambda}} \mathcal{L}$ denote the gradients of the total loss function with respect to the network parameters and the biophysical parameters, respectively. The two learning rates $\eta_{\theta}(k)$ and $\eta_{\lambda}(k)$ follow independent decay schedules, allowing the network parameters and the physical parameters to evolve at different time scales during training. 

In practice, both learning rates follow an exponential decay schedule with a staircase strategy. This piecewise-constant exponential decay schedule ensures sufficiently large learning rates during the early stage of training to promote efficient exploration of the parameter space, while gradually reducing the step sizes to enable stable refinement and accurate convergence during the later stage of training. More detailed hyperparameter choices for learning rate schedule will be discussed in Section \ref {sec:NumExample}.

\section{Numerical Examples}
\label{sec:NumExample} 

In this section, we evaluate the performance of the proposed PINN framework through a series of numerical experiments using three neuron models settings: (1) a spiking Morris--Lecar model (Subsection \ref{subsec:Ex1}), a canonical conductance-based model for studying neuronal excitability \cite{morris1981voltage,ermentrout2010mathematical}, (2) a bursting Morris--Lecar model (Subsection \ref{subsec:Ex2}), and (3) a more biologically detailed model of pre-B\"{o}tzinger complex (pBC) respiratory neurons (Subsection \ref{subsec:Ex3}), a group of neurons in the brainstem that is critical for driving inspiration \cite{smith1991pre,butera1999models,del2002persistent,wang2016multiple}. These numerical experiments are designed to access the quality and robustness of biophysical parameter estimation and the accuracy of state reconstruction under different dynamical behaviors and noise levels. For the Morris--Lecar model, we consider three excitability mechanisms characterized by distinct bifurcation structures underlying repetitive spiking, including Hopf, saddle-node on an invariant circle (SNIC), and homoclinic (HC) bifurcations, as well as bursting dynamics in both square-wave and elliptic bursting regimes 
\cite{ermentrout2010mathematical}. 

Detailed settings of data generation, neural network design, training strategies, model equations, and error metrics are provided in the subsequent subsections.

\vspace{0.4em}
\noindent\textbf{Data Generation.}
For the Morris--Lecar model, the ground-truth data $V^{\rm true}(t)$ are generated by numerically integrating the governing differential equations using a modified Euler method \cite{moye2018data} with a fixed time step of $\Delta t = 0.1$ ms. The total simulation time $T$ is set to $0.2$ s for the spiking Morris--Lecar model and $2$ s for the bursting Morris--Lecar model. The observation time step is chosen to be identical to the simulation time step, i.e., $\Delta t_{\mathrm{obs}} = \Delta t$.

For the pBC model, the system is first integrated up to the total simulation time $T=6$ s with a time step of $\Delta t = 0.01$ ms. The observation data is collected by  uniformly down-sampling the simulated trajectory in every ten time steps, yielding an effective observation time step of $\Delta t_{\text{obs}} = 0.1$ ms. The numbers of observed voltage data points for each model are summarized in Table~\ref{tab:total_number_obs}. 

\begin{table}[t]
    \centering
    \caption{Total number of data points for the observed voltage signal for each model.}
    \label{tab:total_number_obs}
    \begin{tabular}{c|c|c}
        \hline 
        Model & $T$ & Data size \\
        \hline 
        Spiking Morris--Lecar & 0.2 s & 2001 \\
        Bursting Morris--Lecar & 2 s & 20001 \\
        pBC & 6 s & 60001 \\
        \hline 
    \end{tabular}
\end{table}

To assess the robustness of the proposed method against measurement noise, voltage observations are contaminated with additive zero-mean Gaussian noise at two different levels. Specifically, the observed voltage signal is defined as
\begin{equation}
V^{\text{obs}}(t) = V^{\text{true}}(t) + r\varepsilon(t), \quad \varepsilon(t) \sim \mathcal{N}(0, \sigma_V^2),
\label{eq:stand_noise}
\end{equation}
where $r$ is the noise level, and $\sigma_V = \text{std}(V^{\text{true}}(t))$ represents the standard deviation of the true voltage signal. We consider $r = 1\%$ for the small-noise setting and $r = 5\%$ for the large-noise setting. For the pBC model, the large-noise level is increased to $r = 10\%$.
In addition, we examine a more realistic observational scenario by introducing additive absolute noise to the voltage measurements in the pBC model. 

\vspace{0.4em}
\noindent\textbf{Neural Network Design.} 
For the spiking Morris--Lecar (SML) model, we employ a two-layer fully connected architecture with hidden-layer widths $[50,\,50]$. For the bursting Morris--Lecar (BML) model and the pre-B\"{o}tzinger complex (pBC) model, deeper networks with three hidden layers of width $[50,\,50,\,50]$ are adopted to accommodate the higher dynamical complexity of these systems. All neural networks use the sigmoid function as the activation function. 

Due to the use of Fourier feature embedding, the effective input dimensionality of each network depends on the number of extracted FFT-based dominant frequencies and the trainable frequencies as discussed in Subsection \ref{subsubsec:ffe}. Table~\ref{tab:fourier_freq_summary} summarizes the number of dominant Fourier frequencies extracted for different examples. In the SML model, a relatively small threshold $p=95$ is sufficient, yielding only a few dominant frequencies for each regime, and the results remain unchanged as the noise level increases from $1\%$ to $5\%$. In contrast, the BML model requires a larger threshold $p=99$ due to its richer spectral content, resulting in significantly more dominant frequencies for both the square-wave and elliptic bursting regimes. These frequency counts also remain stable under moderate noise. For the pBC neuron model, an even larger threshold $p=99.92$ is necessary to accurately capture the dominant oscillatory components, reflecting the broadband spectral characteristics of the pBC dynamics. As the noise level increases, the number of extracted dominant frequencies shows a slight increase, indicating that stronger noise leads to a broader effective spectral support. 
\begin{table}[t]
\centering
\caption{Number of dominant Fourier frequencies extracted under different noise levels for the spiking Morris--Lecar (SML), bursting Morris--Lecar (BML), and pBC neuron models.}
\label{tab:fourier_freq_summary}
\resizebox{0.8\textwidth}{!}{
\begin{tabular}{c|c|c|c}
\hline
Model & Regime & Noise Level & FFT Freq. Count \\ \hline
\multirow{6}{*}{SML ($p=95$)} 
& Hopf        & $1\%$ & $4$ \\
& SNIC        & $1\%$ & $8$ \\
& HC  & $1\%$ & $3$ \\
& Hopf        & $5\%$ & $4$ \\
& SNIC        & $5\%$ & $8$ \\
& HC  & $5\%$ & $3$ \\ \hline
\multirow{4}{*}{BML ($p=99$)} 
& square-wave & $1\%$ & $19$ \\
& elliptic    & $1\%$ & $23$ \\
& square-wave & $5\%$ & $19$ \\
& elliptic    & $5\%$ & $23$ \\ \hline
\multirow{2}{*}{pBC ($p=99.92$)} 
& --- & $1\%$  & $49$ \\
& --- & $10\%$ & $53$ \\ \hline
\end{tabular}
}
\end{table}

As a result, the total number of trainable parameters (i.e., the network size) varies across different models, dynamical regimes, noise levels, and between the observed voltage variable $V$ and the unobserved state variables (gating variables $n$, $Ca$, and $h$). The detailed network sizes for all cases are summarized in Table~\ref{tab:network_size_summary}.

\begin{table}[t]
\centering
\caption{Total number of trainable parameters (network size) for the neural network of the observed voltage $V$ and of the unobserved variables $n$, $Ca$, and $h$ across different models, bifurcation regimes, and noise levels, respectively.}
\label{tab:network_size_summary}
\resizebox{0.9\textwidth}{!}{
\begin{tabular}{c|c|c|c|c|c|c}
\hline
Model & Regime & Noise & Network $V$ & Network $n$ & Network $Ca$ & Network $h$\\ \hline
\multirow{6}{*}{SML}
& Hopf        & $1\%$ & 3516  & 4237  & --- & --- \\
& SNIC        & $1\%$ & 4230  & 5672 & --- & --- \\
& HC  & $1\%$ & 3618  & 4442  & --- & --- \\
\cline{2-7}
& Hopf        & $5\%$ & 3516  & 4237  & --- & --- \\
& SNIC        & $5\%$ & 4332  & 5877 & --- & --- \\
& HC  & $5\%$ & 3720  & 4647  & --- & --- \\ 
\hline
\multirow{4}{*}{BML}
& square-wave & $1\%$ & 8972 & 12577 & 12577 & --- \\
& elliptic    & $1\%$ & 8666 & 11962 & 11962 & --- \\
& square-wave & $5\%$ & 8972 & 12577 & 12577 & --- \\
& elliptic    & $5\%$ & 8768 & 12167 & 12167 & --- \\ 
\hline
\multirow{2}{*}{pBC}
& ---         & $1\%$  & 10400 & 15447 & --- & 15447 \\
& ---         & $10\%$ & 10808 & 16267 & --- & 16267 \\ 
\hline
\end{tabular}}
\end{table}

\vspace{0.4em}
\noindent\textbf{Training Strategy.} 
All trainable parameters in the PINN framework, including the neural network parameters $\boldsymbol{\theta}$ (network weights and trainable Fourier frequencies) and the biophysical parameters $\boldsymbol{\lambda}$, are optimized using the Adam optimizer. During the data-driven pre-training stage, the voltage network $V$ for all three neuronal models is trained using the Adam optimizer with a staircase exponential decay learning rate schedule. The initial learning rate is set to $10^{-3}$ and is reduced by a factor of $0.5$ every $10{,}000$ iterations. The total number of training iterations in this stage is $20{,}000$. Mini-batch training is adopted at each iteration, with a batch size of $500$ for the spiking Morris--Lecar model and $1{,}000$ for the bursting Morris--Lecar model and the pBC model.

In the physics-informed training stage, the gating-variable networks and the biophysical parameters are jointly optimized under the physics loss imposed by the governing differential equations. In this stage, the residual losses associated with the dynamical system dominate the objective and enforce the physical consistency of both the recovered states and the inferred parameters. A summary of the training configurations used in the physics-informed training stage for all three neuronal models is provided in Table~\ref{tab:physics_stage_training}. The biophysical parameters $\boldsymbol{\lambda}$ are optimized using a constant learning rate, which is set to $10^{-4}$ for both the spiking and bursting Morris--Lecar models and to $10^{-3}$ for the pBC model.

\begin{table}[t]
\centering
\caption{Training configurations for the physics-constrained stage of different neuronal models. All gating-variable networks are optimized using the Adam optimizer with staircase exponential decay learning rates (lr), while the biophysical parameters $\boldsymbol{\lambda}$ are updated using constant learning rates as described in the text.}
\label{tab:physics_stage_training}
\resizebox{0.6\textwidth}{!}{
\begin{tabular}{c|c|c|c}
\hline
Model & Gating Networks & Initial lr & Total Iterations \\ \hline
SML 
& $n$ 
& $10^{-4}$ 
& $800{,}000$ \\ \hline
BML 
& $n,\, Ca$ 
& $10^{-4}$ 
& $1{,}000{,}000$ \\ \hline
pBC 
& $n,\, h$ 
& $10^{-3}$ 
& $200{,}000$ \\ \hline
\end{tabular}}
\end{table}

\vspace{0.4em}
\noindent\textbf{Error Metrics.}
To evaluate the performance of the proposed PINN framework, we consider both state reconstruction errors and parameter estimation errors as evaluation metrics.

\begin{itemize}
    \item \textbf{State reconstruction error.}
    The normalized $L^2$ error is used to measure the discrepancy between the predicted and reference state trajectories:
    \begin{equation}
        e_x = \frac{\|x - \hat{x}\|_2}{\|x\|_2}
        = \frac{\sqrt{\sum_{i=1}^{N}(x_i - \hat{x}_i)^2}}
        {\sqrt{\sum_{i=1}^{N}x_i^2}},
    \end{equation}
    where $x$ denotes the reference state trajectory with $N$ time sample points and $\hat{x}$ denotes the corresponding PINN prediction.
    
    \item \textbf{Parameter estimation error.}
    The relative error for each biophysical parameter $\lambda_k$ is defined as
    \begin{equation}
        e_{\lambda_k} = \frac{|\hat{\lambda}_k - \lambda_k|}{|\lambda_k|},
    \end{equation}
    where $\lambda_k$ and $\hat{\lambda}_k$ denote the ground-truth and the estimated parameter values, respectively.
\end{itemize}

\vspace{0.4em}
\noindent\textbf{Baseline Solutions.}
To assess the performance of the proposed PINN-based framework, we compare it with two widely used data assimilation techniques, the Unscented Kalman Filter (UKF) and 4D-Var \cite{moye2018data}. To this end, we apply all three methods - the proposed PINN-based framework, the UKF, and 4D-Var - to the same 0.2-second voltage trace (2,001 time points) in each dynamical regime, with identical noise levels to ensure a fair comparison. The UKF and 4D-Var implementations were adapted from \cite{moye2018data} to match our model setup and initial conditions. Since the available code in \cite{moye2018data} applies to the spiking Morris--Lecar model, comparative results are reported only for this model. 

We note that in \cite{moye2018data}, different observation lengths were used. Specifically, the UKF was implemented using a long trajectory of $20$ seconds (200,001 time points), whereas the 4D-Var method was restricted to a much shorter trajectory of 0.2 seconds (2,001 time points) due to computational and memory storage constraints. Here, we adopt the shorter 0.2-second observation window for all three methods to evaluate effectiveness and robustness under short observation windows.

\subsection{Application to the Spiking Morris-Lecar Model}
\label{subsec:Ex1}
The two-dimensional spiking Morris-Lecar (SML) Model \cite{morris1981voltage,ermentrout2010mathematical} is described by the following  equations:
\begin{align}
    C_m \frac{dV}{dt} &= I_{\text{app}} - g_L(V - E_L) - g_K n (V - E_K) - g_{Ca} m_{\infty}(V) (V - E_{Ca}) \\
    \frac{dn}{dt} &= \phi (n_{\infty}(V) - n)/\tau_n(V). 
\end{align}
with
\begin{align}
    m_{\infty} &= \frac{1}{2}(1 + \tanh((V - V_1)/V_2)) \\
    \tau_n &= 1/\cosh((V - V_3)/2V_4), \\
    n_{\infty} &= \frac{1}{2}(1 + \tanh((V - V_3)/V_4)).
\end{align}
We assume that $C_m = 20$, $E_{Ca} = 120$, $E_K = -84$, $E_L = -60$, and $I_{\text{app}} = 100$ are known parameters. The parameters to be estimated are $g_L$, $g_K$, $g_{Ca}$, $\phi$, $V_1$, $V_2$, $V_3$, and $V_4$. 

\begin{figure}[t]
    \centering
    \includegraphics[width=\linewidth]{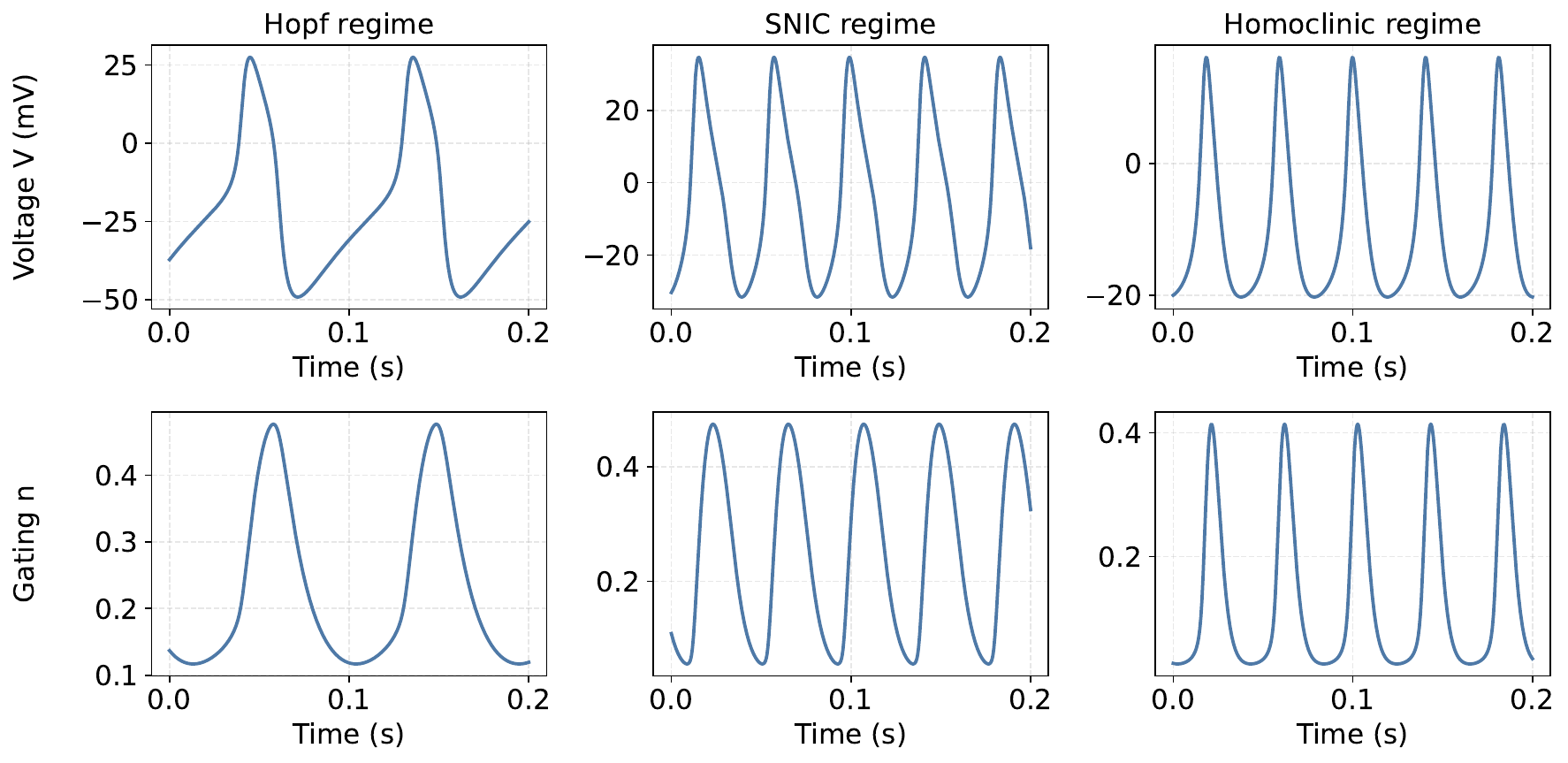}
    \caption{\small{Time trajectories of the solution of the spiking Morris–Lecar model under three distinct bifurcation regimes: Hopf (first column), SNIC (second column) and Homoclinic (third column). Each row corresponds to one state variable: voltage $V$, gating variable $n$.}}
    \label{fig:SR_ML_true_bifurcation}
\end{figure}

The SML model exhibits multiple excitability mechanisms, characterized by distinct bifurcation structures through which stable oscillations emerge. These include the Hopf, saddle-node on an invariant circle (SNIC), and homoclinic (HC) bifurcations, with the corresponding parameter values (ground-true values to be estimated) listed in Table \ref{tab:SR_ML_true_param}. The true trajectories for $V(t)$ and $n(t)$ for these three important bifurcation regimes are shown in Figure \ref{fig:SR_ML_true_bifurcation}, computed over a total simulation time of $T=0.2$ s. 

\begin{table}[t]
\centering
\footnotesize
\caption{\small{The true parameter values of Morris-Lecar model for three different bifurcation regimes, Hope, saddle-node on an invariant circle (SNIC), and Homoclinic.} }
\label{tab:SR_ML_true_param}
\begin{tabular}{c|c|c|c}
\hline
&  Hopf  &  SNIC & Homoclinic \\
\hline 
$\phi$ & 0.04 & 0.067 & 0.23 \\
gCa & 4 & 4 & 4 \\
$V_3$ & 2 & 12 & 12 \\
$V_4$ & 30 & 17.4 & 17.4 \\
gK & 8 & 8 & 8 \\
gL & 2 & 2 & 2 \\
$V_1$ & -1.2 & -1.2 & -1.2 \\
$V_2$ & 18 & 18 & 18 \\
\hline
\end{tabular}
\end{table} 

All biological parameters were bounded by the ranges specified in Table \ref{tab:SR_ML_param_bound} to ensure biological plausibility. These sign constraints are enforced through the following reparameterization:
\begin{equation}
\lambda_j =
\begin{cases}
e^{z_j}, & \text{if }\lambda_j > 0, \\
-e^{z_j}, & \text{if }\lambda_j < 0,
\end{cases}
\end{equation}
where each $z_j \in \mathbb{R}$ is an unconstrained trainable variable.

\begin{table}[t]
\centering
\footnotesize
\caption{\small{Bounds  for the Spiking Morris-Lecar Model in the training stage of PINNs.} }
\label{tab:SR_ML_param_bound}
\begin{tabular}{c|c|c}
\hline
&  Lower bound  &  Upper bound  \\
\hline 
$\phi$ & 0 & $+\infty$ \\
gCa & 0 & $+\infty$ \\
$V_3$ & 0 & $+\infty$ \\
$V_4$ & 0 & $+\infty$ \\
gK & 0 & $+\infty$ \\
gL & 0 & $+\infty$ \\
$V_1$ & $-\infty$ & 0 \\
$V_2$ & 0 & $+\infty$ \\
\hline
\end{tabular}
\end{table} 

Below, we apply our PINN framework to estimate model parameters from partially observed data generated by the SML model across multiple bifurcation regimes (i.e., using observations of $V$ over a short observation window) and contaminated with varying levels of noise. For each bifurcation regime, the algorithm is initialized using parameter guesses corresponding to a different regime \cite{WANG2021113938} as well as a non-informative initialization. We evaluate performance by comparing the estimated parameters with the corresponding ground-truth values. Notably, parameter estimation in each case is performed using data obtained at a fixed value of the applied current $I_{\rm app}$. To assess whether the estimated parameter sets recover the correct qualitative dynamics, we also numerically compute the resulting bifurcation diagram of the SML model as $I_{\rm app}$ varies based on the estimated parameters and compare it with the true underlying bifurcation structure. 

\subsubsection{Hopf regime}
For the Hopf regime, the PINN is evaluated under two noise levels ($1\%$ and $5\%$) using initial guesses from the SNIC regime, the Homoclinic regime, and a non-informative initialization in which all parameters are initialized to be $1$.

{\bf Parameter estimation}. 
Table \ref{tab:combined_results_SR_ML_hopf} summarizes the parameter estimation results for the Hopf regime when the initial guesses are taken from the SNIC and Homoclinic regimes, as well as from a non-informative initialization. All three methods, PINN, UKF, and 4D-Var, are evaluated using the same set of observation data under different measurement noise levels. It is important to note that, when using initial guesses from the SNIC or Homoclinic regimes, the corresponding parameter sets differ from the Hopf regime only in three parameters ($\phi$, $V_3$, and $V_4$), while all remaining parameters are the same (see Table \ref{tab:SR_ML_true_param}).

When the initial guess is taken from the SNIC regime, the PINN method demonstrates a clear advantage over both UKF and 4D-Var. As shown in Table~\ref{tab:hopf_est_1} at the $1\%$ noise level, the PINN consistently achieves accurate parameter estimates, with the largest relative error among estimated parameters being only $2.9\%$ (for $V_1$). In contrast, 4D-Var performs noticeably better than UKF but still shows significant error for parameter $V_3$. The UKF exhibits the poorest performance, with the relative error for $V_3$, indicating a strong sensitivity to initial guesses even under small noise.

A similar performance trend holds when the initial guess is taken from the Homoclinic regime. Again, the PINN maintains stable and accurate performance, with all relative errors remaining below $2.9\%$. The 4D-Var method shows its largest error for $V_3$  at $75.1\%$, while the UKF yields substantially larger estimation errors across multiple parameters, particularly for $V_3$. We emphasize that these results are obtained using a short observation window of 0.2 seconds. For filtering-based methods such as the UKF, such a limited trajectory length may be insufficient for state and parameter convergence, particularly when the initial guess lies in a distant dynamical regime. As demonstrated in \cite{moye2018data}, increasing the observation window could substantially improve UKF performance.

We further test a more challenging scenario using a non-informative initial guess, where all parameters are initialized to $1$. As shown in Table \ref{tab:combined_results_SR_ML_hopf}, this experiment highlights a fundamental limitation of both UKF and 4D-Var when used as baseline methods in this initialization setting.
Specifically, both approaches rely on classical numerical solvers to march the underlying ODE system in time, making their performance sensitive to the initial guesses of the model parameters. In our experimental setup, non-informative initializations correspond to parameter values that lie far from the true dynamical regime. As a result, the ODE solver could fail to generate physically meaningful forward trajectories, leading to divergence of both UKF and 4D-Var methods. In contrast, the PINN framework does not depend on explicit time-marching of the ODE with fixed parameter values. Instead, state trajectories and parameters are inferred simultaneously through a unified optimization process, which could reduces sensitivity to poor initial guesses of the biophysical parameters. Consequently, the PINN consistently converges to accurate parameter estimates with both regime-based initial guesses and non-informative initial guesses.

Finally, at a large noise level of $5\%$, the PINN method maintains strong robustness. 
As shown in Table~\ref{tab:hopf_est_5}, the PINN accurately estimates all parameters within $5\%$ relative error except $V_1$ and $V_3$, whose relative errors are below $10.7\%$ and $14.2\%$, respectively. In contrast, both UKF and 4D-Var exhibit noticeably larger estimation errors. The UKF performs poorly for both SNIC- and Homoclinic-based initializations, with multiple parameters (e.g., $\phi, V_1, V_3, V_4$) exhibiting relative errors above $40\%$. While 4D-Var performs better than the UKF, achieving relative errors below $10\%$ for most parameters, it yields a significant error in $V_3$, with relative errors exceeding $50\%$ under both initial guesses.

\begin{table}[t]
    \centering   
    \caption{\small{Estimated biophysical parameters for the Hopf regime under two observational noise levels ($1\%$ and $5\%$), evaluated using two initial guess strategies (regime-based initial guess and non-informative initial guess where all parameters are set to $1$), with observed data from bifurcation regime `t' and initial parameter guesses corresponding to bifurcation regime `g'.}}
    \label{tab:combined_results_SR_ML_hopf}
    \begin{subtable}{\textwidth}
        \centering
        \caption{HOPF ($1\%$ noise)}
        \resizebox{\textwidth}{!}{
        \begin{tabular}{c|ccc|cc|cc}
        \hline
        t: Hopf & \multicolumn{3}{c}{PINN} & \multicolumn{2}{c}{UKF} & \multicolumn{2}{c}{4D-Var}\\
        \hline
         & g: SNIC & g: Homoclinic & g: 1 & g: SNIC & g: Homoclinic & g: SNIC & g: Homoclinic\\
         \hline
          & Est. (Rel. Err.) & Est. (Rel. Err.) & Est. (Rel. Err.) & Est. (Rel. Err.) & Est. (Rel. Err.) & Est. (Rel. Err.) & Est. (Rel. Err.) \\
        \hline
        $g_{\rm L}$ & $1.974 (1.3\%)$ & $1.973 (1.3\%)$ & $1.973 (1.3\%)$ & $2.516 (25.8\%)$ & $2.380 (19.0\%)$ & $1.983 (0.8\%)$ & $1.986 (0.7\%)$ \\
        $g_{\rm K}$ & $8.019 (0.2\%)$ & $8.019 (0.2\%)$ & $8.021 (0.3\%)$ & $8.764 (9.5\%)$ & $8.606 (7.6\%)$ & $8.283 (3.5\%)$ & $8.366 (4.6\%)$ \\
        $g_{\rm Ca}$ & $3.983 (0.4\%)$ & $3.982 (0.4\%)$ & $3.982 (0.4\%)$ & $3.722 (7.0\%)$ & $3.701 (7.5\%)$ & $3.931 (1.7\%)$ & $3.952 (1.2\%)$ \\
        $\phi$ & $0.040 (0.6\%)$ & $0.040 (0.6\%)$ & $0.040 (0.6\%)$ & $0.018 (56.0\%)$ & $0.013 (66.5\%)$ & $0.039 (3.6\%)$ & $0.039 (2.7\%)$ \\
        $V_1$ & $-1.165 (2.9\%)$ & $-1.165 (2.9\%)$ & $-1.165 (2.9\%)$ & $-1.180 (1.7\%)$ & $-1.152 (4.0\%)$ & $-1.220 (1.7\%)$ & $-1.196 (0.3\%)$ \\
        $V_2$ & $17.991 (<0.1\%)$ & $17.991 (0.1\%)$ & $17.991 (0.1\%)$ & $17.801 (1.1\%)$ & $17.847 (0.9\%)$ & $17.823 (1.0\%)$ & $17.896 (0.6\%)$ \\
        $V_3$ & $1.963 (1.8\%)$ & $1.962 (1.9\%)$ & $1.972 (1.4\%)$ & $11.746 (487.3\%)$ & $11.799 (489.9\%)$ & $3.076 (53.8\%)$ & $3.501 (75.1\%)$ \\
        $V_4$ & $29.983 (0.1\%)$ & $29.983 (0.1\%)$ & $29.992 (<0.1\%)$ & $17.503 (41.7\%)$ & $17.439 (41.9\%)$ & $29.896 (0.3\%)$ & $30.362 (1.2\%)$ \\
        \hline
        \end{tabular}
        }
        \label{tab:hopf_est_1}
    \end{subtable}
    
    \begin{subtable}[h]{\textwidth}
        \centering
        \caption{HOPF ($5\%$ noise)}
        \resizebox{\textwidth}{!}{
        \begin{tabular}{c|ccc|cc|cc}
        \hline
        t: Hopf & \multicolumn{3}{c}{PINN} & \multicolumn{2}{c}{UKF} & \multicolumn{2}{c}{4D-Var}\\
        \hline
         & g: SNIC & g: Homoclinic & g: 1 & g: SNIC & g: Homoclinic & g: SNIC & g: Homoclinic\\
         \hline
          & Est. (Rel. Err.) & Est. (Rel. Err.) & Est. (Rel. Err.) & Est. (Rel. Err.) & Est. (Rel. Err.) & Est. (Rel. Err.) & Est. (Rel. Err.) \\
        \hline
        $g_{\rm L}$ & $1.935 (3.3\%)$ & $1.934 (3.3\%)$ & $1.933 (3.3\%)$ & $2.464 (23.2\%)$ & $2.383 (19.2\%)$ & $1.858 (7.1\%)$ & $1.857 (7.1\%)$ \\
        $g_{\rm K}$ & $8.027 (0.3\%)$ & $8.028 (0.4\%)$ & $8.033 (0.4\%)$ & $8.670 (8.4\%)$ & $8.525 (6.6\%)$ & $8.516 (6.5\%)$ & $8.573 (7.2\%)$ \\
        $g_{\rm Ca}$ & $3.915 (2.1\%)$ & $3.915 (2.1\%)$ & $3.914 (2.2\%)$ & $3.636 (9.1\%)$ & $3.653 (8.7\%)$ & $3.818 (4.5\%)$ & $3.824 (4.4\%)$ \\
        $\phi$ & $0.038 (4.2\%)$ & $0.038 (4.2\%)$ & $0.038 (4.2\%)$ & $0.018 (56.0\%)$ & $0.015 (61.5\%)$ & $0.037 (7.0\%)$ & $0.037 (6.7\%)$ \\
        $V_1$ & $-1.071 (10.7\%)$ & $-1.071 (10.7\%)$ & $-1.072 (10.7\%)$ & $-1.111 (7.4\%)$ & $-1.096 (8.7\%)$ & $-1.170 (2.5\%)$ & $-1.161 (3.3\%)$ \\
        $V_2$ & $17.972 (0.2\%)$ & $17.972 (0.2\%)$ & $17.969 (0.2\%)$ & $17.881 (0.7\%)$ & $17.906 (0.5\%)$ & $17.638 (2.0\%)$ & $17.663 (1.9\%)$ \\
        $V_3$ & $1.717 (14.2\%)$ & $1.720 (14.0\%)$ & $1.738 (13.1\%)$ & $11.759 (488.0\%)$ & $11.805 (490.2\%)$ & $3.476 (73.8\%)$ & $3.737 (86.8\%)$ \\
        $V_4$ & $29.387 (2.0\%)$ & $29.388 (2.0\%)$ & $29.402 (2.0\%)$ & $17.540 (41.5\%)$ & $17.471 (41.8\%)$ & $29.980 (0.1\%)$ & $30.222 (0.7\%)$ \\
        \hline
        \end{tabular}
        }
        \label{tab:hopf_est_5}
    \end{subtable}  
\end{table}

\begin{figure}[t]
\centering
\begin{minipage}{\textwidth} 
    \centering
    \begin{subfigure}[b]{\textwidth}
        \centering
        \includegraphics[width=\textwidth]{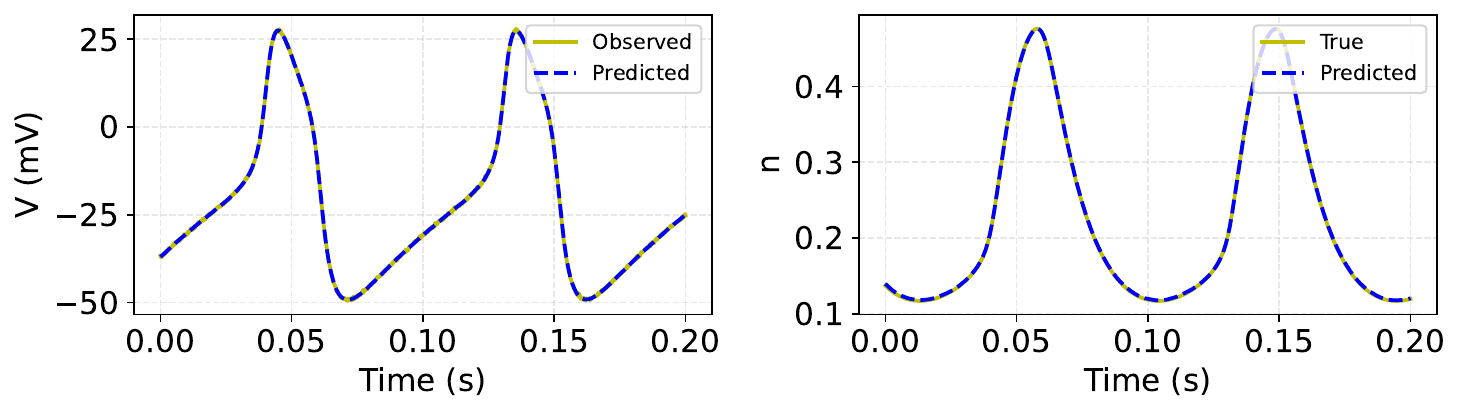}
        \caption{Small Noise Level: True: Hopf; Initial: 1}
    \end{subfigure}
    \hfill
    \begin{subfigure}[b]{\textwidth}
        \centering
        \includegraphics[width=\textwidth]{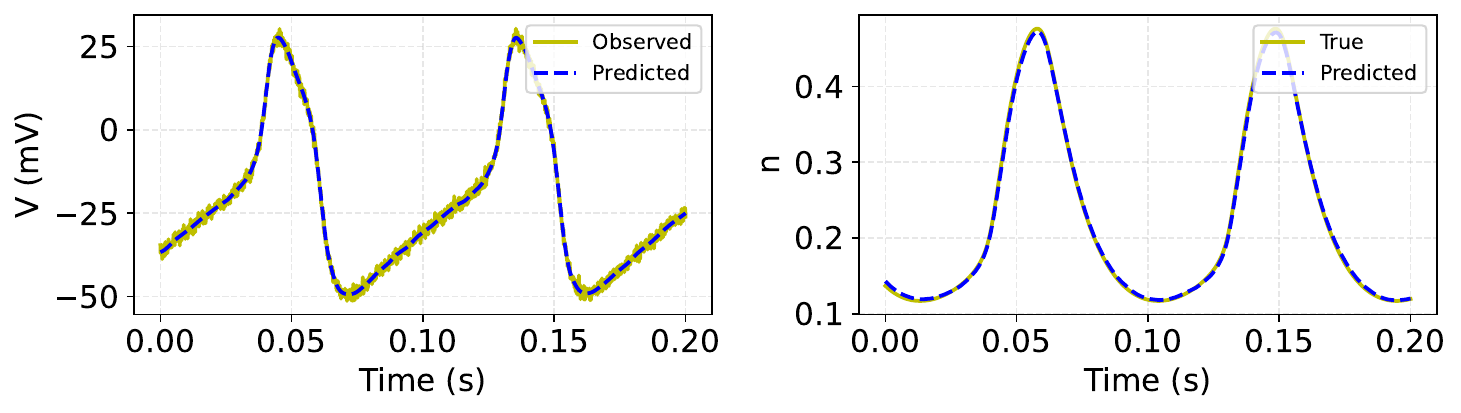}
        \caption{Large Noise Level: True: Hopf; Initial: 1}
    \end{subfigure} 
    \caption{\small{Comparison between the ground-truth trajectories and the PINN-reconstructed trajectories for the Hopf regime under two noise levels based on the non-informative initial guess. PINN predictions are shown as dashed blue lines, while the observation voltage $V$ and ground-truth gating $n$ trajectories are shown as solid yellow lines.}}
    \label{fig:SR_ML_tHOPF_Prediction}
\end{minipage}
\end{figure}

{\bf Forward trajectory reconstruction}. 
Next, we demonstrate that the PINN framework not only achieves accurate parameter estimation, but also reliably reconstructs the full system state trajectory, including unobserved variables. In our experimental setup for the SML model, only the membrane voltage $V(t)$ is directly measured, while the gating variable $n(t)$ remains completely unobserved. Despite this, the PINN successfully recovers both variables. As shown in Figure~\ref{fig:SR_ML_tHOPF_Prediction}, the reconstructed solutions for $V(t)$ and $n(t)$  closely match the ground truth trajectories even when initialized from a non-informative starting point.

Quantitatively, Table~\ref{tab:SR_ML_tHOPF_Prediction} reports the state reconstruction errors for both variables under three initial guesses (SNIC, Homoclinic, and non-informative). At a low noise level, reconstruction errors remain below 0.5\%. When the noise is increased to 5\%, the observation data become visibly noisier, yet the PINN-reconstructed trajectories remain smooth and dynamically consistent. As a result, reconstruction errors increase only slightly, staying below 1.3\% for both variables. Importantly, even under higher noise, the PINN continues to accurately capture the qualitative spiking dynamics and phase relationships.
 
\begin{table}[t]
    \centering
    \caption{\small{Comparison between the ground-truth trajectories and the PINN-reconstructed
solutions for the SNIC regime under two noise levels using state reconstruction errors.}}
    \label{tab:SR_ML_tHOPF_Prediction}
    \resizebox{0.6\textwidth}{!}{
    \begin{tabular}{c|c|c|c|c}
        \hline
        \textbf{Noise Level} & \textbf{True Regime} & \textbf{Initial Guess} & \textbf{$e_{\hat{V}}$} & \textbf{$e_{\hat{n}}$} \\
        \hline
         $1\%$ & Hopf & SNIC & $0.15\%$ & $0.31\%$  \\
         $1\%$ & Hopf & HC & $0.15\%$ & $0.32\%$  \\
         $1\%$ & Hopf & $1$ & $0.15\%$ & $0.32\%$  \\
        \hline
         $5\%$ & Hopf & SNIC & $0.62\%$ & $1.21\%$  \\
         $5\%$ & Hopf & HC & $0.62\%$ & $1.21\%$  \\
         $5\%$ & Hopf & $1$ & $0.62\%$ & $1.25\%$  \\
        \hline
    \end{tabular}}
\end{table}

\subsubsection{SNIC regime}
For the SNIC regime, the PINN is evaluated under two noise levels ($1\%$ and $5\%$) using initial guesses from the Hopf regime, the Homoclinic regime, and a non-informative initialization in which all parameters are set to $1$.

{\bf Parameter estimation.} 
Table~\ref{tab:combined_results_SR_ML_snic} summarizes the parameter estimation results for the SNIC regime under different initial guesses and noise levels. The PINN is evaluated using initial guesses from the Hopf regime, the Homoclinic regime, and a non-informative with all parameters initialized to $1$. For UKF and 4D-Var,   since non-informative initial guess leads to divergence or invalid forward trajectories, only regime-based initial guesses are considered.
Under the small-noise level shown in Table~\ref{tab:snic_est_1}, the PINN consistently achieves accurate parameter estimation regardless of the initial guess. When initialized from the Hopf, Homoclinic, or non-informative regimes, all parameter errors remain below $6\%$, with most below $2\%$. This demonstrates that PINN's performance is largely insensitive to the choice of initial guess.

The performance of UKF and 4D-Var, however, is more dependent on the starting regime. When the initial guess is taken from the Homoclinic regime, both methods perform well---likely due to the parameter similarity between the SNIC and Homoclinic regimes, which differ only in the value of $\phi$. Here, UKF achieves a maximum error of $1.1\%$, and 4D-Var keeps all errors below $3\%$. In contrast, when initialized from the more challenging Hopf regime, UKF performance deteriorates substantially, with large errors for $V_3$ ($81.2\%$) and $V_4$ ($70.9\%$). 4D-Var remains more robust, with all errors below $5\%$. Even in this scenario, the PINN continues to deliver stable estimates with errors consistently below $6\%$, highlighting its robustness to regime mismatch.

Under the large-noise level, PINN maintains its accuracy across all initial guesses, with errors still below approximately $6\%$ and most below $2\%$, confirming its robustness to both increased noise and initial-guess variation as shown Table~\ref{tab:snic_est_5}. In contrast, UKF exhibits strong sensitivity to both factors. While it yields reasonable estimates when initialized from the Homoclinic regime, using the Hopf regime leads to errors exceeding $80\%$ for $V_3$ and approximately $70\%$ for $V_4$. For the SNIC regime, 4D-Var achieves highly accurate parameter estimation under both noise levels when initialized from appropriate regime-based guesses. As shown in Table~\ref{tab:combined_results_SR_ML_snic}, the maximum relative error remains below approximately $8.1\%$, with most parameters estimated within a few percent error.  However, this accuracy relies on the availability of informative initial guesses. In contrast, the PINN does not require regime-specific initializations and is able to consistently recover accurate parameter estimates even from non-informative initial guesses.

\begin{table}[t]
    \centering   
    \caption{\small{Estimated biophysical parameters for the SNIC regime under two observational noise levels ($1\%$ and $5\%$), evaluated using two initial guess strategies (regime-based initial guess and non-informative initial guess where all parameters are set to $1$), with observed data from bifurcation regime `t' and initial parameter guesses corresponding to bifurcation regime `g'.}}
    \label{tab:combined_results_SR_ML_snic}
    \begin{subtable}{\textwidth}
        \centering
        \caption{SNIC ($1\%$ noise)}
        \resizebox{\textwidth}{!}{
        \begin{tabular}{c|ccc|cc|cc}
        \hline
        t: SNIC & \multicolumn{3}{c}{PINN} & \multicolumn{2}{c}{UKF} & \multicolumn{2}{c}{4D-Var}\\
        \hline
         & g: Hopf & g: Homoclinic & g: 1 & g: Hopf & g: Homoclinic & g: Hopf & g: Homoclinic\\
         \hline
          & Est. (Rel. Err.) & Est. (Rel. Err.) & Est. (Rel. Err.) & Est. (Rel. Err.) & Est. (Rel. Err.) & Est. (Rel. Err.) & Est. (Rel. Err.) \\
        \hline
        $g_{\rm L}$ & $1.996 (0.2\%)$ & $1.996 (0.2\%)$ & $1.995 (0.2\%)$ & $0.735 (63.2\%)$ & $2.014 (0.7\%)$ & $1.998 (0.1\%)$ & $2.042 (2.1\%)$ \\
        $g_{\rm K}$ & $8.057 (0.7\%)$ & $8.056 (0.7\%)$ & $8.058 (0.7\%)$ & $7.386 (7.7\%)$ & $7.987 (0.2\%)$ & $8.154 (1.9\%)$ & $7.971 (0.4\%)$ \\
        $g_{\rm Ca}$ & $4.035 (0.9\%)$ & $4.035 (0.9\%)$ & $4.034 (0.9\%)$ & $4.305 (7.6\%)$ & $4.016 (0.4\%)$ & $3.986 (0.3\%)$ & $4.015 (0.4\%)$ \\
        $\phi$ & $0.068 (2.0\%)$ & $0.068 (2.0\%)$ & $0.068 (2.0\%)$ & $0.115 (71.2\%)$ & $0.068 (1.1\%)$ & $0.066 (1.5\%)$ & $0.066 (1.2\%)$ \\
        $V_1$ & $-1.129 (5.9\%)$ & $-1.129 (5.9\%)$ & $-1.130 (5.9\%)$ & $-0.901 (24.9\%)$ & $-1.189 (0.9\%)$ & $-1.225 (2.1\%)$ & $-1.213 (1.1\%)$ \\
        $V_2$ & $18.105 (0.6\%)$ & $18.105 (0.6\%)$ & $18.103 (0.6\%)$ & $18.455 (2.5\%)$ & $17.983 (0.1\%)$ & $17.942 (0.3\%)$ & $18.008 (<0.1\%)$ \\
        $V_3$ & $12.166 (1.4\%)$ & $12.166 (1.4\%)$ & $12.169 (1.4\%)$ & $2.257 (81.2\%)$ & $12.001 (<0.1\%)$ & $12.393 (3.3\%)$ & $11.973 (0.2\%)$ \\
        $V_4$ & $17.855 (2.6\%)$ & $17.851 (2.6\%)$ & $17.854 (2.6\%)$ & $29.731 (70.9\%)$ & $17.399 (<0.1\%)$ & $17.351 (0.3\%)$ & $16.986 (2.4\%)$ \\
        \hline
        \end{tabular}
        }
        \label{tab:snic_est_1}
    \end{subtable}
    
    \begin{subtable}[h]{\textwidth}
        \centering
        \caption{SNIC ($5\%$ noise)}
        \resizebox{\textwidth}{!}{
        \begin{tabular}{c|ccc|cc|cc}
        \hline
        t: SNIC & \multicolumn{3}{c}{PINN} & \multicolumn{2}{c}{UKF} & \multicolumn{2}{c}{4D-Var}\\
        \hline
         & g: Hopf & g: Homoclinic & g: 1 & g: Hopf & g: Homoclinic & g: Hopf & g: Homoclinic\\
         \hline
          & Est. (Rel. Err.) & Est. (Rel. Err.) & Est. (Rel. Err.) & Est. (Rel. Err.) & Est. (Rel. Err.) & Est. (Rel. Err.) & Est. (Rel. Err.) \\
        \hline
        $g_{\rm L}$ & $2.115 (5.8\%)$ & $2.118 (5.9\%)$ & $2.105 (5.3\%)$ & $0.710 (64.5\%)$ & $2.016 (0.8\%)$ & $2.037 (1.8\%)$ & $2.046 (2.3\%)$ \\
        $g_{\rm K}$ & $7.927 (0.9\%)$ & $7.922 (1.0\%)$ & $7.948 (0.6\%)$ & $7.325 (8.4\%)$ & $7.989 (0.1\%)$ & $8.086 (1.1\%)$ & $8.018 (0.2\%)$ \\
        $g_{\rm Ca}$ & $4.076 (1.9\%)$ & $4.077 (1.9\%)$ & $4.071 (1.8\%)$ & $4.117 (2.9\%)$ & $4.020 (0.5\%)$ & $4.045 (1.1\%)$ & $4.065 (1.6\%)$ \\
        $\phi$ & $0.066 (1.9\%)$ & $0.066 (1.9\%)$ & $0.066 (1.8\%)$ & $0.109 (63.0\%)$ & $0.067 (0.2\%)$ & $0.068 (1.2\%)$ & $0.069 (2.3\%)$ \\
        $V_1$ & $-1.257 (4.8\%)$ & $-1.257 (4.8\%)$ & $-1.260 (5.0\%)$ & $-1.162 (3.1\%)$ & $-1.185 (1.3\%)$ & $-1.137 (5.3\%)$ & $-1.103 (8.1\%)$ \\
        $V_2$ & $18.062 (0.3\%)$ & $18.064 (0.4\%)$ & $18.048 (0.3\%)$ & $18.404 (2.2\%)$ & $17.966 (0.2\%)$ & $17.920 (0.4\%)$ & $17.968 (0.2\%)$ \\
        $V_3$ & $11.890 (0.9\%)$ & $11.883 (1.0\%)$ & $11.930 (0.6\%)$ & $2.209 (81.6\%)$ & $12.003 (<0.1\%)$ & $12.359 (3.0\%)$ & $12.209 (1.7\%)$ \\
        $V_4$ & $16.509 (5.1\%)$ & $16.489 (5.2\%)$ & $16.581 (4.7\%)$ & $29.662 (70.5\%)$ & $17.402 (<0.1\%)$ & $17.422 (0.1\%)$ & $17.494 (0.5\%)$ \\
        \hline
        \end{tabular}
        }
        \label{tab:snic_est_5}
    \end{subtable}
\end{table}

\begin{figure}[t]
\begin{minipage}{\textwidth}  
    \centering
    \begin{subfigure}[b]{\textwidth}
        \centering
        \includegraphics[width=\textwidth]{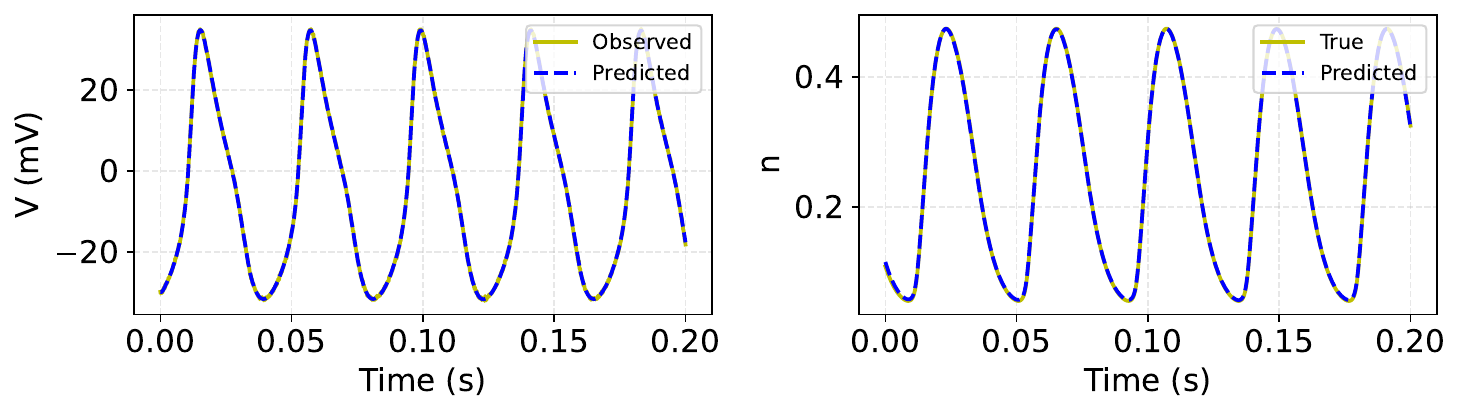}
        \caption{Small Noise: True: SNIC; Initial: 1}
    \end{subfigure}
    \hfill
    \begin{subfigure}[b]{\textwidth}
        \centering
        \includegraphics[width=\textwidth]{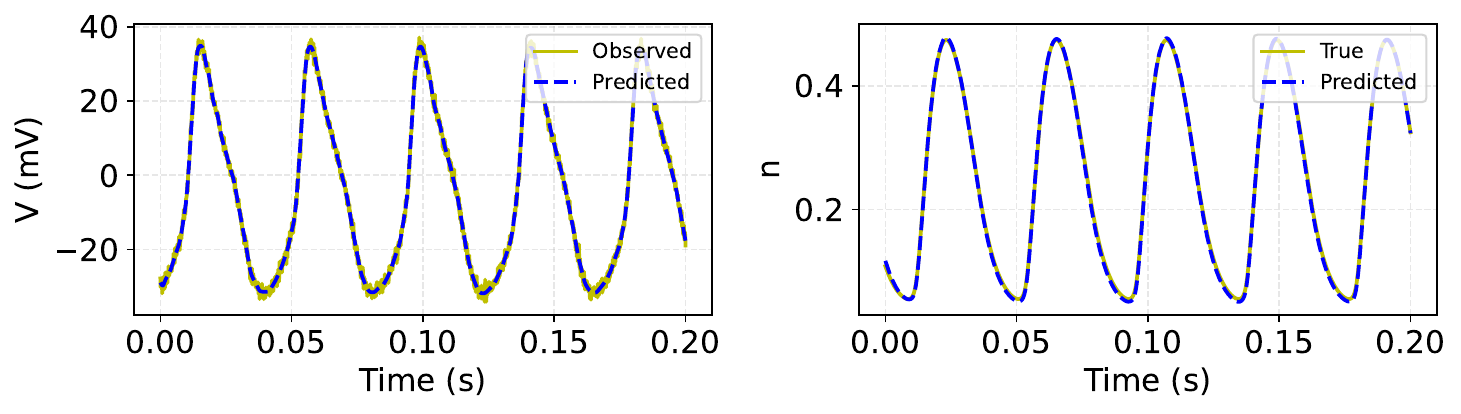}
        \caption{Large Noise: True: SNIC; Initial: 1}
    \end{subfigure}
    \caption{\small{Comparison between the ground-truth trajectories and the PINN-reconstructed solutions for the SNIC regime under two noise levels based on the non-informative initial guess, where the PINN predictions are shown as dashed blue lines and the observation voltage $V$ and ground-truth gating $n$ trajectories are shown as solid yellow lines.}}
    \label{fig:SR_ML_tSNIC_Prediction}
\end{minipage}
\end{figure}

{\bf Forward trajectory reconstruction.} 
Figure~\ref{fig:SR_ML_tSNIC_Prediction} and Table~\ref{tab:SR_ML_tSNIC_Prediction} illustrate the state reconstruction performance of the PINN for the SNIC regime across both small- and large-noise levels. Under the small-noise level, the PINN achieves highly accurate state reconstruction. The relative $L^2$ error for the observed voltage is approximately $0.27\%$, while the error for the unobserved gating variable $n(t)$ remains around $0.5\%$---both well below $0.5\%$. When the noise level is increased, the reconstruction accuracy decreases only modestly. 
Despite increased noise level, both errors of the voltage and the the gating variable $n$ remain below $1.5\%$, and the PINN is able to successfully capture the qualitative spiking dynamics while accurately reconstructing the states of both observed and unobserved variables.

\begin{table}[t]
    \centering
    \caption{\small{Comparison between the ground-truth trajectories and the PINN-reconstructed solutions for the SNIC regime under two noise levels using state reconstruction errors.}}
    \label{tab:SR_ML_tSNIC_Prediction}
    \resizebox{0.6\textwidth}{!}{
    \begin{tabular}{c|c|c|c|c}
        \hline
        \textbf{Noise Level} & \textbf{True Regime} & \textbf{Initial Guess} & \textbf{$e_{\hat{V}}$} & \textbf{$e_{\hat{n}}$} \\
        \hline
         $1\%$ & SNIC & Hopf & $0.27\%$ & $0.49\%$  \\
         $1\%$ & SNIC & HC & $0.27\%$ & $0.48\%$  \\
         $1\%$ & SNIC & $1$ & $0.27\%$ & $0.49\%$  \\
        \hline
         $5\%$ & SNIC & Hopf & $1.33\%$ & $1.24\%$  \\
         $5\%$ & SNIC & HC & $1.33\%$ & $1.26\%$  \\
         $5\%$ & SNIC & $1$ & $1.33\%$ & $1.18\%$  \\
        \hline
    \end{tabular}}
\end{table}

\subsubsection{Homoclinic regime}
For the Homoclinic regime, the PINN is evaluated under two noise levels ($1\%$ and $5\%$) using initial guesses from the Hopf regime, the SNIC regime, and a non-informative initialization in which all parameters are set to $1$.

{\bf Parameter estimation.}
Table~\ref{tab:combined_results_SR_ML_homo} summarizes the parameter estimation results for the Homoclinic regime under different initial guesses. Under the small-noise level ($1\%$, Table~\ref{tab:homoclinic_est_1}), the PINN demonstrates consistently accurate estimation across all initial guesses, with all relative errors remaining below approximately $1.5\%$.

The performance of UKF and 4D-Var, however, depends strongly on the similarity between the initial guess and the true dynamical regime. When initialized from the nearby SNIC regime, which differs from the Homoclinic regime in only a small subset of parameters (primarily $V_1$, $V_3$, and $V_4$), both methods perform well. In this case, the UKF yields relative errors below $1\%$ for all parameters, and 4D-Var also deliver reasonably accurate estimates for most of the parameters with the maximum relative error  $15.4\%$ (for $V_3$). In contrast, when initialized from the more distant Hopf regime, where multiple key parameters differ substantially from the Homoclinic regime, both methods degrade markedly even under low noise. The UKF exhibits large estimation errors exceeding $60\%$ for several parameters, including $g_L$, $V_1$, $V_3$, and $V_4$. The 4D-Var method performs even more poorly, with relative errors exceeding $100\%$ for parameters such as $g_{Ca}$ and $V_1$. These results indicate a pronounced sensitivity to regime mismatch in the initial guess, even at the $1\%$ noise level.

Under the higher noise level ($5\%$, Table~\ref{tab:homoclinic_est_5}), these differences in robustness become more pronounced. The PINN maintains stable performance across all initial guesses, with errors increasing only moderately compared to the low-noise case and remaining below approximately $5\%$. In contrast, the UKF again shows strong dependence on initialization. It performs reasonably when started from the SNIC regime but deteriorates significantly under Hopf initialization, with several parameters exhibiting relative errors exceeding $60\%$. The 4D-Var method is further affected by noise; while it remains acceptable under SNIC initialization, its performance under Hopf initialization degrades severely, particularly for  $V_1$.

\begin{figure}[h!]
\begin{minipage}{\textwidth}
    \centering
    \begin{subfigure}[b]{\textwidth}
        \centering
        \includegraphics[width=\textwidth]{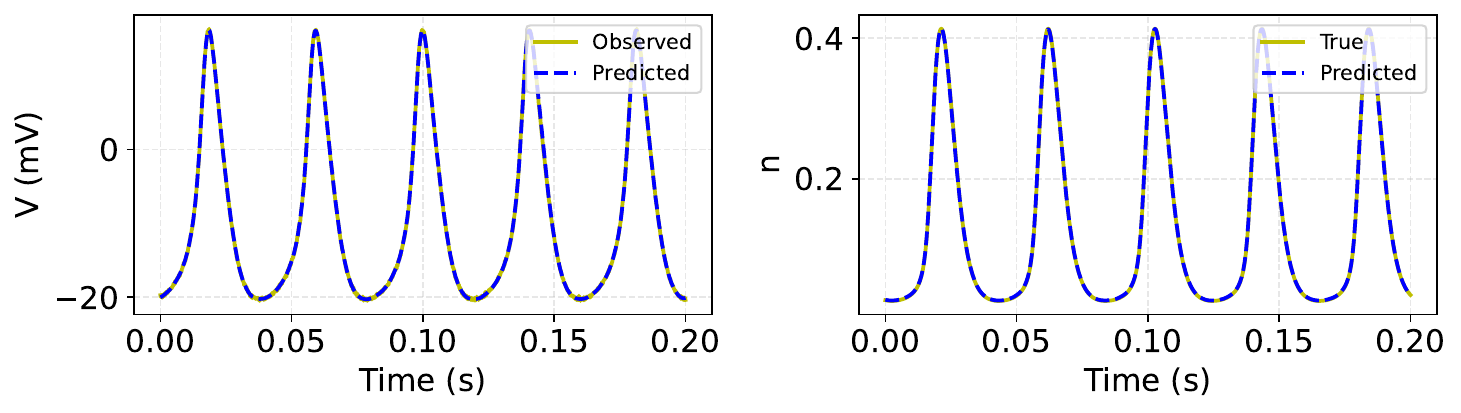}
        \caption{Small Noise: True: Homoclinic; Initial: 1}
    \end{subfigure}
    \hfill
    \begin{subfigure}[b]{\textwidth}
        \centering
        \includegraphics[width=\textwidth]{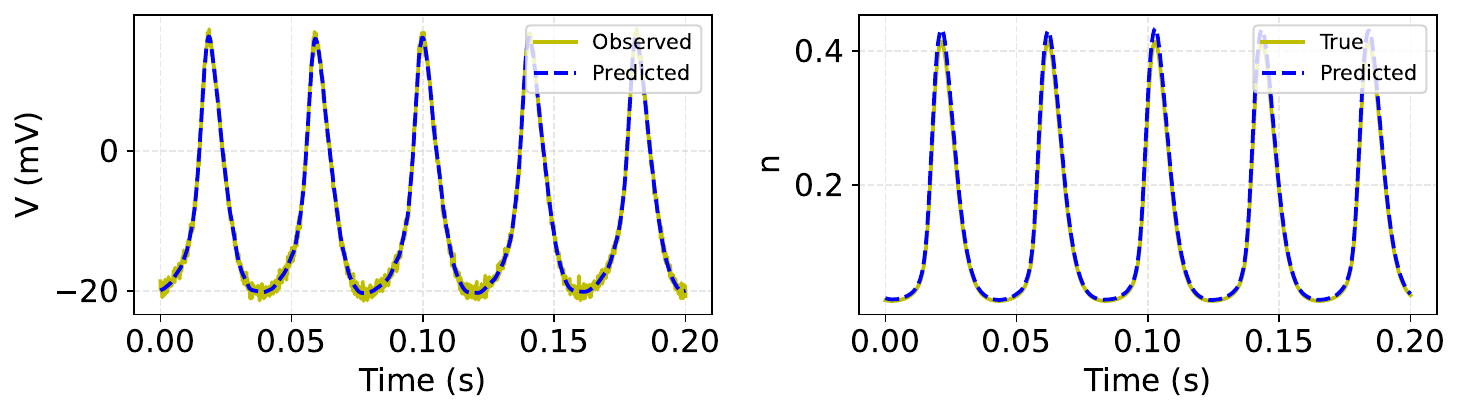}
        \caption{Large Noise: True: Homoclinic; Initial: 1}
    \end{subfigure}
    \caption{\small{Comparison between the ground-truth trajectories and the PINN-reconstructed trajectories for the Homoclinic regime under two noise levels based on the non-informative initial guess, where the PINN predictions are shown as dashed blue lines and the observation voltage $V$ and ground-truth gating $n$ trajectories are shown as solid yellow lines.}}
    \label{fig:SR_ML_tHomoclinic_Prediction}
\end{minipage}
\end{figure}

{\bf Forward trajectory reconstruction.}
Table~\ref{tab:SR_ML_tHomoclinic_Prediction} and Figure~\ref{fig:SR_ML_tHomoclinic_Prediction} present the forward state reconstruction performance of the PINN for the Homoclinic regime. Under the small-noise level ($1\%$), the PINN achieves highly accurate reconstruction across all initial guesses. The relative \(L^2\) error for the observed voltage remains at $0.21\%$, while errors for the unobserved gating variable \(n(t)\) stay below $1\%$. When the noise level is increased to $5\%$, reconstruction errors increase moderately but remain well-controlled. In this case, the voltage reconstruction error rises to approximately $0.97\%$, and the reconstruction errors for the unobserved variable $n(t)$ remain below approximately $6.5\%$. Despite the higher observational noise, the PINN continues to accurately recover the forward states of both observed and unobserved variables, while preserving the qualitative dynamical behavior of the Homoclinic regime.

Overall, these results demonstrate that the PINN provides robust parameter estimation and forward state reconstruction for the Homoclinic regime across different noise levels and initial guesses, even when only partial observations are available.

\begin{table}[t]
    \centering   
    \caption{\small{Estimated biophysical parameters for the Homoclinic regime under two observational noise levels ($1\%$ and $5\%$), evaluated using two initial guess strategies (regime-based initial guess and non-informative initial guess where all parameters are set to $1$), with observed data from bifurcation regime `t' and initial parameter guesses corresponding to bifurcation regime `g'.}}
    \label{tab:combined_results_SR_ML_homo}
    \begin{subtable}{\textwidth}
        \centering
        \caption{Homoclinic ($1\%$ noise)}
        \resizebox{\textwidth}{!}{
        \begin{tabular}{c|ccc|cc|cc}
        \hline
        t: Homoclinic & \multicolumn{3}{c}{PINN} & \multicolumn{2}{c}{UKF} & \multicolumn{2}{c}{4D-Var}\\
        \hline
         & g: Hopf & g: SNIC & g: 1 & g: Hopf & g: SNIC & g: Hopf & g: SNIC\\
         \hline
          & Est. (Rel. Err.) & Est. (Rel. Err.) & Est. (Rel. Err.) & Est. (Rel. Err.) & Est. (Rel. Err.) & Est. (Rel. Err.) & Est. (Rel. Err.) \\
        \hline
        $g_{\rm L}$ & $1.976 (1.2\%)$ & $1.981 (1.0\%)$ & $1.980 (1.0\%)$ & $0.316 (84.2\%)$ & $2.002 (0.1\%)$ & $3.166 (58.3\%)$ & $1.954 (2.3\%)$ \\
        $g_{\rm K}$ & $8.044 (0.6\%)$ & $8.049 (0.6\%)$ & $8.052 (0.6\%)$ & $8.143 (1.8\%)$ & $8.007 (0.1\%)$ & $9.596 (20.0\%)$ & $8.949 (11.9\%)$ \\
        $g_{\rm Ca}$ & $3.981 (0.5\%)$ & $4.003 (0.1\%)$ & $3.999 (<0.1\%)$ & $4.599 (15.0\%)$ & $4.005 (0.1\%)$ & $9.139 (128.5\%)$ & $3.953 (1.2\%)$ \\
        $\phi$ & $0.231 (0.3\%)$ & $0.232 (0.8\%)$ & $0.232 (0.7\%)$ & $0.296 (28.6\%)$ & $0.230 (0.2\%)$ & $0.422 (83.5\%)$ & $0.225 (2.3\%)$ \\
        $V_1$ & $-1.215 (1.3\%)$ & $-1.189 (0.9\%)$ & $-1.194 (0.5\%)$ & $-2.018 (68.2\%)$ & $-1.208 (0.7\%)$ & $2.891 (340.9\%)$ & $-1.265 (5.4\%)$ \\
        $V_2$ & $17.912 (0.5\%)$ & $17.953 (0.3\%)$ & $17.945 (0.3\%)$ & $19.077 (6.0\%)$ & $17.988 (0.1\%)$ & $24.695 (37.2\%)$ & $17.962 (0.2\%)$ \\
        $V_3$ & $12.078 (0.6\%)$ & $12.071 (0.6\%)$ & $12.080 (0.7\%)$ & $2.613 (78.2\%)$ & $11.994 (0.1\%)$ & $6.533 (45.6\%)$ & $13.848 (15.4\%)$ \\
        $V_4$ & $17.477 (0.4\%)$ & $17.565 (0.9\%)$ & $17.550 (0.9\%)$ & $29.426 (69.1\%)$ & $17.399 (<0.1\%)$ & $27.765 (59.6\%)$ & $18.223 (4.7\%)$ \\
        \hline
        \end{tabular}
        }
        \label{tab:homoclinic_est_1}
    \end{subtable}
    
    \begin{subtable}[h]{\textwidth}
        \centering
        \caption{Homoclinic ($5\%$ noise)}
        \resizebox{\textwidth}{!}{
        \begin{tabular}{c|ccc|cc|cc}
        \hline
        t: Homoclinic & \multicolumn{3}{c}{PINN} & \multicolumn{2}{c}{UKF} & \multicolumn{2}{c}{4D-Var}\\
        \hline
         & g: Hopf & g: SNIC & g: 1 & g: Hopf & g: SNIC & g: Hopf & g: SNIC\\
         \hline
          & Est. (Rel. Err.) & Est. (Rel. Err.) & Est. (Rel. Err.) & Est. (Rel. Err.) & Est. (Rel. Err.) & Est. (Rel. Err.) & Est. (Rel. Err.) \\
        \hline
        $g_{\rm L}$ & $1.925 (3.8\%)$ & $1.988 (0.6\%)$ & $1.954 (2.3\%)$ & $-0.170 (108.5\%)$ & $2.005 (0.2\%)$ & $3.049 (52.4\%)$ & $1.905 (4.8\%)$ \\
        $g_{\rm K}$ & $7.765 (2.9\%)$ & $7.920 (1.0\%)$ & $7.787 (2.7\%)$ & $6.932 (13.4\%)$ & $8.014 (0.2\%)$ & $9.637 (20.5\%)$ & $9.092 (13.6\%)$ \\
        $g_{\rm Ca}$ & $3.861 (3.5\%)$ & $4.157 (3.9\%)$ & $3.990 (0.2\%)$ & $3.190 (20.3\%)$ & $4.008 (0.2\%)$ & $8.905 (122.6\%)$ & $3.955 (1.1\%)$ \\
        $\phi$ & $0.228 (0.9\%)$ & $0.244 (6.3\%)$ & $0.236 (2.4\%)$ & $0.230 (0.1\%)$ & $0.231 (0.5\%)$ & $0.418 (81.8\%)$ & $0.229 (0.6\%)$ \\
        $V_1$ & $-1.430 (19.2\%)$ & $-1.084 (9.6\%)$ & $-1.274 (6.2\%)$ & $-1.931 (60.9\%)$ & $-1.207 (0.6\%)$ & $2.680 (323.3\%)$ & $-1.300 (8.3\%)$ \\
        $V_2$ & $17.522 (2.7\%)$ & $18.067 (0.4\%)$ & $17.756 (1.4\%)$ & $19.410 (7.8\%)$ & $17.975 (0.1\%)$ & $24.396 (35.5\%)$ & $17.800 (1.1\%)$ \\
        $V_3$ & $11.495 (4.2\%)$ & $11.536 (3.9\%)$ & $11.425 (4.8\%)$ & $2.630 (78.1\%)$ & $11.993 (0.1\%)$ & $6.792 (43.4\%)$ & $14.036 (17.0\%)$ \\
        $V_4$ & $16.908 (2.8\%)$ & $18.143 (4.3\%)$ & $17.408 (<0.1\%)$ & $29.307 (68.4\%)$ & $17.398 (<0.1\%)$ & $27.841 (60.0\%)$ & $18.651 (7.2\%)$ \\
        \hline
        \end{tabular}
        }
        \label{tab:homoclinic_est_5}
    \end{subtable}
\end{table}

\begin{table}[t]
    \centering
    \caption{\small{Comparison between the ground-truth trajectories and the PINN-reconstructed solutions for the Homoclinic regime under two noise levels using state reconstruction errors.}}
    \label{tab:SR_ML_tHomoclinic_Prediction}
    \resizebox{0.6\textwidth}{!}{
    \begin{tabular}{c|c|c|c|c}
        \hline
        \textbf{Noise Level} & \textbf{True Regime} & \textbf{Initial Guess} & \textbf{$e_{\hat{V}}$} & \textbf{$e_{\hat{n}}$} \\
        \hline
         $1\%$ & HC & Hopf & $0.21\%$ & $0.37\%$  \\
         $1\%$ & HC & SNIC & $0.21\%$ & $0.38\%$  \\
         $1\%$ & HC & $1$ & $0.21\%$ & $0.35\%$  \\
        \hline
         $5\%$ & HC & Hopf & $0.97\%$ & $2.32\%$  \\
         $5\%$ & HC & SNIC & $0.97\%$ & $6.47\%$  \\
         $5\%$ & HC & $1$ & $0.97\%$ & $4.45\%$  \\
        \hline
    \end{tabular}}
\end{table}

\subsubsection{Bifurcation diagrams} 

\begin{figure}
    \centering
    \includegraphics[width=0.32\linewidth]{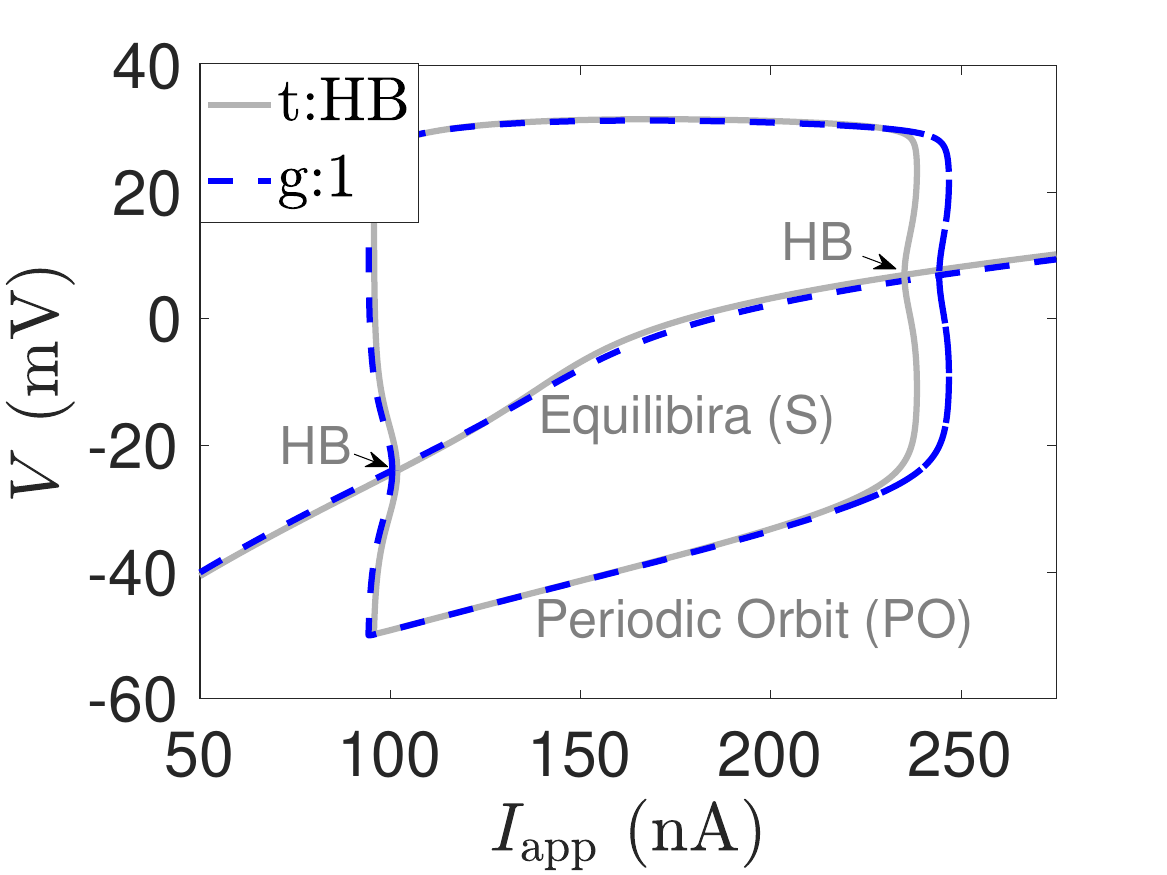}
    \includegraphics[width=0.32\linewidth]{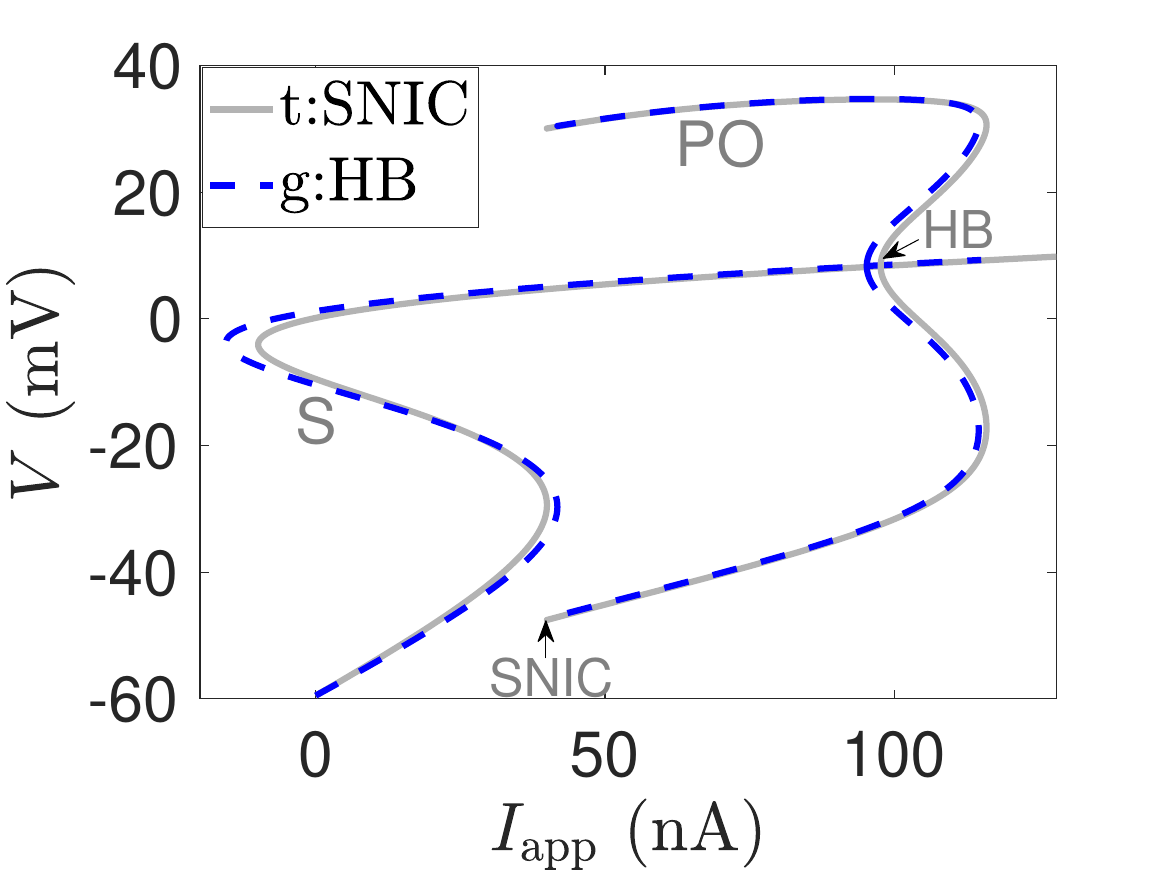}
    \includegraphics[width=0.32\linewidth]{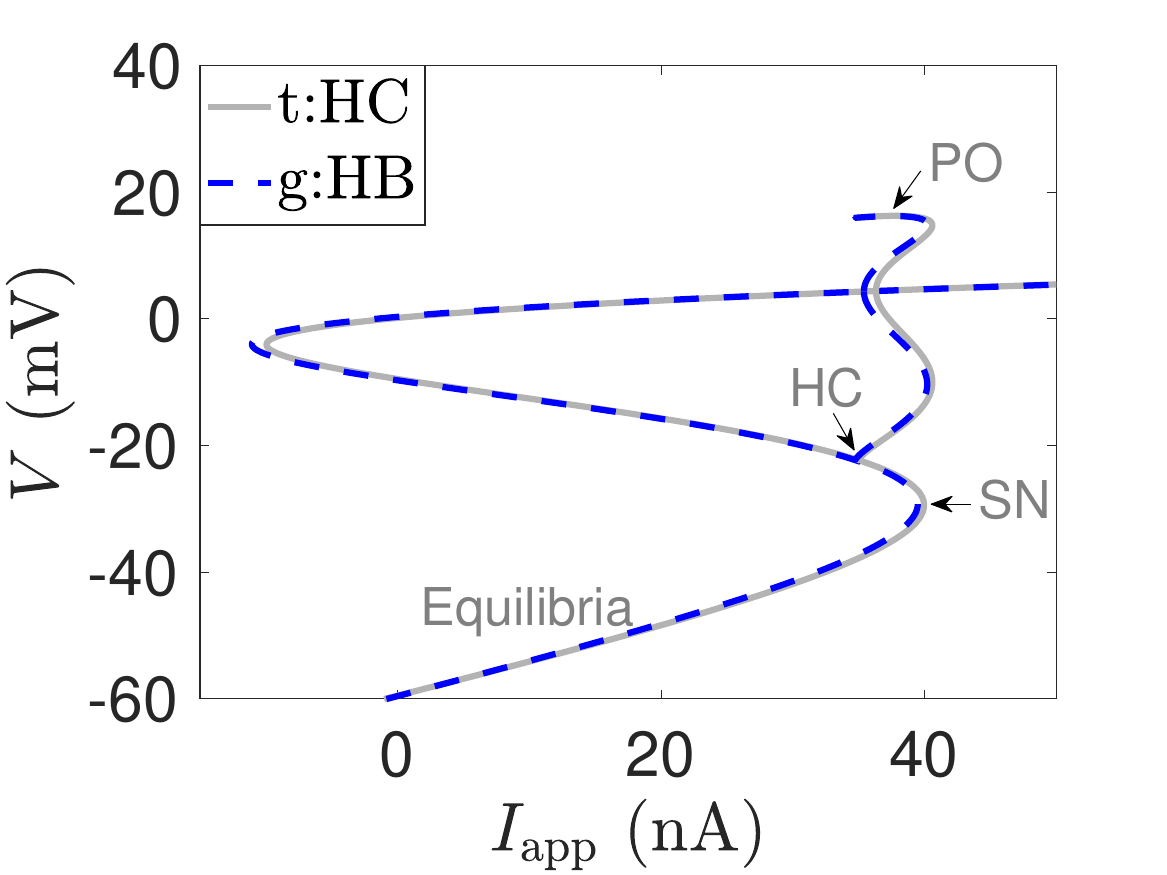}
    \caption{\label{fig:SMLbif} Bifurcation diagrams for the Morris-Lecar Model in three different bifurcation regimes: Hopf (left), SNIC (middle) and Homoclinic (right). Shown are the equilibrium curves and the minimum and maximum $V$ along families of periodic orbit (PO) solutions as functions of the bifurcation parameter $I_{\rm app}$. Key bifurcation points on the ground-truth bifurcation diagrams are indicated. 
    Gray curves represent the true diagrams, while blue dotted curves correspond to the diagrams produced from PINN-estimated parameters under 5\% relative noise levels (Tables \ref{tab:combined_results_SR_ML_hopf}, \ref{tab:combined_results_SR_ML_snic} and \ref{tab:combined_results_SR_ML_homo}). The displayed results correspond to representative worst-case estimates. In the legend, the notation ``t:" denotes the true bifurcation regime and ``g:" indicates initial parameter guesses. }
\end{figure}

Although the PINN framework yields highly accurate and robust parameter estimates and state trajectory reconstructions across multiple noise levels and initialization schemes, even small parameter discrepancies may induce qualitative changes in the system dynamics, a phenomena called bifurcations in dynamical systems \cite{izhikevich2007dynamical}. As a result, pointwise agreement in parameter values or time-series trajectories alone is not sufficient to guarantee that the underlying dynamical structure has been faithfully recovered. To examine this issue, we further validate the PINN-inferred models by examining whether the estimated parameter sets reproduce the correct bifurcation structure of the SML system as the applied current $I_{\rm app}$ is varied. For each bifurcation regime, we present only the worst case with the largest parameter estimation error and compare the resulting bifurcation diagram (Figure \ref{fig:SMLbif}, blue dotted curves) with the corresponding ground-truth diagram (Figure \ref{fig:SMLbif}, gray curves).
As shown in Figure \ref{fig:SMLbif}, the PINN-based estimates not only recover the correct bifurcation structure, but also demonstrate close quantitative agreement with the true bifurcation diagrams. Key bifurcation points (HB, SNIC, HC, and saddle node (SN)) are indicated. Additional background on the bifurcation analysis of the SML model can be found in \cite{izhikevich2007dynamical,ermentrout2010mathematical}. 

\subsection{Applications to the Bursting Morris–Lecar Model}
\label{subsec:Ex2}

\begin{figure}[t]
    \centering
     \includegraphics[width=0.8\linewidth]{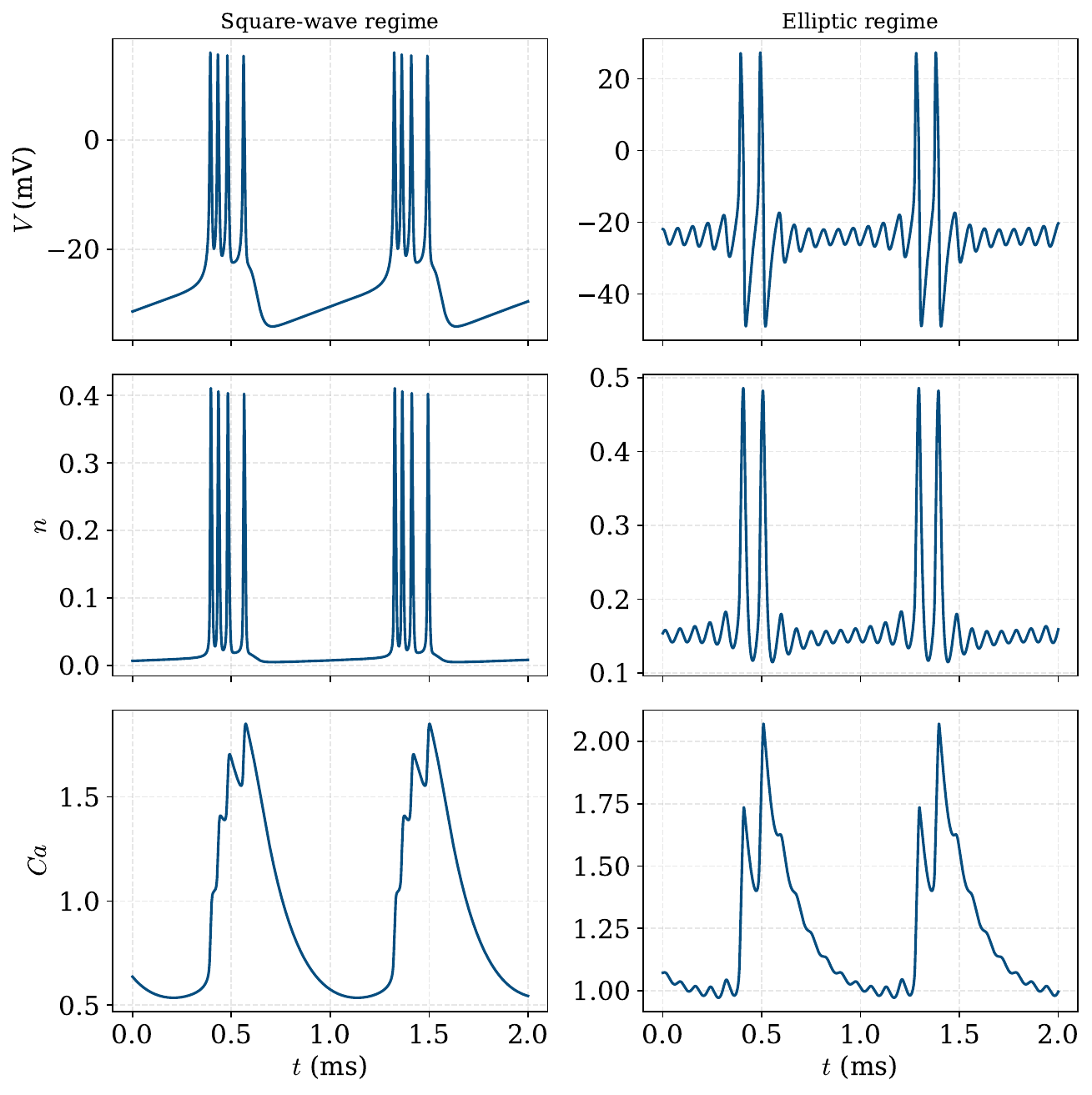}
    \caption{Ground-truth dynamics of the Morris–Lecar model under the two distinct bursting regimes: square-wave (left panel) and elliptic (right panel). Each row corresponds to one state variable: membrane potential $V$, gating variable $n$, and intracellular calcium concentration $Ca$.}
    \label{fig:BR_ML_true_bifurcation}
\end{figure}

Next, we consider a three-dimensional Morris–Lecar (ML) model that incorporates slow calcium dynamics and the $K_{\mathrm{Ca}}$ current, which gives rise to two types of bursting regimes: square-wave and elliptic bursting. The equations for the bursting Morris-Lecar (BML) model \cite{moye2018data, wang2016multiple} are given as follows:
\begin{align}
\label{BR_ML_Eq_V} C_m \frac{dV}{dt} &= I_{\text{app}} - g_L(V - E_L) - g_K n (V - E_K) \\
&- g_{Ca} m_{\infty}(V) (V - E_{Ca}) - g_{KCa} z (V - E_K), \notag\\
\label{BR_ML_Eq_n} \frac{dn}{dt} &= \phi (n_{\infty}(V) - n)/\tau_n(V), \\
\label{BR_ML_Eq_Ca} \frac{dCa}{dt} &= \epsilon (-\mu I_{Ca} - Ca),
\end{align}
with
\begin{align}
m_{\infty}(V) &= \frac{1}{2}(1 + \tanh((V - V_1)/V_2)) \\
\tau_n &= 1/\cosh((V - V_3)/2V_4), \\
n_{\infty}(V) &= \frac{1}{2}(1 + \tanh((V - V_3)/V_4)), \\
I_{Ca} &= g_{Ca} m_{\infty}(V) (V - E_{Ca}), \\
z &= \frac{Ca}{Ca + 1}.
\end{align}
In the BML model, a total of nine biophysical parameters are estimated across the two bursting regimes, with their ground-truth values summarized in Table~\ref{tab:bursting_params}, while the remaining parameters are assumed to be known. $I_{\mathrm{app}}$ is $45$ for square-wave regime and $120$ for elliptic regime. In both bursting regimes, $\varepsilon=0.005$ and $\mu=0.02$. All other known parameters are the same as those in the spiking Morris--Lecar model, with $C_m = 20$, $E_{Ca} = 120$, $E_K = -84$, and $E_L = -60$ assumed to be known and fixed.
The true trajectories for $V(t)$, $n(t)$, and $Ca(t)$ with parameters given in Table~\ref{tab:bursting_params} are shown in Figure~\ref{fig:BR_ML_true_bifurcation}. 

\begin{table}[t]
\centering
\caption{Parameters for the bursting Morris--Lecar model \eqref{BR_ML_Eq_V}--\eqref{BR_ML_Eq_Ca}.}
\label{tab:bursting_params}
\footnotesize
\begin{tabular}{c|c|c}
\hline
Parameter & square-wave & elliptic \\
\hline
$\phi$            & $0.23$  & $0.04$  \\
$g_{\mathrm{Ca}}$ & $4.0$   & $4.4$   \\
$V_3$             & $12.0$  & $2.0$   \\
$V_4$             & $17.4$  & $30.0$  \\
$g_K$             & $8.0$   & $8.0$   \\
$g_L$             & $2.0$   & $2.0$   \\
$V_1$             & $-1.2$ & $-1.2$ \\
$V_2$             & $18.0$  & $18.0$  \\
$g_{\mathrm{KCa}}$ & $0.25$  & $0.75$  \\
\hline
\end{tabular}
\end{table}

\subsubsection{Square-wave regime}
For the square-wave regime, the PINN is evaluated under two noise levels ($1\%$ and $5\%$) using initial guesses from the elliptic bursting regime and a non-informative initialization where all parameters are set to $1$. 

{\bf Parameter estimation.}
Table~\ref{tab:combined_results_BR_ML_sw} summarizes the parameter estimation results for BML model in the square-wave bursting regime using two different initial guess strategies. Under both initializations, the estimated parameters align closely with their ground-truth values. 
In the case of a small-noise level ($1\%$, Table~\ref{tab:square_wave_est1}), the PINN delivers accurate parameter estimation for all variables. When initialized from the elliptic bursting regime, most relative errors remain below $1\%$, with only one parameter ($V_1$) exhibiting a slightly larger deviation of $4.3\%$. When initialized from the non-informative initial guess, all estimated parameters maintain relative errors below $2\%$, with the largest deviation being $2.0\%$ for $V_1$.

Table~\ref{tab:square_wave_est5} presents the parameter estimation results at the $5\%$ noise level. Despite the increased noise level, the inferred parameters remain close to their true values for both initial guesses. Most parameters exhibit relative errors below approximately $5\%$, while a small number of parameters (notably $g_K$, $V_1$, and $V_3$) show larger deviations, with the maximum relative error reaching $16.1\%$.

\begin{table}[t]
    \centering   
    \caption{\small{Estimated biophysical parameters for the SML model in the square-wave bursting regime under two observational noise levels ($1\%$ and $5\%$), evaluated using two initial guess strategies (regime-based initial guess and non-informative initial guess where all parameters are set to $1$).}}
    \label{tab:combined_results_BR_ML_sw}
    \begin{subtable}[h]{0.9\textwidth}
        \centering
        \caption{Square-wave ($1\%$ noise)}
        \label{tab:square_wave_est1}
        \resizebox{\textwidth}{!}{
        \begin{tabular}{c|c|c|c|c|c}
            \hline
            Parameter & True Value & Initial Guess: Elliptic & Relative Error & Initial Guess: $1$ & Relative Error \\
            \hline
            $g_{\rm L}$ & $2.0$ & $1.996$ & $0.2\%$ & $1.998$ & $0.1\%$ \\
            $g_{\rm K}$ & $8.0$ & $7.996$ & $0.1\%$ & $8.009$ & $0.1\%$ \\
            $g_{\rm Ca}$ & $4.0$ & $3.968$ & $0.8\%$ & $3.990$ & $0.3\%$ \\
            $\phi$ & $0.23$ & $0.228$ & $0.9\%$ & $0.229$ & $0.4\%$ \\
            $V_1$ & $-1.2$ & $-1.251$ & $4.3\%$ & $-1.224$ & $2.0\%$ \\
            $V_2$ & $18.0$ & $17.941$ & $0.3\%$ & $17.976$ & $0.1\%$ \\
            $V_3$ & $12.0$ & $12.024$ & $0.2\%$ & $12.035$ & $0.3\%$ \\
            $V_4$ & $17.4$ & $17.275$ & $0.7\%$ & $17.380$ & $0.1\%$ \\
            $g_{\rm KCa}$ & $0.25$ & $0.250$ & $0.1\%$ & $0.250$ & $0.2\%$ \\
            \hline
        \end{tabular}}
    \end{subtable}
    \begin{subtable}[h]{0.9\textwidth}
        \centering
        \caption{Square-wave ($5\%$ noise)}
        \label{tab:square_wave_est5}
        \resizebox{\textwidth}{!}{
        \begin{tabular}{c|c|c|c|c|c}
            \hline
            Parameter & True Value & Initial Guess: Elliptic & Relative Error & Initial Guess: 1 & Relative Error \\
            \hline
            $g_L$ & $2.0$ & $1.996$ & $0.2\%$ & $2.021$ & $1.1\%$ \\
            $g_K$ & $8.0$ & $8.943$ & $11.8\%$ & $9.033$ & $12.9\%$ \\
            $g_{Ca}$ & $4.0$ & $3.907$ & $2.3\%$ & $4.132$ & $3.3\%$ \\
            $\phi$ & $0.23$ & $0.218$ & $5.2\%$ & $0.232$ & $0.9\%$ \\
            $V_1$ & $-1.2$ & $-1.322$ & $10.1\%$ & $-1.043$ & $13.0\%$ \\
            $V_2$ & $18.0$ & $17.985$ & $0.1\%$ & $18.346$ & $1.9\%$ \\
            $V_3$ & $12.0$ & $13.828$ & $15.2\%$ & $13.931$ & $16.1\%$ \\
            $V_4$ & $17.4$ & $17.579$ & $1.0\%$ & $18.695$ & $7.4\%$ \\
            $g_{KCa}$ & $0.25$ & $0.251$ & $0.6\%$ & $0.251$ & $0.3\%$ \\
            \hline
        \end{tabular}}
    \end{subtable}
\end{table}

\begin{figure}[t]
    \centering
    \includegraphics[width=\linewidth]{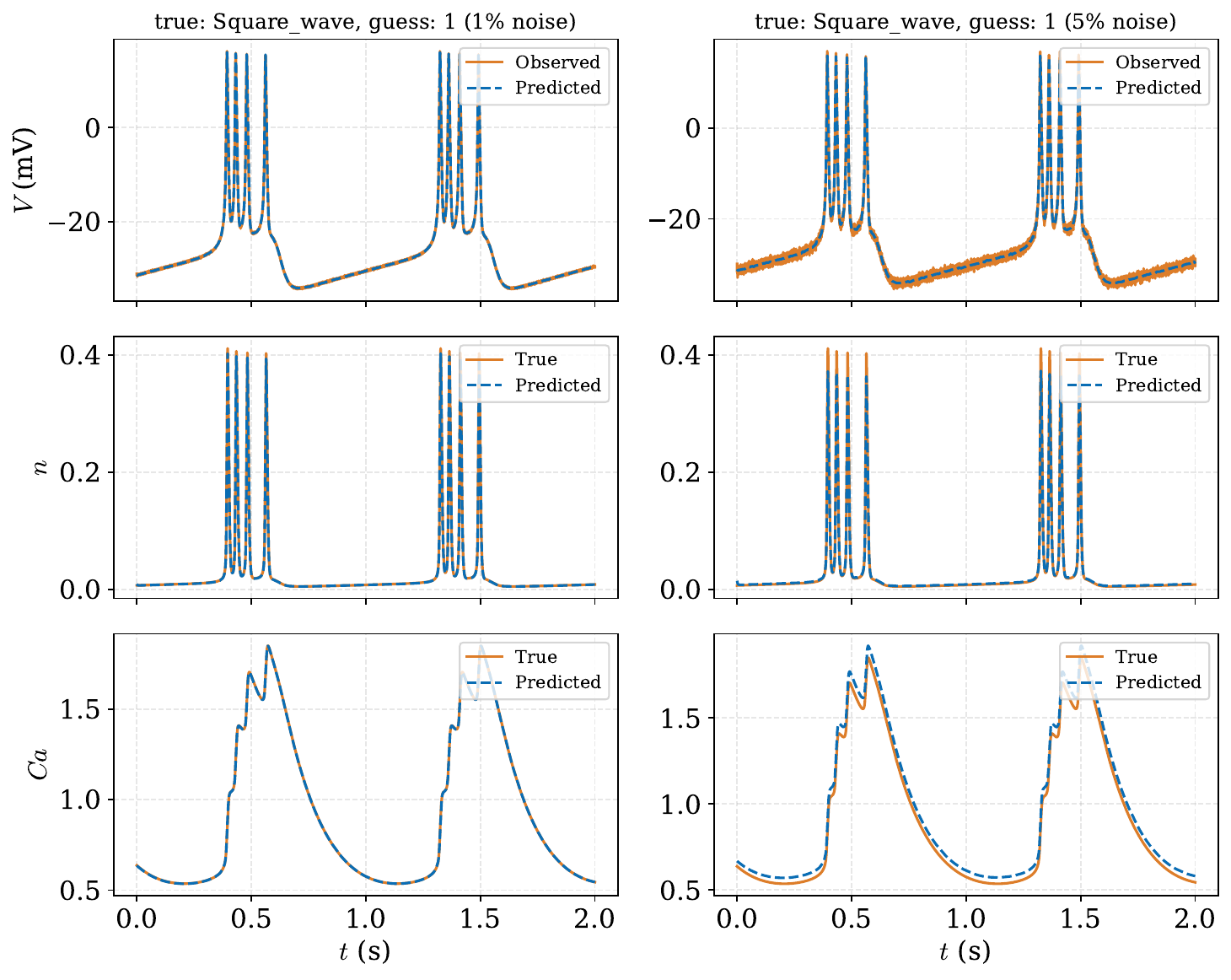}
    \caption{\small{Comparison between the ground-truth trajectories and the PINN-reconstructed solutions for the BML model in the square-wave bursting regime under two noise levels based on the non-informative initial guess. The PINN predictions are shown as dashed blue line, while the true trajectories are shown as solid orange line.}}
    \label{fig:BR_ML_Square_wave_Prediction}
\end{figure}

\begin{table}[t]
    \centering
    \caption{\small{Comparison between the ground-truth trajectories and the PINN-reconstructed solutions for the BML model in the square-wave bursting regime under two noise levels using state reconstruction errors.}}
    \label{tab:BR_ML_Square_wave_Prediction}
    \resizebox{0.7\textwidth}{!}{
    \begin{tabular}{c|c|c|c|c|c}
        \hline
        \textbf{Noise Level} & \textbf{True Regime} & \textbf{Initial Guess} &\textbf{$e_{\hat{V}}$} & \textbf{$e_{\hat{n}}$} & \textbf{$e_{\widehat{Ca}}$} \\
        \hline
        $1\%$ & square-wave & Elliptic & $0.09\%$ & $0.79\%$ & $0.85\%$ \\
        $1\%$ & square-wave & $1$ & $0.09\%$ & $0.36\%$ & $0.30\%$ \\
        \hline
        $5\%$ & square-wave & Elliptic & $0.25\%$ & $16.79\%$ & $8.93\%$ \\
        $5\%$ & square-wave & $1$ & $0.25\%$ & $11.32\%$ & $2.85\%$ \\
        \hline
    \end{tabular}}
\end{table}

{\bf Forward trajectory reconstruction.}
Figure~\ref{fig:BR_ML_Square_wave_Prediction} illustrates the state reconstruction performance of the PINN for the square-wave bursting regime under both small-noise and large-noise levels based on the non-informative initial guess. The reconstructed solutions obtained from the trained PINN are compared with the ground-truth solutions. Despite this non-informative initial guess, the PINN successfully recovers both the correct qualitative bursting pattern and quantitatively accurate state solutions.  

Besides, Table~\ref{tab:BR_ML_Square_wave_Prediction} shows that under the $1\%$ noise level, the relative $L^2$ error for the observed membrane voltage $V$ remains as low as $0.09\%$ for both initializations, while the reconstruction errors for all unobserved state variables remain below $1\%$. At the higher noise level of $5\%$, voltage reconstruction errors increase moderately, whereas the reconstruction errors for the unobserved variables, in particular the gating variable $n(t)$, increase more substantially. Nevertheless, Figure~\ref{fig:BR_ML_Square_wave_Prediction} shows that the dominant source of reconstruction error is magnitude deviation rather than phase shift, indicating that the PINN preserves the underlying dynamical structure even under elevated observational noise.

\subsubsection{Elliptic regime}
For the elliptic regime, the PINN is evaluated under two noise levels ($1\%$ and $5\%$) using initial guesses from the square-wave bursting regime, and a non-informative initialization in which all parameters are set to $1$.

{\bf Parameter estimation.}
Table~\ref{tab:combined_results_BR_ML_e} reports the estimated biophysical parameters for the elliptic bursting regime under small-noise ($1\%$) and large-noise ($5\%$) levels, using  different initial guess strategies. Under the small-noise level,parameter estimation is highly accurate:  the maximum relative error among all estimated parameters is $1.7\%$ (for $V_3$) when using the non-informative initial guess, and $1.4\%$ (also for $V_3$) when using the square-wave regime initial guess. All remaining parameters are recovered with relative errors below $0.3\%$, indicating highly accurate parameter inference. When the noise level increases to $5\%$, the estimation accuracy degrades mildly. The maximum relative error among all parameters remains reasonably small, reaching $2.7\%$ for both initial guesses (for $V_1$), while all other parameters exhibit relative errors below $1.4\%$. Across both noise levels and both initialization strategies, the estimated parameters consistently agree well with their ground-truth values, demonstrating the strong robustness of the PINN to observational noise and initial guess selection in the elliptic bursting regime.

\begin{table}[t]
    \centering   
    \caption{\small{Estimated biophysical parameters for the Elliptic bursting regime under two observational noise levels ($1\%$ and $5\%$), evaluated using two initial guess strategies (regime-based initial guess and non-informative initial guess where all parameters are set to $1$).}}
    \label{tab:combined_results_BR_ML_e}
    \begin{subtable}[h]{0.9\textwidth}
        \centering
        \caption{Elliptic ($1\%$ noise)}
        \label{tab:elliptic_est1}
        \resizebox{\textwidth}{!}{
        \begin{tabular}{c|c|c|c|c|c}
            \hline
            Parameter & True Value & Initial Guess: Square-wave & Relative Error & Initial Guess: 1 & Relative Error \\
            \hline
            $g_L$ & $2.0$ & $2.004$ & $0.2\%$ & $2.002$ & $0.1\%$ \\
            $g_K$ & $8.0$ & $8.008$ & $0.1\%$ & $8.012$ & $0.1\%$ \\
            $g_{Ca}$ & $4.4$ & $4.403$ & $0.1\%$ & $4.403$ & $0.1\%$ \\
            $\phi$ & $0.04$ & $0.040$ & $0.1\%$ & $0.040$ & $0.1\%$ \\
            $V_1$ & $-1.2$ & $-1.203$ & $0.2\%$ & $-1.202$ & $0.2\%$ \\
            $V_2$ & $18.0$ & $18.000$ & $<0.1\%$ & $17.998$ & $<0.1\%$ \\
            $V_3$ & $2.0$ & $2.028$ & $1.4\%$ & $2.034$ & $1.7\%$ \\
            $V_4$ & $30.0$ & $30.007$ & $<0.1\%$ & $30.016$ & $0.1\%$ \\
            $g_{KCa}$ & $0.75$ & $0.748$ & $0.2\%$ & $0.748$ & $0.3\%$ \\
            \hline
        \end{tabular}}
    \end{subtable}
    \begin{subtable}[h]{0.9\textwidth}
        \centering
        \caption{Elliptic ($5\%$ noise)}
        \label{tab:elliptic_est5}
        \resizebox{\textwidth}{!}{
        \begin{tabular}{c|c|c|c|c|c}
            \hline
            Parameter & True Value & Initial Guess: Square-wave & Relative Error & Initial Guess: 1 & Relative Error \\
            \hline
            $g_L$ & $2.0$ & $2.029$ & $1.4\%$ & $2.028$ & $1.4\%$ \\
            $g_K$ & $8.0$ & $7.969$ & $0.4\%$ & $7.970$ & $0.4\%$ \\
            $g_{Ca}$ & $4.4$ & $4.429$ & $0.7\%$ & $4.429$ & $0.7\%$ \\
            $\phi$ & $0.04$ & $0.040$ & $0.5\%$ & $0.040$ & $0.5\%$ \\
            $V_1$ & $-1.2$ & $-1.168$ & $2.7\%$ & $-1.168$ & $2.7\%$ \\
            $V_2$ & $18.0$ & $18.086$ & $0.5\%$ & $18.085$ & $0.5\%$ \\
            $V_3$ & $2.0$ & $1.998$ & $0.1\%$ & $1.998$ & $0.1\%$ \\
            $V_4$ & $30.0$ & $29.925$ & $0.3\%$ & $29.925$ & $0.2\%$ \\
            $g_{KCa}$ & $0.75$ & $0.753$ & $0.4\%$ & $0.752$ & $0.3\%$ \\
            \hline
        \end{tabular}}
    \end{subtable}
\end{table}

\begin{figure}[t]
    \centering
    \includegraphics[width=\linewidth]{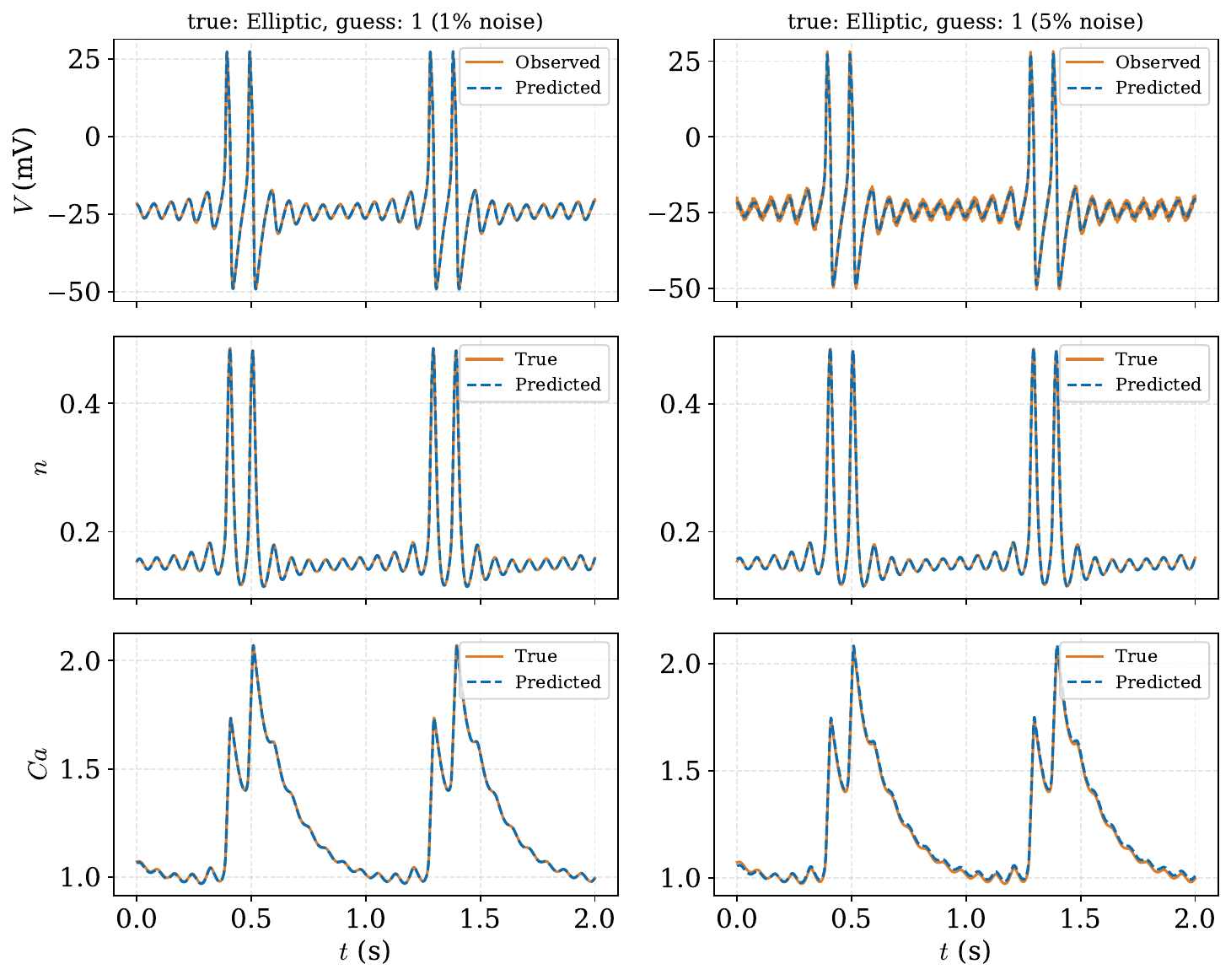}
    \caption{\small{Comparison between the ground-truth trajectories and the PINN-reconstructed solutions for the elliptic bursting regime under two noise levels based on the non-informative initial guess}, where the PINN predictions shown as dashed blue line and true trajectories shown as solid orange line.}
    \label{fig:BR_ML_Elliptic_Prediction}
\end{figure}

\begin{table}[t]
    \centering
    \caption{\small{Comparison between the ground-truth trajectories and the PINN-reconstructed solutions for the elliptic bursting regime under two noise levels using state reconstruction errors.}}
    \label{tab:BR_ML_Elliptic_Prediction}
    \resizebox{0.7\textwidth}{!}{
    \begin{tabular}{c|c|c|c|c|c}
        \hline
        \textbf{Noise Level} & \textbf{True Regime} & \textbf{Initial Guess} &\textbf{$e_{\hat{V}}$} & \textbf{$e_{\hat{n}}$} & \textbf{$e_{\widehat{Ca}}$} \\
        \hline
        $1\%$ & elliptic & square-wave & $0.09\%$ & $0.15\%$ & $0.09\%$ \\
        $1\%$ & elliptic & $1$ & $0.09\%$ & $0.14\%$ & $0.10\%$ \\
        \hline
        $5\%$ & elliptic & square-wave & $0.33\%$ & $0.44\%$ & $0.59\%$ \\
        $5\%$ & elliptic & $1$ & $0.33\%$ & $0.49\%$ & $0.54\%$ \\
        \hline
    \end{tabular}}
\end{table}

{\bf Forward trajectory reconstruction.}
The forward state reconstruction performance of the proposed PINN for the elliptic bursting regime is summarized in Table~\ref{tab:BR_ML_Elliptic_Prediction}. Under the small-noise level ($1\%$), the reconstructed solutions show excellent agreement with the ground-truth solutions, with relative $L^2$ errors remaining below $0.2\%$ across all state variables. These results indicate that the PINN achieves highly accurate state recovery regardless of whether it is initialized from the square-wave regime or from a non-informative initial guess with all parameters set to $1$. When the observational noise level increases to $5\%$, the reconstruction errors increase moderately but remain well controlled. In this case, the relative $L^2$ errors for all state variables remain below $0.6\%$, while the reconstructed solutions continue to accurately reproduce the correct oscillatory behavior of the elliptic bursting dynamics as shown in Figure~\ref{fig:BR_ML_Elliptic_Prediction} (based on the non-informative initial guesses).

\subsubsection{Bifurcation diagrams} 

As we mentioned before, even very small discrepancies in parameter estimates can lead to qualitative different bursting dynamics. We therefore compute bifurcation diagrams for each case using the estimated parameters from the worst-case scenarios (i.e., for the square-wave regime we use the estimates obtained with the non-informative guess, and for the elliptic regime we use the estimates obtained with an initial guess from the square-wave regime, both under the $5\%$ noise level) and compare them to their true counterparts. 

\begin{figure}
    \centering
    \includegraphics[width=0.48\linewidth]{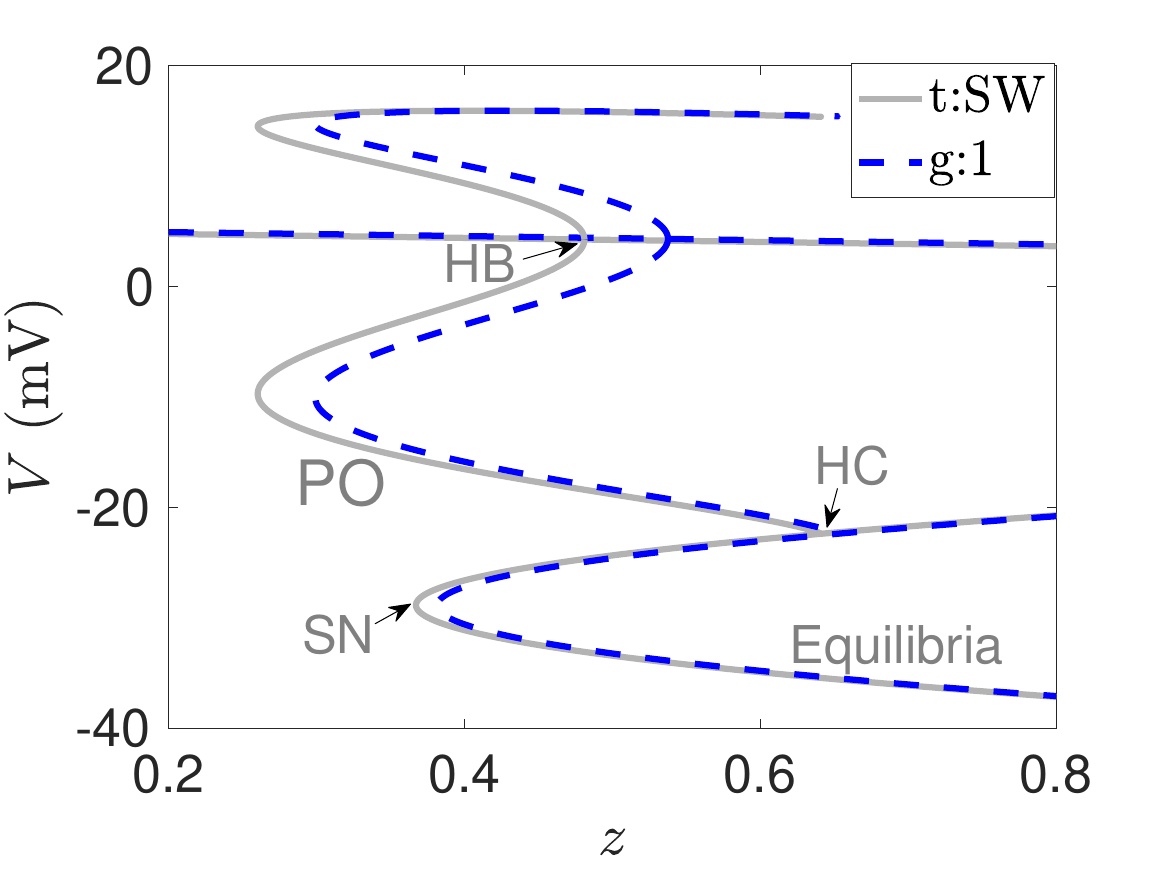}
    \includegraphics[width=0.48\linewidth]{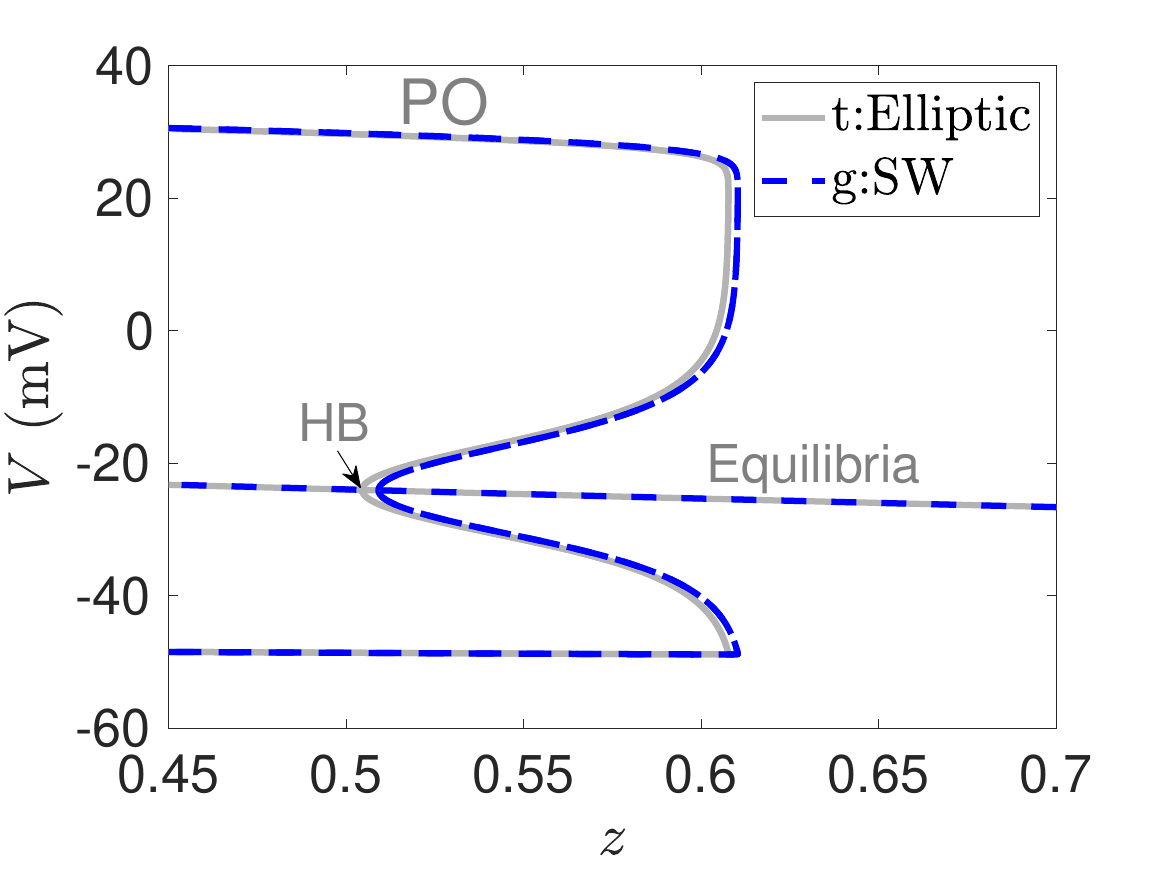}
    \caption{\label{fig:BML-bif} Bifurcation diagrams for the Morris-Lecar Model at two different bursting regimes: Square-Wave (left), and Elliptic (right). The gray lines represent the true diagrams, and the blue dotted lines correspond to the diagrams produced from the PINN-estimated parameters under 5\% relative noise levels (Tables \ref{tab:combined_results_BR_ML_sw} and \ref{tab:combined_results_BR_ML_e}). The displayed results correspond to representative worst-case estimates. In the legend, the notation ``t:" denotes the true bifurcation regime and ``g:" indicates initial parameter guesses.}
\end{figure}

Figure \ref{fig:BML-bif} shows that, in both cases, the estimated models recover the same bifurcation structure as the ground-truth models used to generate the data. For the square-wave bursting regime (Figure \ref{fig:BML-bif}, left panel), the reconstructed bifurcation diagram is nearly identical to the true one, with only a small rightward shift in the upper Hopf bifurcation and associated periodic orbit (PO) branches. Notably, these estimates correspond to the most challenging setting, using a non-informative initial guess and a noise level of $5\%$. These results further demonstrate the robustness of our methods to inaccurate initial parameter guesses.

For the elliptic bursting regime (Figure \ref{fig:BML-bif}, right panel), the reconstructed bifurcation structure shows a near-perfect agreement with the ground-truth, capturing both equilibria and PO branches with high accuracy. This stands in contrast to previous studies reporting that UKF and 4D-Var struggle to estimate parameters such as $g_{\rm KCa}$ due to the pronounced timescale separation between slow $\rm Ca$ dynamics and the fast variables $(V, n)$ \cite{moye2018data}. To mitigate this limitation, prior work has proposed multi-stage estimation strategies that explicitly separate slow and fast dynamics \cite{kadakia2016nonlinear}, or the incorporation of slow calcium observations to improve the performance. In contrast, our PINN-framework achieves accurate parameter estimation and bifurcation reconstruction without requiring such staged procedures, and remains robust for fast-slow systems even when only one fast variable measurements are available.

We remark that although direct comparisons with UKF and 4D-Var are not performed for the BML model, the accuracy of the PINN-based parameter estimates is nevertheless supported by the close agreement between the true and estimated parameters (Tables~\ref{tab:BR_ML_Square_wave_Prediction} and~\ref{tab:BR_ML_Elliptic_Prediction}) and by the strong consistency between the true and reconstructed bifurcation diagrams (Figure~\ref{fig:BML-bif}).

\subsection{pre-B\"{o}tzinger complex (pBC) neuron model}
\label{subsec:Ex3}

Finally, we consider a more biologically detailed bursting model of pre-B\"{o}tzinger complex (pBC) neurons \cite{butera1999models,park2013cooperation,wang2016multiple}, given by the following equations:

\begin{subequations}\label{eq:pbc}
\begin{eqnarray}
\frac{dV}{dt}&=&(-g_{\rm L}(V-V_{\rm L})-g_{\rm K}n^4(V-V_{\rm K})-g_{\rm Na}m_{\infty}^3(V)(1-n)(V-V_{\rm Na})\label{eq:Biov1}\\
&&-g_{\rm NaP}mp_{\infty}(V)h(V-V_{\rm Na}))/C_m\nonumber \\
\frac{dn}{dt}&=&(n_{\infty}(V)-n)/{\tau_{n}(V)}\label{eq:Bion}\\
\frac{dh}{dt}&=&(h_{\infty}(V)-h)/{\tau_{h}(V)}\label{eq:Bioh}
\end{eqnarray}
\end{subequations}
with
\begin{subequations}\label{eq:termbio}
\begin{eqnarray}
x_{\infty}(V)&=& 1/(1+\exp((V-\theta_x)/\sigma_x)), \,\, x\in\{m,\, mp,\, n, \, h\}\\
 \tau_{x}(V)&=&\bar{\tau}_x/{\cosh}((V-\theta_{x})/2\sigma_{x}), \,\, x \in \{n, \, h \}\,.
\end{eqnarray}
\end{subequations}
In system \eqref{eq:Biov1}-\eqref{eq:Bioh}, $V$ denotes the voltage, and $n, h$ are voltage-dependent gating variables associated with the potassium and persistent sodium currents.
Default parameter values and corresponding units for this system are given in Table \ref{tab:parbio}.

\begin{figure}[t]
    \centering
    \includegraphics[width=\linewidth]{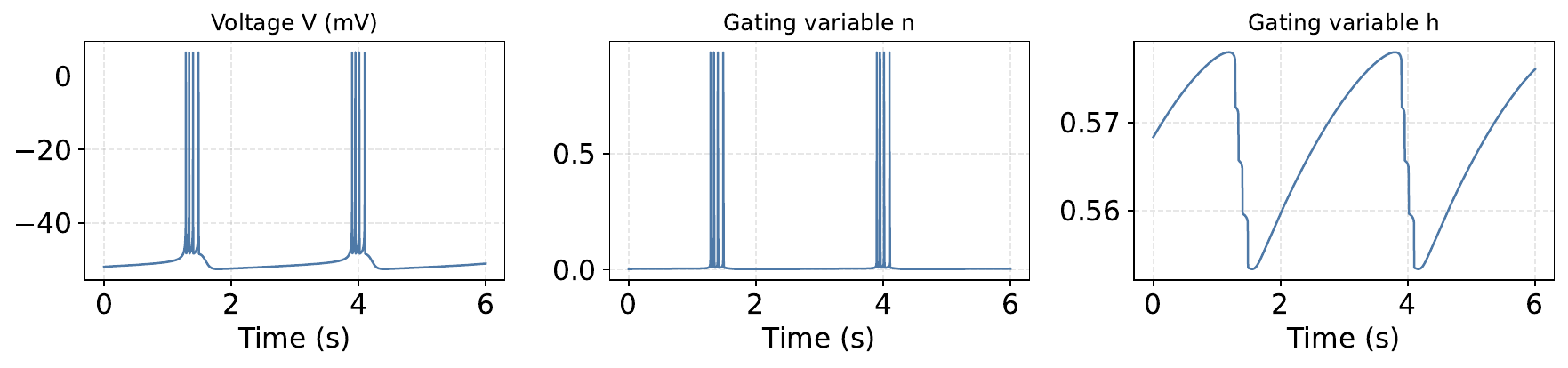}
    \caption{Ground-truth dynamics of the pBC model \eqref{eq:pbc} over two bursting cycles. Each column corresponds to one state variable: voltage $V$, gating variable $n$, and gating variable $h$.}
    \label{fig:pBC_true_bifurcation}
\end{figure}

\begin{table}[t]
\caption{Ground-truth parameter values used in the pre-B\"{o}tzinger complex (pBC) model \eqref{eq:pbc}.
Estimated parameters are listed in the first column, while the remaining parameters are assumed to be known.}
\label{tab:parbio}
\centering
\begin{tabular}{l c | l c | l c }
\hline
\multicolumn{2}{c|}{\textbf{\textcolor{black}{Estimated Parameters}}} &
\multicolumn{4}{c}{\textbf{Fixed Parameters}} \\
\hline

$\textcolor{black}{g_{\rm Na}}$ & \textcolor{black}{$28~{\rm nS}$}
& $C_m$ & $21~{\rm pF}$
& $\sigma_h$ & $5~{\rm mV}$\\

$\textcolor{black}{g_{\rm K}}$ & \textcolor{black}{$11.2~{\rm nS}$}
& $\bar{\tau}_h$ & $10{,}000~{\rm ms}$
& $\sigma_m$ & $-5~{\rm mV}$ \\

$\textcolor{black}{g_{\rm L}}$ & \textcolor{black}{$2.3~{\rm nS}$}
& $\sigma_{mp}$ & $-6~{\rm mV}$
& $\theta_m$ & $-34~{\rm mV}$ \\

$\textcolor{black}{g_{\rm NaP}}$ & \textcolor{black}{$2.0~{\rm nS}$}
& $\theta_{mp}$ & $-40~{\rm mV}$
& $\theta_h$ & $-48~{\rm mV}$ \\

$\textcolor{black}{V_{\rm Na}}$ & \textcolor{black}{$50~{\rm mV}$}
& $\bar{\tau}_n$ & $10~{\rm ms}$ & $\theta_n$ & $-29~{\rm mV}$\\

$\textcolor{black}{V_{\rm K}}$ & \textcolor{black}{$-85~{\rm mV}$}
& $\sigma_n$ & $-4~{\rm mV}$ \\

$\textcolor{black}{V_{\rm L}}$ & \textcolor{black}{$-58~{\rm mV}$}
 \\

\hline
\end{tabular}
\end{table}
The true trajectories of $V(t)$, $n(t)$, and $h(t)$ for the pBC model, computed using the ground-truth parameters in Table \ref{tab:parbio} and a modified Euler method, are shown in Figure~\ref{fig:pBC_true_bifurcation}.

\subsubsection{Relative  noise setting}
This subsection reports results obtained under the standard observational noise assumption, which is shown in \eqref{eq:stand_noise}. 
{\bf Parameter estimation.} 
Table~\ref{tab:combined_results_pBC} summarizes the parameter estimation results for the pBC model under two noise levels ($1\%$ and $10\%$). In this model, bursting is initiated via a fold bifurcation and terminated by a homoclinic bifurcation, corresponding to a square-wave type of bursting  \cite{wang2016multiple,rinzel1989analysis}. In this case, all estimated parameters for the pBC model are initialized to $1$ in the algorithm, without relying on any regime-based initial guesses. Note that the ground-truth values of most estimated parameters differ substantially from $1$ (see Table \ref{tab:parbio}, first column). 

Under a small-noise level ($1\%$, Table~\ref{tab:pBC_est1}), the PINN framework accurately recovers all biophysical parameters.  The largest deviation occurs for $g_{\rm K}$ ($9.2\%$ relative error), followed by $g_{\rm NaP}$ ($5.9\%$), while all remaining parameters are estimated with relative errors below $5\%$. These results demonstrate that the PINN framework is capable of reliable parameter recovery for the pBC model even without informative initial guesses. When the noise level increases to $10\%$ (Table~\ref{tab:pBC_est10}), estimation errors increase moderately, as expected. The largest deviations are around $10.2\%$ for $g_{\rm L}$ and $g_{\rm K}$, while all remaining parameters are recovered with relative errors below $5\%$. Overall, these results indicate that the PINN framework maintains stable and robust parameter estimation performance for the pBC model even under relatively high level of added noise.

\begin{table}[t]
    \centering   
    \caption{Estimated biophysical parameters for the pBC model under two observational noise levels ($1\%$ and $10\%$), evaluated using non-informative initial guess where all parameters are set to $1$.}
    \label{tab:combined_results_pBC}
    \begin{subtable}[h]{0.6\textwidth}
        \centering
        \caption{($1\%$ noise)}
        \label{tab:pBC_est1}
        \resizebox{\textwidth}{!}{
        \begin{tabular}{c|c|c|c}
            \hline
            Parameter & True Value & Initial Guess: 1 & Relative Error \\
            \hline
            $g_{\rm NaP}$ & $2.0$ & $2.118$ & $5.9\%$ \\
            $g_{\rm L}$ & $2.3$ & $2.378$ & $3.4\%$ \\
            $g_{\rm K}$ & $11.2$ & $12.228$ & $9.2\%$ \\
            $g_{\rm Na}$ & $28.0$ & $27.730$ & $1.0\%$ \\
            $V_{\rm L}$ & $-58.0$ & $-58.129$ & $0.2\%$ \\
            $V_{\rm K}$ & $-85.0$ & $-82.431$ & $3.0\%$ \\
            $V_{\rm Na}$ & $50.0$ & $50.660$ & $1.3\%$ \\
            \hline
        \end{tabular}}
    \end{subtable}
    \begin{subtable}[h]{0.6\textwidth}
        \centering
        \caption{($10\%$ noise)}
        \label{tab:pBC_est10}
        \resizebox{\textwidth}{!}{
         \begin{tabular}{c|c|c|c}
            \hline
            Parameter & True Value & Initial Guess: 1 & Relative Error \\
            \hline
            $g_{\rm NaP}$ & $2.0$ & $1.897$ & $5.2\%$ \\
            $g_{\rm L}$ & $2.3$ & $2.066$ & $10.2\%$ \\
            $g_{\rm K}$ & $11.2$ & $12.341$ & $10.2\%$ \\
            $g_{\rm Na}$ & $28.0$ & $28.956$ & $3.4\%$ \\
            $V_{\rm L}$ & $-58.0$ & $-58.103$ & $0.2\%$ \\
            $V_{\rm K}$ & $-85.0$ & $-81.421$ & $4.2\%$ \\
            $V_{\rm Na}$ & $50.0$ & $48.299$ & $3.4\%$ \\
            \hline
        \end{tabular}}
    \end{subtable}
\end{table}

\begin{figure}[t]
    \centering
    \includegraphics[width=\linewidth]{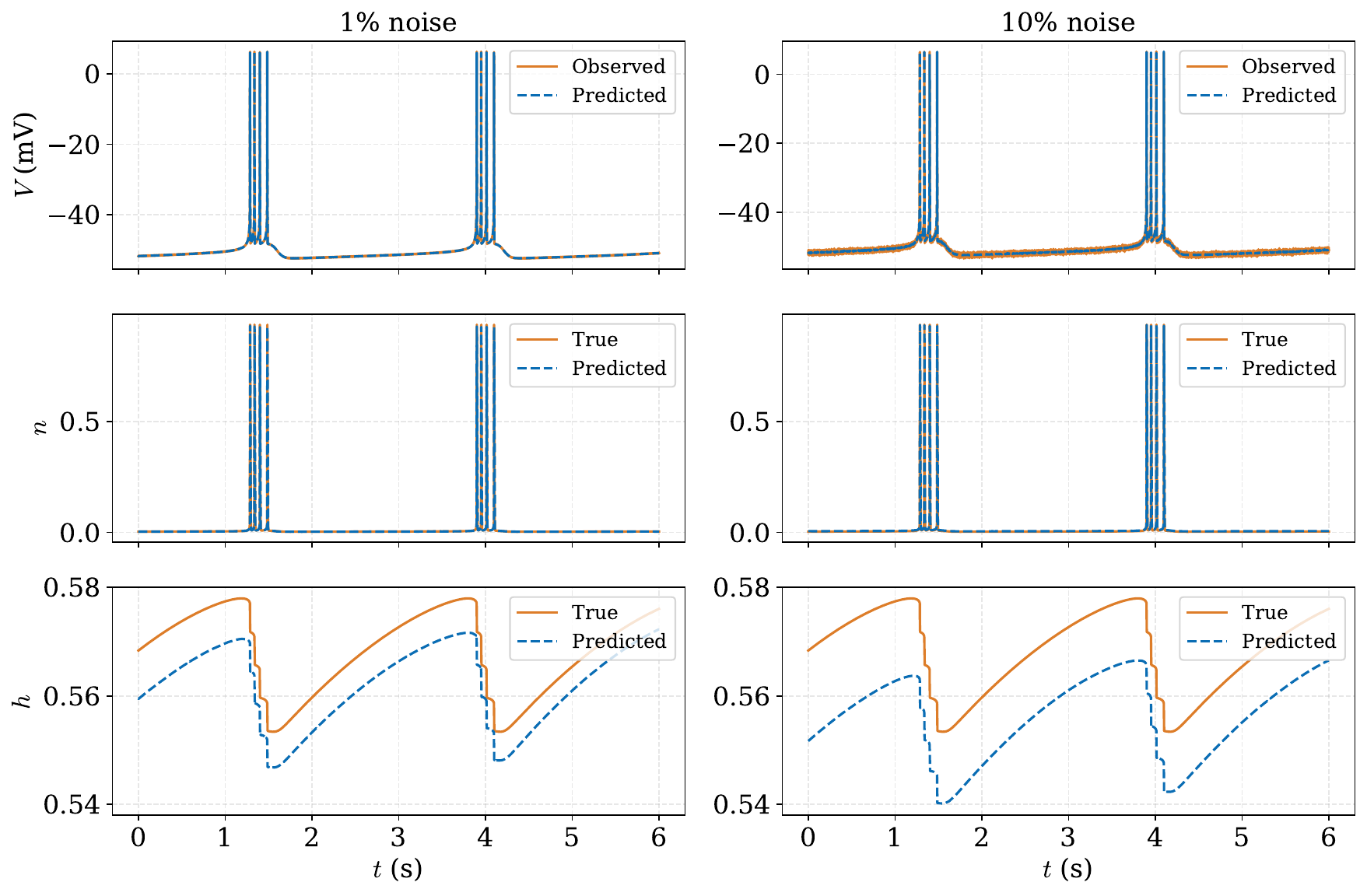}
    \caption{\small{Comparison between the ground-truth trajectories and the PINN-reconstructed solutions for the pBC model under two noise levels based on the non-informative initial guess, where the PINN predictions shown as dashed blue line and true trajectories shown as solid orange line.}}
    \label{fig:pBC_Prediction}
\end{figure}

{\bf Forward trajectory reconstruction.} 
Figure~\ref{fig:pBC_Prediction} illustrates the forward state reconstruction performance of the PINN framework for the pBC model under both $1\%$ and $10\%$ noise levels, using a non-informative initial guess with all parameters set to $1$. 
In both cases, the PINN-predicted solutions closely match the ground‑truth solutions, successfully capturing the temporal structure of spiking activity as well as the unobserved gating dynamics. Although the reconstruction of the gating variable 
$h$ shows some magnitude deviation under higher noise, the phase error remains negligible, indicating that the PINN preserves the essential dynamical structure even under elevated noise. 

Under the small-noise level ($1\%$), the PINN achieves highly accurate forward reconstruction across all state variables. As summarized in Table~\ref{tab:pBC_prediction_eror}, the relative $L^2$ errors remain below approximately $1.2\%$ for all variables. When the observational noise level is increased to $10\%$, reconstruction errors increase moderately. In this case, the relative $L^2$ error of the voltage remains below $0.2\%$, while the errors for the unobserved gating variables stay below approximately $4\%$. 

\begin{table}[t]
    \centering
    \caption{\small{Comparison between the ground-truth trajectories and the PINN-reconstructed solutions for pBC model under two noise levels using state reconstruction errors.}}
    \label{tab:pBC_prediction_eror}
    \resizebox{0.4\textwidth}{!}{
    \begin{tabular}{c|c|c|c}
        \hline
        \textbf{Noise Level} & \textbf{$e_{\hat{V}}$} & \textbf{$e_{\hat{n}}$} & \textbf{$e_{\hat{h}}$} \\
        \hline
        $1\%$ & $0.05\%$ & $1.14\%$ & $1.09\%$ \\
        \hline
        $10\%$ & $0.14\%$ & $3.65\%$ & $2.18\%$ \\
        \hline
    \end{tabular}}
\end{table}

\subsubsection{Absolute noise setting}
\begin{figure}[t]
    \centering
    \includegraphics[width=\linewidth]{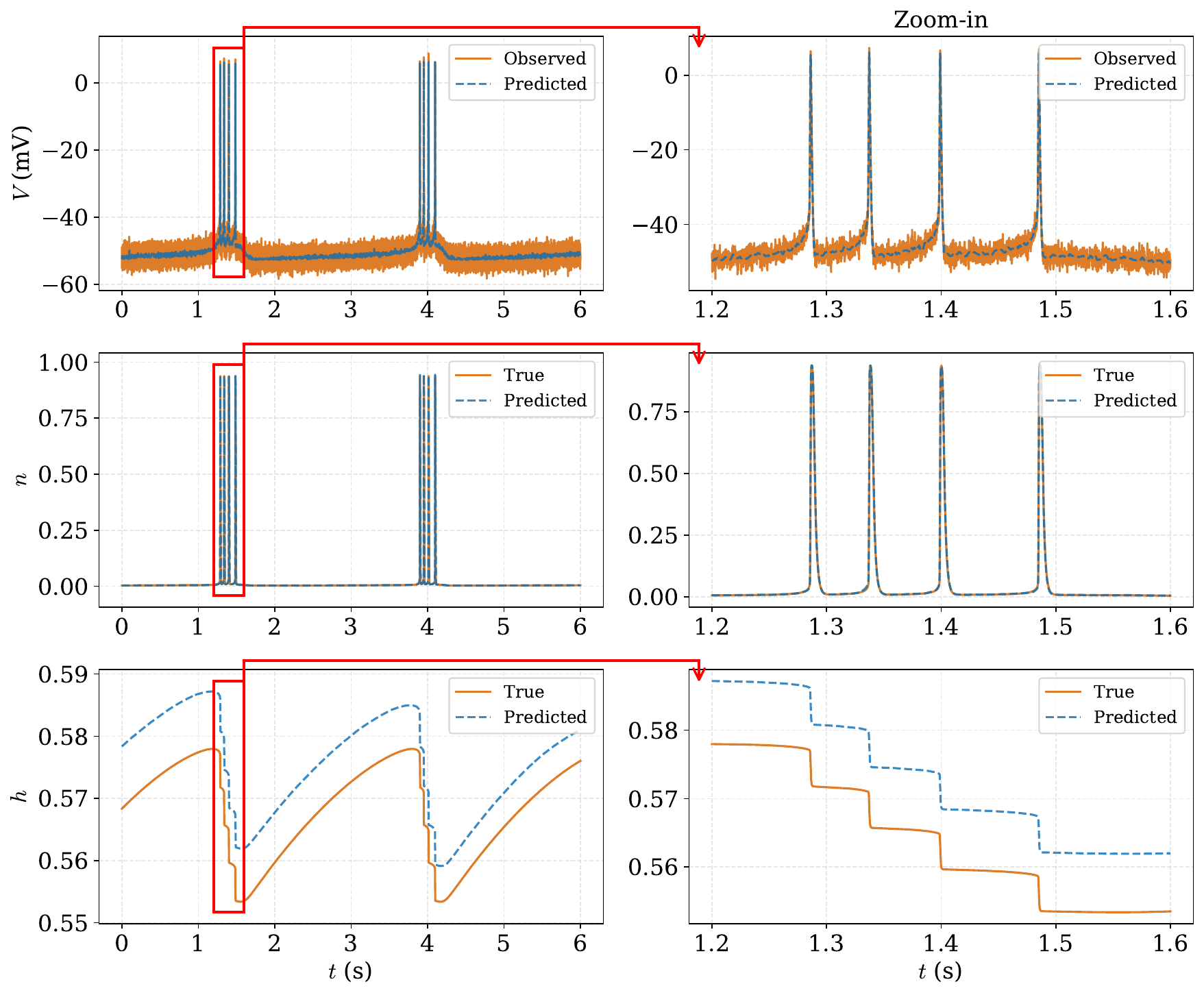}
    \caption{\small{Comparison between the ground-truth trajectories and the PINN-reconstructed solutions for the pBC model under the absolute observational noise level, where the PINN predictions shown as dashed blue line and true trajectories shown as solid orange line.}}
    \label{fig:pBC_Prediction_with_zoom}
\end{figure}

We further challenge our algorithm by considering a more realistic noise setting using an absolute observational noise setting: 
\begin{equation}
V^{\text{obs}}(t) = V^{\text{true}}(t) +
    r\,\mathbb{E}\!\left[ V^{\text{true}}(t) \right]\varepsilon(t), \quad
    \varepsilon(t) \sim \mathcal{N}(0,1),
\end{equation}
where $\mathbb{E}[\cdot]$ denotes the empirical mean over the observation window, and here $r = 0.03$ is chosen to control the noise amplitude. Compared with the relative noise settings considered previously, this formulation introduces a substantially stronger and more realistic level of observational noise (compare Figures \ref{fig:pBC_Prediction} and \ref{fig:pBC_Prediction_with_zoom}).

{\bf Parameter estimation.} 
Table~\ref{tab:pBC_est3} reports the estimated biophysical parameters of the pBC model under the absolute observational noise setting, using a non-informative initial guess with all parameters initialized to $1$. Despite the noise having a broader distribution than that in the relative noise setting, the PINN framework remains capable of recovering all biophysical parameters with reasonable accuracy. Most parameters are recovered with relative errors below $5\%$, while a small subset (e.g., $g_{\rm NaP}$, $g_{\rm Na}$, $V_{\rm Ca}$) exhibits moderately larger deviations on the order of $10\%$. Notably, even these larger errors remain within a moderate range given the severity of the imposed noise.

\begin{table}[t]
    \centering   
    \caption{Estimated biophysical parameters for the pBC model under the absolute observational noise level, evaluated using non-informative initial guess where all parameters are set to $1$.}
    \label{tab:pBC_est3}
    \resizebox{0.6\textwidth}{!}{
     \begin{tabular}{c|c|c|c}
        \hline
        Parameter & True Value & Initial Guess: 1 & Relative Error \\
        \hline
        $g_{\rm NaP}$ & $2.0$ & $1.807$ & $9.6\%$ \\
        $g_{\rm L}$ & $2.3$ & $2.319$ & $0.8\%$ \\
        $g_{\rm K}$ & $11.2$ & $11.710$ & $4.6\%$ \\
        $g_{\rm Na}$ & $28.0$ & $25.158$ & $10.2\%$ \\
        $V_{\rm L}$ & $-58.0$ & $-57.808$ & $0.3\%$ \\
        $V_{\rm K}$ & $-85.0$ & $-81.790$ & $3.8\%$ \\
        $V_{\rm Na}$ & $50.0$ & $55.594$ & $11.2\%$ \\
        \hline
    \end{tabular}}
\end{table}

{\bf Forward trajectory reconstruction.} 
Figure~\ref{fig:pBC_Prediction_with_zoom} presents the forward state solutions performance of the PINN framework for the pBC model under the absolute observational noise setting. The reconstructed trajectories of the voltage $V(t)$ and the gating variables $n(t)$ and $h(t)$ are compared with the corresponding ground-truth solutions, and representative spike windows are highlighted through zoom-in views. The zoom-in panels further indicate that the dominant reconstruction discrepancy arises from small amplitude offsets rather than phase errors, while the spike timing and waveform morphology are well preserved. Despite the strong noise contamination in the voltage observations, the PINN-recovered model accurately captures the timing and shape of all spiking events within each burst, as well as the dynamics of the two unobserved gating variable. As quantified in Table~\ref{tab:pBC_prediction_abs_eror}, the state reconstruction errors for all state variables remain below $2\%$.

\begin{table}[t]
    \centering
    \caption{\small{Comparison between the ground-truth trajectories and the PINN-reconstructed solutions for pBC model under the absolute observational noise level using state reconstruction errors.}}
    \label{tab:pBC_prediction_abs_eror}
    \resizebox{0.3\textwidth}{!}{
    \begin{tabular}{c|c|c}
        \hline
         \textbf{$e_{\hat{V}}$} & \textbf{$e_{\hat{n}}$} & \textbf{$e_{\hat{h}}$} \\
        \hline
         $0.58\%$ & $1.88\%$ & $1.32\%$ \\
        \hline
    \end{tabular}}
\end{table}

\subsubsection{Bifurcation diagrams} 

\begin{figure}[!htp]
    \centering
    \includegraphics[width=0.32\linewidth]{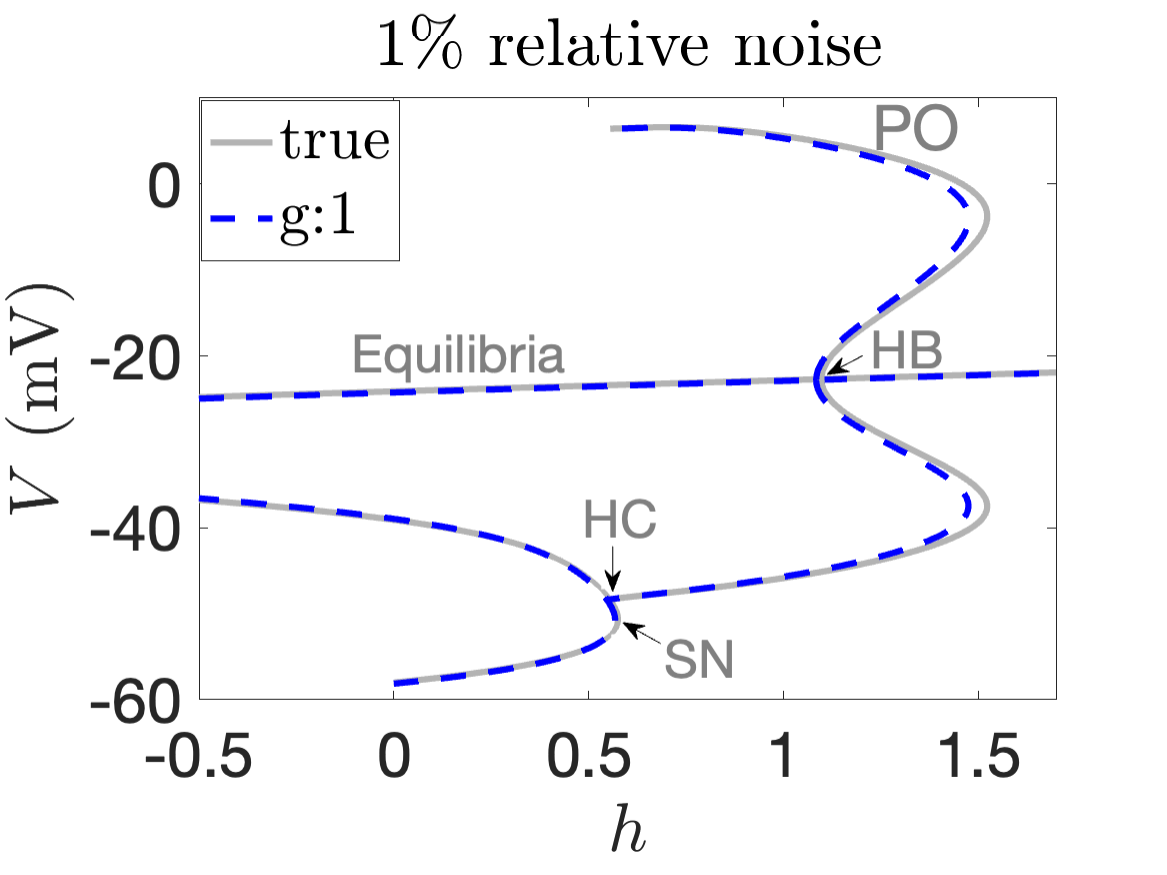}
    \includegraphics[width=0.32\linewidth]{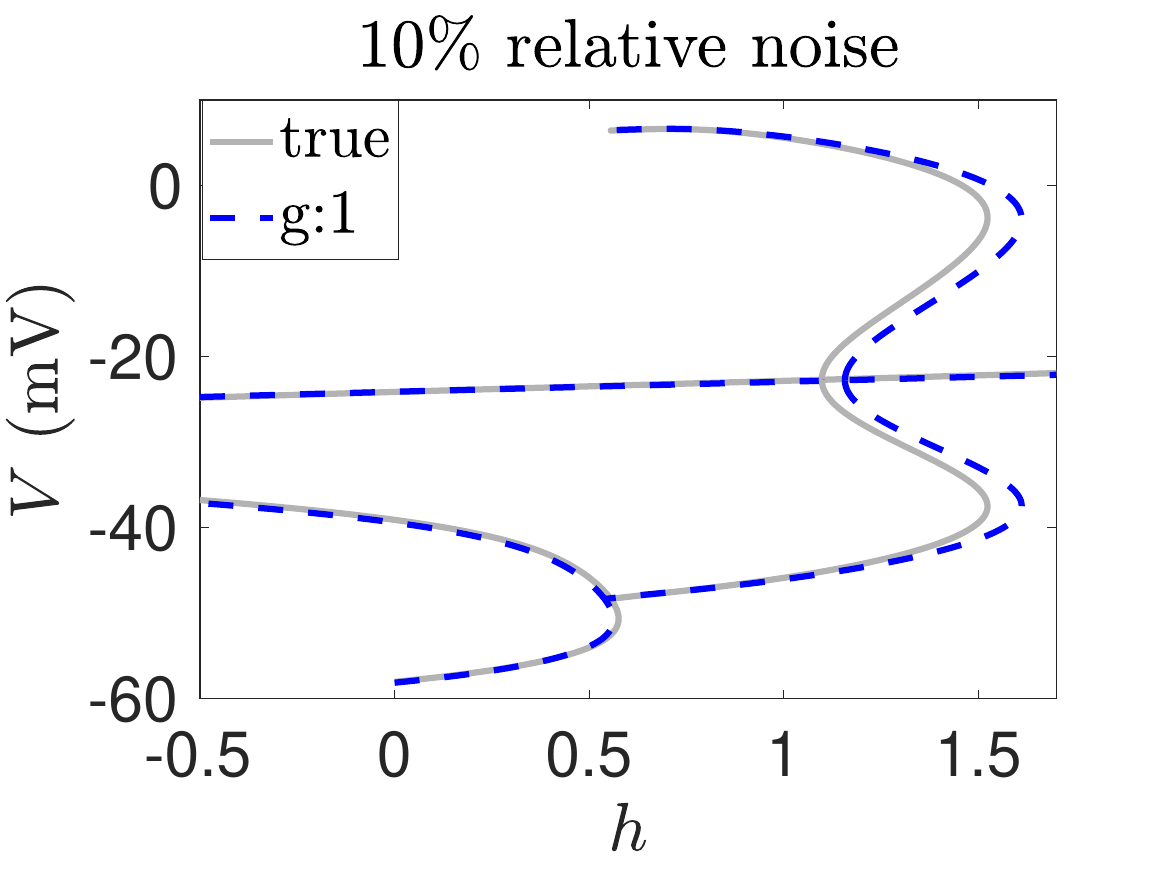}
    \includegraphics[width=0.32\linewidth]{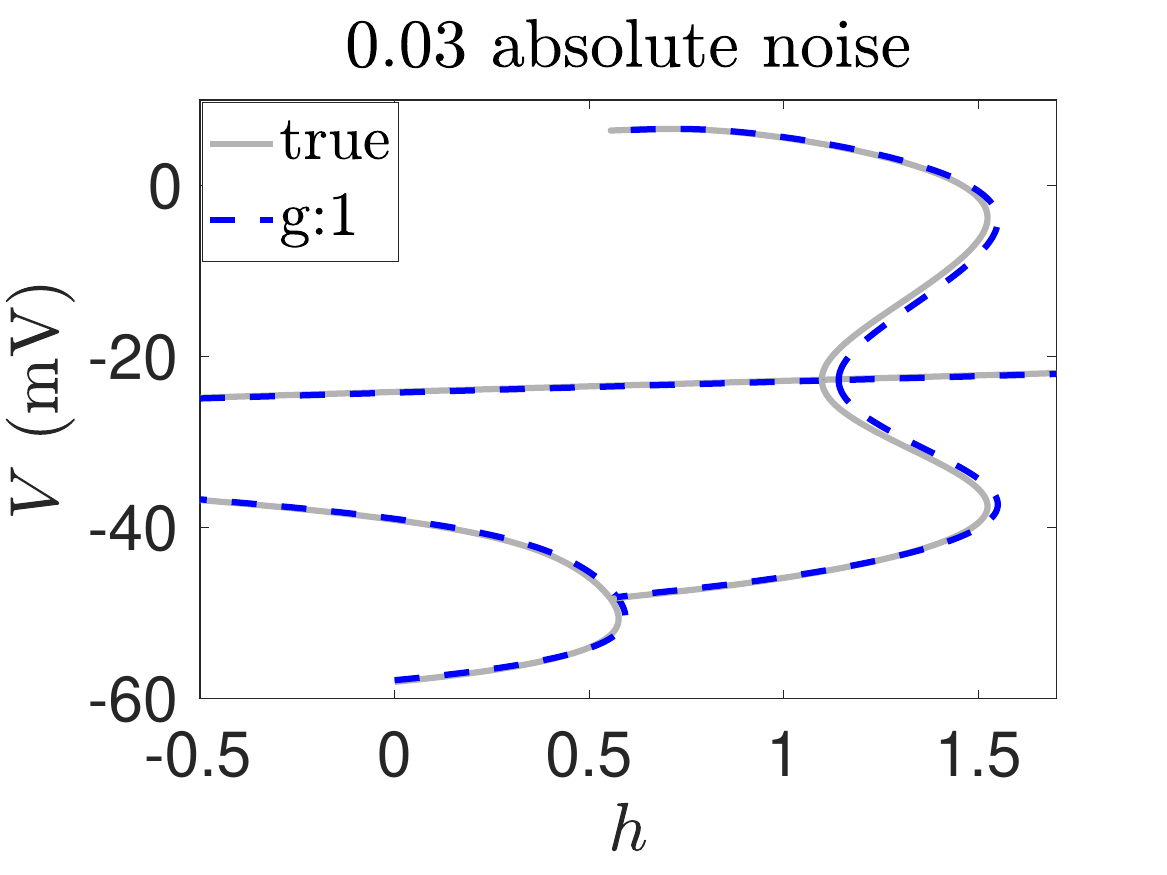}
    \caption{\label{fig:pbc-bif}Bifurcation diagrams for the pBC Model. The gray lines represent the true diagram, while the blue dotted lines correspond to the diagrams generated using the PINN-estimated parameters obtained from a noninformative initial guess (g:1) under (left) 1\% relative noise, (middle) 10\% relative noise, and (right) 3\% absolute observational noise levels.}
\end{figure}

The bifurcation diagrams of the PINN-estimated pBC models, with the slow variable $h$ treated as a bifurcation parameter, under all three noise levels are shown in Figure \ref{fig:pbc-bif}. The case with $1\%$ relative noise exhibits the closest agreement with the ground-truth bifurcation structure. Interestingly, although the noise level is substantially larger in the third case, the corresponding reconstructed bifurcation diagram appears closer to the ground truth than that obtained under the $10\%$ relative noise setting, despite being associated with a less accurate parameter estimates and an overall larger relative errors. This suggests that accurate recovery of qualitative bifurcation structure does not necessarily require uniformly minimal parameter estimation errors, and highlights the necessity of validating algorithm performance using both quantitative comparisons of parameters and trajectories and qualitative comparisons of bifurcation structures.
Overall, the consistently strong agreement between the estimated and true bifurcation structures further highlights the effectiveness of our PINN framework in reconstructing the underlying model, despite partial observations, high levels of observational noise, non-informative initial guesses, and the presence of fast-slow dynamics. 

\section{Discussion}
\label{sec:summary}

This work presents a physics-informed neural network framework for joint state reconstruction and parameter estimation in conductance-based neuronal dynamical systems under partial observability. From a mathematical perspective, the proposed framework integrates several recent advances in physics-informed learning to address inverse problems for stiff, multiscale ordinary differential equations, including Fourier feature embeddings to resolve fast–slow temporal dynamics, random weight factorization decouples the scale and direction of weight matrices in neural networks, which improves the conditioning of the optimization landscape and stabilizes gradient propagation across layers, a two-stage training strategy to stabilize optimization, adaptive loss balancing with residual scaling to improve training convergence, and separate learning rate schedules for network and biophysical parameters to accommodate their distinct training dynamics. From a biological perspective, these methodological developments enable accurate recovery of biophysical parameters and hidden state variables from a single short-duration voltage trace in conductance-based neuron models across multiple dynamical regimes, including type I and type II spiking, as well as square-wave and elliptic bursting \cite{ermentrout2010mathematical}. Across these regimes, the proposed approach achieves stable inference over a range of noise levels and initialization conditions, demonstrating robustness in multiscale neuronal dynamics.
More broadly, the framework offers a general tool for parameter inference in multiscale dynamical systems, with potential applications to other biological processes involving fast-slow interactions \cite{keener2009mathematicalI,keener2009mathematicalII}. 

Our results suggest that the primary strength of the PINN framework lies in its robustness to the choice of initial parameter guess under different noise levels. While classical data assimilation methods can achieve high estimation accuracy when initialized with parameter values that are close to the true dynamical regime, their performance can degrade substantially when such prior knowledge is unavailable, particularly when initial guesses are far from the true regime. In contrast, the proposed PINN-based approach consistently converges to accurate solutions even when initialized from non-informative or regime-mismatched parameter values, making it well suited to experimental settings where reliable prior estimates are difficult to obtain. Moreover, the proposed PINN framework achieves accurate state reconstruction and parameter inference when only a limited number of oscillatory cycles are observed, while sequential filtering methods, such as UKF, generally rely on sufficiently long observation windows to drive the joint convergence of hidden states and unknown parameters. We remark that although our primary focus is on limited-data scenarios, the proposed method remains effective when longer time-series recordings are available.

To validate the accuracy of the proposed PINN framework, we complemented quantitative error metrics with a geometric comparison of bifurcation structures. Specifically, we compared the bifurcation diagrams of the PINN-estimated models with those of the true system, providing an additional qualitative assessment of estimation success. This step is particularly important because even small discrepancies in biophysical parameter estimates can shift a model into fundamentally different dynamical regimes \cite{izhikevich2007dynamical}. Remarkably, the estimated models recovered the correct bifurcation structure with high accuracy even when the algorithm was initialized in a parameter region corresponding to a different dynamical regime, or from non-informative initial guesses in which all parameters were set be $1$. Notably, the underlying geometric and timescale structure was not encoded in the training procedure, yet it emerged naturally from the inference process. A promising future direction is to integrate known geometric constraints into the PINN inference framework to further enhance reliability (e.g., \cite{tien2008parameter}). 

Several limitations of the present study warrant discussion. First, while the proposed PINN framework shows a remarkable robustness to initialization, its performance depends on the choice of network architecture, feature representations, and loss-weighting strategies, which may require problem-specific tuning. Developing more principled guidelines for selecting these components remains an important direction for future work. Second, the numerical experiments in this study are conducted on synthetic data generated from established conductance-based models. Applying the proposed framework to realistic experimental datasets will be essential for evaluating its practical effectiveness and for gaining further insight into the mechanisms underlying neuronal excitability and rhythm generation under physiological and pathological conditions. Experimental neuronal recordings inherently exhibit complex multiscale spiking or bursting dynamics, undergo developmental changes in intrinsic cellular properties, and are constantly shaped by various neuromodulators  \cite{wang2017timescales,wang2020complex,venkatakrishnan2025dual,tryba2026noradrenergic,tien2008parameter}. Extending the framework to such settings would provide a powerful tool for inferring underlying modulatory mechanisms and potential pathological alternations from limited voltage observations. In such problems, incorporating uncertainty quantification into the proposed framework would be particularly beneficial for assessing confidence in inferred parameters, hidden state trajectories, and dynamical regimes.
Finally, the present work focuses on single neuron  systems. Extending the framework to neuronal networks will require careful consideration of computational cost, scalability, and parameter identifiability. Developing efficient training strategies and regularization techniques for network-level inference represents an important and challenging avenue for future research.

\section{Acknowledgment}

Y. Wang was partially supported by National Institutes of Health (NIH) under Award No. NIH/NIDA R01DA057767, as part of the Collaborative Research in Computational Neuroscience Program. X. Zhu was supported
by the Simons Foundation (MPS-TSM-00007740).

\bibliographystyle{abbrv}
\bibliography{biblio}

@article{tryba2026noradrenergic,
  title={Noradrenergic neuromodulation produces a NMDAR-dependent network state of respiratory rhythmogenesis in the preBotzinger Complex},
  author={Tryba, Andrew Kieran and Viemari, Jean-Charles and Wang, Yangyang and Garcia III, Alfredo},
  journal={bioRxiv},
  pages={2026--02},
  year={2026},
  publisher={Cold Spring Harbor Laboratory}
}

@article{tien2008parameter,
  title={Parameter estimation for bursting neural models},
  author={Tien, Joseph H and Guckenheimer, John},
  journal={Journal of Computational Neuroscience},
  volume={24},
  number={3},
  pages={358--373},
  year={2008},
  publisher={Springer}
}

@article{bano2021daily,
  title={Daily electrical activity in the master circadian clock of a diurnal mammal},
  author={Bano-Otalora, Beatriz and Moye, Matthew J and Brown, Timothy and Lucas, Robert J and Diekman, Casey O and Belle, Mino DC},
  journal={Elife},
  volume={10},
  pages={e68179},
  year={2021},
  publisher={eLife Sciences Publications Limited}
}

@book{asch2016data,
  title={Data assimilation: methods, algorithms, and applications},
  author={Asch, Mark and Bocquet, Marc and Nodet, Ma{\"e}lle},
  year={2016},
  publisher={SIAM}
}

@article{ghorbani2025physics,
  title={Physics-aware tuning of the unscented Kalman filter: statistical framework for solving inverse problems involving nonlinear dynamical systems and missing data},
  author={Ghorbani, Esmaeil and Dollon, Quentin and Gosselin, Frederick P},
  journal={Nonlinear Dynamics},
  volume={113},
  number={5},
  pages={4301--4323},
  year={2025},
  publisher={Springer}
}

@article{iglesias2011mixed,
  title={Mixed mode oscillations in mouse spinal motoneurons arise from a low excitability state},
  author={Iglesias, Caroline and Meunier, Claude and Manuel, Marin and Timofeeva, Yulia and Delestr{\'e}e, Nicolas and Zytnicki, Daniel},
  journal={Journal of Neuroscience},
  volume={31},
  number={15},
  pages={5829--5840},
  year={2011},
  publisher={Society for Neuroscience}
}

@article{moehlis2006canards,
  title={Canards for a reduction of the Hodgkin-Huxley equations},
  author={Moehlis, Jeff},
  journal={Journal of mathematical biology},
  volume={52},
  number={2},
  pages={141--153},
  year={2006},
  publisher={Springer}
}

@article{desroches2012mixed,
  title={Mixed-mode oscillations with multiple time scales},
  author={Desroches, Mathieu and Guckenheimer, John and Krauskopf, Bernd and Kuehn, Christian and Osinga, Hinke M and Wechselberger, Martin},
  journal={Siam Review},
  volume={54},
  number={2},
  pages={211--288},
  year={2012},
  publisher={SIAM}
}

@book{keener2009mathematicalII,
  title={Mathematical physiology: II: Systems physiology},
  author={Keener, James and Sneyd, James},
  year={2009},
  publisher={Springer}
}

@book{keener2009mathematicalI,
  title={Mathematical physiology I: Cellular physiology},
  author={Keener, JP and Sneyd, James},
  volume={2},
  year={2009},
  publisher={Springer New York, NY, USA}
}

@article{park2013cooperation,
  title={Cooperation of intrinsic bursting and calcium oscillations underlying activity patterns of model pre-B{\"o}tzinger complex neurons},
  author={Park, Choongseok and Rubin, Jonathan E},
  journal={Journal of Computational Neuroscience},
  volume={34},
  number={2},
  pages={345--366},
  year={2013},
  publisher={Springer}
}

@article{venkatakrishnan2025dual,
  title={Dual Mechanisms for Heterogeneous Responses of Inspiratory Neurons to Noradrenergic Modulation},
  author={Venkatakrishnan, Sreshta and Tryba, Andrew K and Wang, Yangyang and others},
  journal={arXiv preprint arXiv:2507.19416},
  year={2025}
}

@article{nan2015understanding,
  title={Understanding and distinguishing three-time-scale oscillations: Case study in a coupled Morris--Lecar system},
  author={Nan, Pingyu and Wang, Yangyang and Kirk, Vivien and Rubin, Jonathan E},
  journal={SIAM journal on applied dynamical systems},
  volume={14},
  number={3},
  pages={1518--1557},
  year={2015},
  publisher={SIAM}
}

@article{phan2024mixed,
  title={Mixed-mode oscillations in a three-timescale coupled Morris--Lecar system},
  author={Phan, Ngoc Anh and Wang, Yangyang},
  journal={Chaos: An Interdisciplinary Journal of Nonlinear Science},
  volume={34},
  number={5},
  year={2024},
  publisher={AIP Publishing}
}

@article{wang2017timescales,
  title={Timescales and mechanisms of sigh-like bursting and spiking in models of rhythmic respiratory neurons},
  author={Wang, Yangyang and Rubin, Jonathan E},
  journal={The Journal of Mathematical Neuroscience},
  volume={7},
  number={1},
  pages={3},
  year={2017},
  publisher={Springer}
}

@book{ermentrout2010mathematical,
  title={Mathematical foundations of neuroscience},
  author={Ermentrout, Bard and Terman, David Hillel},
  volume={35},
  year={2010},
  publisher={Springer}
}

@article{wang2020complex,
  title={Complex bursting dynamics in an embryonic respiratory neuron model},
  author={Wang, Yangyang and Rubin, Jonathan E},
  journal={Chaos: An Interdisciplinary Journal of Nonlinear Science},
  volume={30},
  number={4},
  year={2020},
  publisher={AIP Publishing}
}

@article{morris1981voltage,
  title={Voltage oscillations in the barnacle giant muscle fiber},
  author={Morris, Catherine and Lecar, Harold},
  journal={Biophysical journal},
  volume={35},
  number={1},
  pages={193--213},
  year={1981},
  publisher={Elsevier}
}

@article{del2002persistent,
  title={Persistent sodium current, membrane properties and bursting behavior of pre-botzinger complex inspiratory neurons in vitro},
  author={Del Negro, Christopher A and Koshiya, Naohiro and Butera Jr, Robert J and Smith, Jeffrey C},
  journal={Journal of neurophysiology},
  volume={88},
  number={5},
  pages={2242--2250},
  year={2002},
  publisher={American Physiological Society Bethesda, MD}
}

@article{smith1991pre,
  title={Pre-B{\"o}tzinger complex: a brainstem region that may generate respiratory rhythm in mammals},
  author={Smith, Jeffrey C and Ellenberger, Howard H and Ballanyi, Klaus and Richter, Diethelm W and Feldman, Jack L},
  journal={Science},
  volume={254},
  number={5032},
  pages={726--729},
  year={1991},
  publisher={American Association for the Advancement of Science}
}

@article{butera1999models,
  title={Models of respiratory rhythm generation in the pre-Botzinger complex. I. Bursting pacemaker neurons},
  author={Butera Jr, Robert J and Rinzel, John and Smith, Jeffrey C},
  journal={Journal of neurophysiology},
  volume={82},
  number={1},
  pages={382--397},
  year={1999},
  publisher={American Physiological Society Bethesda, MD}
}

@book{johnston1994foundations,
  title={Foundations of cellular neurophysiology},
  author={Johnston, Daniel and Wu, Samuel Miao-Sin},
  year={1994},
  publisher={MIT press}
}

@article{kadakia2016nonlinear,
  title={Nonlinear statistical data assimilation for HVC RA neurons in the avian song system},
  author={Kadakia, Nirag and Armstrong, Eve and Breen, Daniel and Morone, Uriel and Daou, Arij and Margoliash, Daniel and Abarbanel, Henry DI},
  journal={Biological cybernetics},
  volume={110},
  number={6},
  pages={417--434},
  year={2016},
  publisher={Springer}
}

@article{kainth2025physics,
  title={Physics-Informed Neural ODEs with Scale-Aware Residuals for Learning Stiff Biophysical Dynamics},
  author={Kainth, Kamalpreet Singh and Joshi, Prathamesh Dinesh and Dandekar, Raj Abhijit and Dandekar, Rajat and Panat, Sreedat},
  journal={arXiv preprint arXiv:2511.11734},
  year={2025}
}

@article{bertaglia2022asymptotic,
  title={Asymptotic-Preserving Neural Networks for multiscale hyperbolic models of epidemic spread},
  author={Bertaglia, Giulia and Lu, Chuan and Pareschi, Lorenzo and Zhu, Xueyu},
  journal={Mathematical Models and Methods in Applied Sciences},
  volume={32},
  number={10},
  pages={1949--1985},
  year={2022},
  publisher={World Scientific}
}

@article{wang2021understanding,
  title={Understanding and mitigating gradient flow pathologies in physics-informed neural networks},
  author={Wang, Sifan and Teng, Yujun and Perdikaris, Paris},
  journal={SIAM Journal on Scientific Computing},
  volume={43},
  number={5},
  pages={A3055--A3081},
  year={2021}
}

@article{hodgkin1952quantitative,
  title={A quantitative description of membrane current and its application to conduction and excitation in nerve},
  author={Hodgkin, Alan L and Huxley, Andrew F},
  journal={The Journal of Physiology},
  volume={117},
  number={4},
  pages={500--544},
  year={1952}
}

@book{izhikevich2007dynamical,
  title={Dynamical Systems in Neuroscience: The Geometry of Excitability and Bursting},
  author={Izhikevich, Eugene M},
  publisher={MIT Press},
  year={2007}
}

@article{paninski2009statistical,
  title={Statistical models for neural encoding, decoding, and optimal stimulus design},
  author={Paninski, Liam and Pillow, Jonathan and Lewi, Jeremy},
  journal={Progress in Brain Research},
  volume={165},
  pages={493--507},
  year={2009}
}

@article{raue2009structural,
  title={Structural and practical identifiability analysis of partially observed dynamical models by exploiting the profile likelihood},
  author={Raue, Andreas and Kreutz, Clemens and Maiwald, Thomas and Bachmann, Jens and Schilling, Moritz and Klingm{\"u}ller, Ursula and Timmer, Jens},
  journal={Bioinformatics},
  volume={25},
  number={15},
  pages={1923--1929},
  year={2009}
}

@article{vlachas2020ukf,
  title={Data-driven forecasting of high-dimensional chaotic systems with long short-term memory networks},
  author={Vlachas, Pantelis R and Pathak, Jaideep and Hunt, Brian R and Sapsis, Themistoklis P and Girvan, Michelle and Ott, Edward and Koumoutsakos, Petros},
  journal={Proceedings of the Royal Society A},
  volume={476},
  number={2240},
  pages={20200349},
  year={2020}
}

@article{li2023adaptiveUKF,
  title={Adaptive unscented Kalman filtering for nonlinear systems with unknown noise statistics},
  author={Li, Xiaoyan and Zhang, Wei and Wang, Jun},
  journal={IEEE Transactions on Automatic Control},
  year={2023},
  note={early access}
}

@article{golightly2011bayesian,
  title={Bayesian parameter inference for stochastic biochemical network models using particle Markov chain Monte Carlo},
  author={Golightly, Andrew and Wilkinson, Darren J},
  journal={Interface Focus},
  volume={1},
  number={6},
  pages={807--820},
  year={2011}
}

@article{calderhead2011accelerating,
  title={Accelerating Bayesian inference over nonlinear differential equations with Gaussian processes},
  author={Calderhead, Ben and Girolami, Mark and Lawrence, Neil D},
  journal={Advances in Neural Information Processing Systems},
  volume={24},
  year={2011}
}

@article{transtrum2015sloppiness,
  title={Sloppiness and emergent theories in physics, biology, and beyond},
  author={Transtrum, Mark K and Machta, Benjamin B and Sethna, James P},
  journal={Physical Review E},
  volume={83},
  number={3},
  pages={036701},
  year={2011}
}

@article{gonccalves2023sbi,
  title={Training deep neural density estimators to identify mechanistic models of neural dynamics},
  author={Gon{\c{c}}alves, Pedro J and Lueckmann, Jan-Matthis and Deistler, Michael and others},
  journal={PLOS Computational Biology},
  volume={19},
  number={1},
  pages={e1010836},
  year={2023}
}

@article{zhang2022neuralsbi,
  title={Neural posterior estimation for mechanistic models of neural dynamics},
  author={Zhang, Yi and Wang, Rui and Paninski, Liam},
  journal={Neurocomputing},
  volume={475},
  pages={56--68},
  year={2022}
}

@article{raissi2019physics,
  title={Physics-informed neural networks: A deep learning framework for solving forward and inverse problems involving nonlinear partial differential equations},
  author={Raissi, Maziar and Perdikaris, Paris and Karniadakis, George Em},
  journal={Journal of Computational Physics},
  volume={378},
  pages={686--707},
  year={2019}
}

@article{alamstotter2022hhpinn,
  title={Physically constrained neural networks for the Hodgkin--Huxley model},
  author={Alamst{\"o}tter, Markus and others},
  journal={arXiv preprint arXiv:2209.11998},
  year={2022}
}

@article{moye2018data,
  title={Data assimilation methods for neuronal state and parameter estimation},
  author={Moye, Matthew J and Diekman, Casey O},
  journal={The Journal of Mathematical Neuroscience},
  volume={8},
  pages={1--38},
  year={2018},
  publisher={Springer}
}

@article{WANG2021113938,
title = {On the eigenvector bias of Fourier feature networks: From regression to solving multi-scale PDEs with physics-informed neural networks},
journal = {Computer Methods in Applied Mechanics and Engineering},
volume = {384},
pages = {113938},
year = {2021},
issn = {0045-7825},
doi = {https://doi.org/10.1016/j.cma.2021.113938},
url = {https://www.sciencedirect.com/science/article/pii/S0045782521002759},
author = {Sifan Wang and Hanwen Wang and Paris Perdikaris},
}

@misc{tancik2020,
      title={Fourier Features Let Networks Learn High Frequency Functions in Low Dimensional Domains}, 
      author={Matthew Tancik and Pratul P. Srinivasan and Ben Mildenhall and Sara Fridovich-Keil and Nithin Raghavan and Utkarsh Singhal and Ravi Ramamoorthi and Jonathan T. Barron and Ren Ng},
      year={2020},
      eprint={2006.10739},
      archivePrefix={arXiv},
      primaryClass={cs.CV},
      url={https://arxiv.org/abs/2006.10739}, 
}

@article{wang2016multiple,
  title={Multiple timescale mixed bursting dynamics in a respiratory neuron model},
  author={Wang, Yangyang and Rubin, Jonathan E},
  journal={Journal of Computational Neuroscience},
  volume={41},
  pages={245--268},
  year={2016},
  publisher={Springer}
}

@misc{rinzel1989analysis,
  title={Analysis of neural excitability and oscillations, Methods of Neural Modeling: From Synapses to Networks (C. Koch and I. Segev, eds.)},
  author={Rinzel, J and Ermentrout, B},
  year={1989},
  publisher={MIT Press}
}

@misc{wang2023,
      title={An Expert's Guide to Training Physics-informed Neural Networks}, 
      author={Sifan Wang and Shyam Sankaran and Hanwen Wang and Paris Perdikaris},
      year={2023},
      eprint={2308.08468},
      archivePrefix={arXiv},
      primaryClass={cs.LG},
      url={https://arxiv.org/abs/2308.08468}, 
}

@misc{wang2022randomweightfactorizationimproves,
      title={Random Weight Factorization Improves the Training of Continuous Neural Representations}, 
      author={Sifan Wang and Hanwen Wang and Jacob H. Seidman and Paris Perdikaris},
      year={2022},
      eprint={2210.01274},
      archivePrefix={arXiv},
      primaryClass={cs.LG},
      url={https://arxiv.org/abs/2210.01274}, 
}

@inproceedings{glorot2010,
  title={Understanding the difficulty of training deep feedforward neural networks},
  author={Glorot, Xavier and Bengio, Yoshua},
  booktitle={Proceedings of the thirteenth international conference on artificial intelligence and statistics},
  pages={249--256},
  year={2010},
  organization={JMLR Workshop and Conference Proceedings}
}

@article{RAISSI2019686,
title = {Physics-informed neural networks: A deep learning framework for solving forward and inverse problems involving nonlinear partial differential equations},
journal = {Journal of Computational Physics},
volume = {378},
pages = {686-707},
year = {2019},
issn = {0021-9991},
doi = {https://doi.org/10.1016/j.jcp.2018.10.045},
url = {https://www.sciencedirect.com/science/article/pii/S0021999118307125},
author = {M. Raissi and P. Perdikaris and G.E. Karniadakis},
}

@misc{kingma2017adam,
      title={Adam: A Method for Stochastic Optimization}, 
      author={Diederik P. Kingma and Jimmy Ba},
      year={2017},
      eprint={1412.6980},
      archivePrefix={arXiv},
      primaryClass={cs.LG},
      url={https://arxiv.org/abs/1412.6980}, 
}

@article{liu1989lbfgs,
  title   = {On the Limited Memory BFGS Method for Large Scale Optimization},
  author  = {Liu, Dong C. and Nocedal, Jorge},
  journal = {Mathematical Programming},
  volume  = {45},
  number  = {1},
  pages   = {503--528},
  year    = {1989}
}

@article{karniadakis2021physics,
  title={Physics-informed machine learning},
  author={Karniadakis, George Em and Kevrekidis, Ioannis G and Lu, Lu and Perdikaris, Paris and Wang, Sifan and Yang, Liu},
  journal={Nature Reviews Physics},
  volume={3},
  number={6},
  pages={422--440},
  year={2021},
  publisher={Nature Publishing Group}
}

@techreport{baker_2019,
title = {Workshop Report on Basic Research Needs for Scientific Machine Learning: Core Technologies for Artificial Intelligence},
author = {Baker, Nathan and Alexander, Frank and Bremer, Timo and Hagberg, Aric and Kevrekidis, Yannis and Najm, Habib and Parashar, Manish and Patra, Abani and Sethian, James and Wild, Stefan and Willcox, Karen and Lee, Steven},
doi = {10.2172/1478744},
url = {https://www.osti.gov/biblio/1478744},
institution = {Office of Scientific and Technical Information},
place = {United States},
year = {2019},
month = {2}
}

\end{document}